\documentclass[preprint,a4paper]{elsarticle}

\usepackage{graphicx}
\usepackage{amsmath}
\usepackage{float}
\usepackage{amsfonts}
\usepackage{a4wide}
\usepackage{algorithm}
\usepackage{algorithmicx}
\usepackage{algpseudocode}
\usepackage{epstopdf}
\usepackage{bm}
\usepackage{tikz}
\usepackage{multirow}
\usepackage{amsmath}
\usetikzlibrary{bayesnet}
\usetikzlibrary{shapes.gates.logic.US,trees,positioning,arrows}
\usepackage{caption}
\usepackage{subcaption}
\usetikzlibrary{trees}
\usepackage{graphicx}
\usepackage{mathtools}
\usepackage{amsmath}
\usepackage{booktabs}
\usepackage{rotating}
\usepackage[page]{appendix}

\usepackage{amsthm}

\usepackage{epstopdf}
\usepackage{amssymb}
\usepackage{microtype}
\usepackage{url}
\usepackage{caption}
\usepackage{subcaption}
\usepackage{cleveref}
\DeclareMathOperator*{\argmax}{argmax}   
\newcommand{\mRVMa}{{mRVM$_1$}}
\newcommand{\mRVMb}{{mRVM$_2$}}

\journal{Mechanical Systems and Signal Processing}

\begin{document}

\begin{frontmatter}

	\title{On robust risk-based active-learning algorithms for enhanced decision support}

	\author[add1]{A.J.\ Hughes\corref{mycorrespondingauthor}}
	\cortext[mycorrespondingauthor]{Corresponding author}
	\ead{ajhughes2@sheffield.ac.uk}

	\author[add2]{L.A.\ Bull}

	\author[add1]{P.\ Gardner}

	\author[add1]{N.\ Dervilis}

	\author[add1]{K.\ Worden}

	\address[add1]{Dynamics Research Group, Department of Mechanical Engineering, University of Sheffield, \\ Sheffield, S1 3JD, UK}

	\address[add2]{The Alan Turing Institute, The British Library, 96 Euston Road, London, NW1 2DB, UK}

	\begin{abstract}
	Classification models are a fundamental component of physical-asset management technologies such as structural health monitoring (SHM) systems and digital twins. Previous work introduced \textit{risk-based active learning}, an online approach for the development of statistical classifiers that takes into account the decision-support context in which they are applied. Decision-making is considered by preferentially querying data labels according to \textit{expected value of perfect information} (EVPI). Although several benefits are gained by adopting a risk-based active learning approach, including improved decision-making performance, the algorithms suffer from issues relating to sampling bias as a result of the guided querying process. This sampling bias ultimately manifests as a decline in decision-making performance during the later stages of active learning, which in turn corresponds to lost resource/utility.
	
	The current paper proposes two novel approaches to counteract the effects of sampling bias: \textit{semi-supervised learning}, and \textit{discriminative classification models}. These approaches are first visualised using a synthetic dataset, then subsequently applied to an experimental case study, specifically, the Z24 Bridge dataset. The semi-supervised learning approach is shown to have variable performance; with robustness to sampling bias dependent on the suitability of the generative distributions selected for the model with respect to each dataset. In contrast, the discriminative classifiers are shown to have excellent robustness to the effects of sampling bias. Moreover, it was found that the number of inspections made during a monitoring campaign, and therefore resource expenditure, could be reduced with the careful selection of the statistical classifiers used within a decision-supporting monitoring system.

	\end{abstract}


\end{frontmatter}

\section{Introduction}

Statistical pattern recognition (SPR) has been established as the state of the art for making data-driven predictions in the context of physical-asset management technologies such as structural health monitoring (SHM) systems \cite{Farrar2013} and digital twins \cite{Grieves2017,Niederer2021}. Statistical classifiers are a fundamental component of the SPR approach to these decision-support technologies -- enabling the categorisation of acquired data into groups, or classes. For example, in the context of SHM, by associating the target classes of a classifier with salient health-states (e.g.\ undamaged, nascent damage, severe damage), inferences can be made regarding the condition of a structure of interest. Recent research has focussed on utilising probabilistic predictions from statistical classifiers in risk-based approaches for supporting the operation and maintenance (O\&M) decision processes engineers are tasked with when managing structural assets \cite{Nielsen2013,Hovgaard2016,Hughes2021}. Defined as a product of probability and cost, risk is akin to negative \textit{expected utility}. Decisions can then be made by selecting strategies that maximise expected utility, or minimise risk.

For SHM and other engineering decision-support applications, the traditional \textit{supervised} and \textit{unsupervised} machine learning paradigms are of limited applicability. This predicament arises because, oftentimes, there is a scarcity of labelled data corresponding to salient damage, operational, and environmental states of a structure.

Active learning is an approach to developing statistical classifiers that is well-suited to online decision-support applications and has seen some application in fields such as SHM \cite{Bull2019,Arellano2019,Chakraborty2015}. Within an active-learning framework, label information is obtained for otherwise unlabelled data via a querying process, such that models may subsequently be learned in a supervised manner. Typically, the querying process is guided according to an uncertainty measure (e.g.\ entropy or likelihood \cite{Bull2019}) given the current classification model. In the context of SHM, new label information queried during the active-learning process corresponds to diagnoses provided by an engineer, following an inspection of a structure. When a statistical classifier is being used within a predefined decision process, \textit{risk-based active learning} can be employed \cite{Hughes2022}. Serving as preliminary work to the current paper, \cite{Hughes2022}, introduces a risk-based formulation of active learning in which queries are guided according
to the \textit{expected value of information} (EVPI) of health-state labels for incipient data. The pre-existing paper contains details on how EVPI may be calculated in the context of SHM decision processes, and how EVPI can be integrated into the active learning process in order to mandate structural inspections. The paper also shows, using two datasets, that decision performance can be more rapidly improved over the course of a monitoring campaign by learning classifiers from data queried according to EVPI. Further details of the work presented in\cite{Hughes2022} are provided in Section \ref{sec:RBAL1} of the current paper. As mentioned, the risk-based approach to active learning requires a predefined decision process. While a brief overview of how such decision processes may be defined is presented in \cite{Hughes2022}, for a detailed explanation of this topic the reader is directed to \cite{Hughes2021}, which comprehensively outlines a methodology for specifying decision processes with reference to a paradigm for SHM that incorporates elements of probabilistic risk assessment such as failure mode analysis using fault trees.

While being apt for dealing with challenges associated with online decision-support systems, active learning algorithms are not without their own pitfalls. Notably, generative models (such as those used in \cite{Bull2019,Hughes2022}), learned via active learning, are often susceptible to sampling bias because the preferential nature of the querying process causes inequity in the amount of data observed for each class. This problem is pertinent, as sampling bias has been found to degrade the performance of classifiers, and decision-making agents alike \cite{Bull2019,Hughes2022,Dasgupta2011}.

\subsection{Novel Contribution}

The current paper proposes and examines two novel methods for addressing sampling bias in active-learning algorithms for asset management technologies, and particularly focusses on the performance of decision-makers utilising the learned classifiers in the context of SHM decision-support.

The first approach involves adapting the algorithm presented in \cite{Hughes2022}, via the introduction of semi-supervised learning. Here, two formulations of semi-supervised learning are considered; expectation-maximisation with respect to the generative model learned in a supervised manner, and latent-state smoothing with respect to the hidden-Markov model that underpins the asset-management decision processes. The second novel approach replaces the generative Gaussian mixture model, used in previous literature \cite{Bull2019,Hughes2022}, with discriminative classifiers. Specifically, two formulations of multiclass relevance vector machines (mRVMs) \cite{Psorakis2010} are considered. In total, these approaches result in four new formulations of risk-based active learning. Finally, further novelty is provided in the form of valuable discussions on the role of statistical classifiers in asset management technology, and how decision-support can be improved via classifier design.

The contents of the current paper are as follows. Background theory is provided for risk-based active learning in Section \ref{sec:RBAL1}. The effects of sampling bias are highlighted using a synthetic dataset in Section \ref{sec:Effects}. Section \ref{sec:Approaches} details the aforementioned approaches to dealing with sampling bias. Section \ref{sec:Z24} introduces the Z24 Bridge dataset as a case study and presents results obtained by applying the proposed methods for addressing sampling bias. A broader discourse on the selection of classifiers for decision-support applications is offered in Section \ref{sec:Discussions}. Finally, concluding remarks are provided in Section \ref{sec:Conclusion}.

\section{Risk-based Active Learning}
\label{sec:RBAL1}

In general, classification is the problem of categorising observations $\mathbf{x} \in X$ according to descriptive labels $y \in Y$, where $X$ and $Y$ denote the \textit{input space} and \textit{label space}, respectively. In learning a statistical classifier, one wishes to obtain a robust mapping from the input space to the label space, i.e.\ $f : X \rightarrow Y$. A probabilistic perspective can be adopted by defining inputs as $D$-dimensional random vectors, such that $X = \mathbb{R}^D$, and descriptive labels as discrete random variables, such that $Y = \{1, \ldots, K\}$, where $K$ is the number of target classes.  In the context of SHM, inputs $\mathbf{x}_i$ correspond to discriminative features extracted from data acquired via a monitoring system. A descriptive label $y_i$ represents structural health, environmental, or operational states pertinent to the O\&M decision process associated with a structure.

Traditionally, statistical classifiers are learned via one of two main paradigms; \textit{supervised} or \textit{unsupervised} machine learning.

In the supervised approach, the mapping $f$ can be learned directly with the use of a fully-labelled training dataset $\mathcal{D}_l$ where \cite{Schwenker2014},

\begin{equation}
	\mathcal{D}_l = \{(\mathbf{x}_i,y_i)|\mathbf{x}_i \in X, y_i \in Y \}^n_{i=1}
\end{equation}

\noindent for $n$ collected data points.

In contrast, unsupervised learning techniques (e.g.\ novelty detection and clustering) seek to identify patterns and structure within unlabelled datasets $\mathcal{D}_u$, where,

\begin{equation}
	\mathcal{D}_u = \{\tilde{\mathbf{x}}_i | \tilde{\mathbf{x}}_i \in X\}^m_{i=1}
\end{equation}

\noindent for $m$ collected data points and where $\tilde{\mathbf{x}}_i$  denote measured data that are unlabelled.

Unfortunately, when considering decision support problems for SHM, the efficacy of both supervised and unsupervised learning techniques are diminished. This is because complete labelled datasets are seldom available prior to the implementation of a monitoring system; specifically, there tends to be a dearth of data corresponding to damage states of the structure. This issue precludes supervised learning because of the absence of comprehensive training targets. Moreover, the issue renders the unsupervised approach inapplicable because the learned patterns within data are without the contextual information necessary for decision-making. An in-depth discussion on the application of supervised and unsupervised machine learning to SHM may be found in \cite{Bull2019}.

\subsection{Active Learning for SHM}

Active learning is a form of \textit{partially-supervised} learning. Partially supervised learning is characterised by the utilisation of both labelled and unlabelled data such that,

\begin{equation}
	\mathcal{D} = \mathcal{D}_l \cup \mathcal{D}_u
\end{equation}

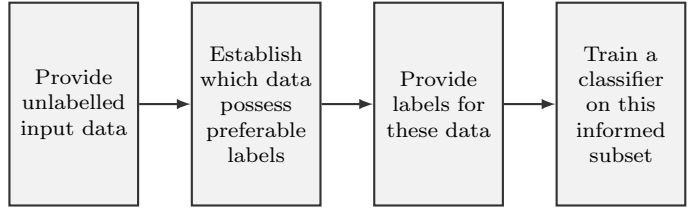
\begin{figure}[ht!]
	\centering
	\begin{tikzpicture}[auto]
		\begin{footnotesize}
			\tikzstyle{block} = [rectangle, thick, draw=black!80, text width=5em, text centered, minimum height=9em, fill=black!5]
			\tikzstyle{line} = [draw, -latex, thick]
			\node [block, node distance=24mm] (A) {Provide\\ unlabelled input data};
			\node [block, right of=A, node distance=24mm] (B) {Establish which data possess preferable labels};
			\node [block, right of=B, node distance=24mm] (C) {Provide labels for these data};
			\node [block, right of=C, node distance=24mm] (D) {Train a classifier on this informed subset};
			\path [line, draw=black!80] (A) -- (B);
			\path [line, draw=black!80] (B) -- (C);
			\path [line, draw=black!80] (C) -- (D);
		\end{footnotesize}
	\end{tikzpicture}
	\caption{A general active learning heuristic.}
	\label{fig:AL1}
\end{figure}

Active learning algorithms construct a labelled data set $\mathcal{D}_l$, by querying labels for the unlabelled data in $D_u$. Labels for incipient data are queried preferentially according to some measure of their desirability. Models are then trained in a supervised manner using $\mathcal{D}_l$. A general heuristic for active learning is presented in Figure \ref{fig:AL1}.

There exists few examples of works applying active learning in the context of health and performance monitoring. In \cite{Feng2017}, active sampling was used in conjunction with artificial neural networks for image classification tasks related to the detection of defects in civil structures. A methodology for tool condition monitoring that incorporates entropy-based active sampling with a Bayesian convolutional neural network is proposed in \cite{Arellano2019}. Moreover, in \cite{Chakraborty2015}, a damage progression model based upon a particle filter aided by actively selected data is presented. In \cite{Bull2019}, an online active learning approach is shown to overcome the primary challenge for classifier development in the context of SHM - initial scarcity of comprehensive labelled data. This result is achieved by the construction of a labelled dataset via a querying process that corresponds to the inspection of a structure by an engineer to determine its health state.
In \cite{Bull2019}, measures of informativeness, high entropy and low likelihood given the current model, are used to guide querying. In \cite{Hughes2022}, an alternative risk-based approach to active learning is proposed in which \textit{expected value of perfect information} (EVPI) is used as a measure to guide querying. The remainder of the current paper will focus on the risk-based formulation of active learning.

\subsection{Expected Value of Perfect Information}

As aforementioned, the risk-based approach to active learning is characterised by the use of an expected value of information as a measure to guide label querying.

The expected value of perfect information (EVPI) can be interpreted as the price that a decision-maker should be willing to pay in order to gain access to perfect information of an otherwise unknown or uncertain state. Here, it should be clarified that the terminology `perfect information' refers to ground-truth states observed without uncertainty.  A more formal definition of EVPI can be obtained by considering influence diagram representations of decision processes.

Influence diagrams are a form of probabilistic graphical model (PGM), that augment Bayesian networks with decision and utility nodes. Within the influence diagram representation of a decision process, random variables are denoted by circular nodes, decisions are denoted by rectangular nodes, and utilities are denoted by rhombic nodes. Observed variables are denoted with shaded nodes. Edges from random variable/decision nodes to random variable/utility nodes denote the conditional dependence of the latter on the former. Edges from random variable/decision nodes to decision nodes are \textit{informational links} and prescribe an order in which observations and decisions are made. Optimal decisions may be found via an influence diagram by conducting probabilistic inference within the model and maximising expected utility. For comprehensive tutorials on PGMs, including algorithms for inference within Bayesian networks and influence diagrams, the reader is directed to \cite{Koller2009,Kjaerulff2008,Sucar2015}.

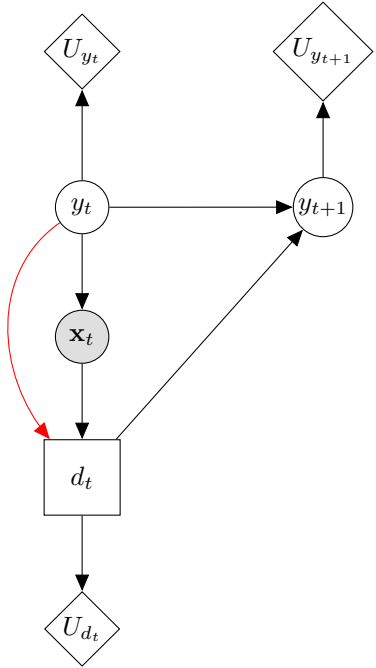
\begin{figure}[ht!]
	\centering
	\begin{tikzpicture}[x=1.7cm,y=1.8cm]

		\node[det] (uf1) {$U_{y_{t}}$} ;
		\node[det, right=2cm of uf1] (uf2) {$U_{y_{t+1}}$} ;
		\node[latent, below=1.2cm of uf1] (x1) {$y_{t}$} ;
		\node[latent, below=1cm of uf2] (x2) {$y_{t+1}$} ;
		\node[obs, below=1cm of x1] (y1) {$\mathbf{x}_{t}$} ;
		\node[rectangle,draw=black,minimum width=1cm,minimum height=1cm,below=1cm of y1] (d1) {$d_{t}$} ;
		\node[det, below=1cm of d1] (u1) {$U_{d_{t}}$} ;

		\edge {x1} {uf1} ; %
		\edge {x2} {uf2} ; %
		\edge {x1} {x2} ; %
		\edge {x1} {y1} ; %
		\edge {d1} {x2} ; %
		\edge {d1} {u1} ; %
		\edge {y1} {d1} ; %

		\draw [red, ->] (x1) to [out=-145,in=130] (d1);

	\end{tikzpicture}
	\caption{An influence diagram of a simplified SHM maintenance decision. The red edge denotes an informational link that implies inspection prior to maintenance and is present in $\mathcal{I}_{y_t \rightarrow d_t}$ and absent in $\mathcal{I}$. $U$ denotes a utility function and $d$ denotes a decision.}
	\label{fig:OverallPGM2}
\end{figure}

Given an influence diagram representation of a decision process $\mathcal{I}$ containing a decision $d$ and dependent on a variable $y$, the expected value of observing $y$ with perfect information prior to making decision $d$ is given by,

\begin{equation}\label{eq:EVPI}
	\text{EVPI}(d | y) := \text{MEU}(\mathcal{I}_{y \rightarrow d}) - \text{MEU}(\mathcal{I})
\end{equation}

\noindent
where $\mathcal{I}_{y \rightarrow d}$ is a modified influence diagram incorporating an additional informational link from $y$ to $d$ and MEU denotes the maximum expected utility. An example calculation of EVPI is presented in \cite{Hughes2022}.

An example influence diagram for a simple SHM maintenance decision process is shown in Figure \ref{fig:OverallPGM2}. In this example, there is a structural health state $y_t$ that evolves in time. The current health state is inferred from an observation of features $\mathbf{x}_t$ via a statistical classifier. The forecasted health state $y_{t+1}$ is conditionally dependent on $y_t$ and the action is decided for decision $d_t$. Finally, there are utility functions conditionally dependent on the health-states and actions. By applying equation (\ref{eq:EVPI}) to this decision process, the EVPI for a label $y_t$ can be determined. Here, it is worth recognising that value of information arises from uncertainty in classifier predictions $p(\tilde{y}_t = k|\tilde{\mathbf{x}}_t)$ between states warranting differing (optimal) courses of action.

\subsection{A Risk-based Active Learning Algorithm}

A convenient criterion for mandating structural inspections can be derived from the EVPI. Simply, if the EVPI of a label $y_t$ for a data point $\tilde{\mathbf{x}}_t$ exceeds the cost of making an inspection $C_{\text{ins}}$ then the ground-truth for $y_t$ should be obtained prior to $d_t$. Subsequently, the labelled dataset $\mathcal{D}_l$ can be extended to include the newly-acquired data-label pair $(\mathbf{x}_t,y_t)$, and the classifier retrained. While the assumption of perfect information may not hold in all cases, the principles and methodologies discussed in the current paper hold in general for value of information. Furthermore, the perfect information assumption may be relaxed via the introduction of an additional probabilistic model that quantifies uncertainty in inspections \cite{Papakonstantinou2014,Hamida2020}.

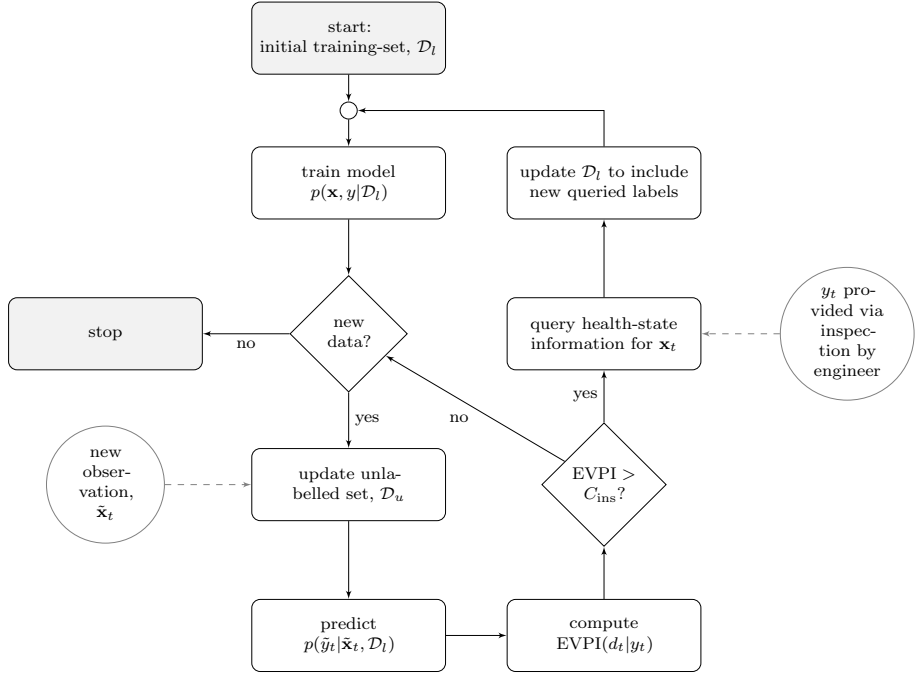
\begin{figure*}[pt]
	\centering
	\scalebox{0.8}{
		\begin{tikzpicture}[auto]
			\begin{footnotesize}
				\tikzstyle{decision} = [diamond, draw, text width=4em, text badly centered, inner sep=2pt]
				\tikzstyle{block} = [rectangle, draw, text width=10em, text centered, rounded corners, minimum height=4em]
				\tikzstyle{block2} = [rectangle, draw, text width=10em, text centered, rounded corners, minimum height=4em, fill=black!5]
				\tikzstyle{line} = [draw, -latex']
				\tikzstyle{cloud} = [draw=black!50, circle, node distance=3cm, minimum height=2em, text width=4em,text centered]
				\tikzstyle{point}=[draw, circle]
				\node [block2, node distance=4em] (start) {start:\\ initial training-set, $\mathcal{D}_l$};
				\node [point, below of=start, node distance=12mm] (point) {};
				\node [block, below of=point, node distance=12mm] (train) {train model\\ $p(\mathbf{x}, y | \mathcal{D}_l)$};
				\node [decision, below of=train, node distance=25mm] (new data) {new data?};
				\node [block2, left of=new data, node distance=40mm] (stop) {stop};
				\node [block, below of=new data, node distance=25mm] (update u) {update unlabelled set, $\mathcal{D}_u$};
				\node [cloud, left of=update u, node distance=40mm] (measured data) {new observation, $\tilde{\mathbf{x}}_t$};
				\node [block, below of=update u, node distance=25mm] (predict) {predict \\ $p(\tilde{y}_t | \tilde{\mathbf{x}}_t,\mathcal{D}_l)$};
				\node [block, right of=predict, node distance=42mm] (EVPI) {compute \\ $\text{EVPI}(d_t | y_t)$};
				\node [decision, above of=EVPI, node distance=25mm] (inspect) {$\text{EVPI} > C_{\text{ins}}$?};
				\node [block, above of=inspect, node distance=25mm] (query) {query health-state information for $\mathbf{x}_t$};
				\node [cloud, right of=query, node distance=40mm] (annotate) {$y_t$ provided via inspection by engineer};
				\node [block, above of=query, node distance=25mm] (update l) {update $\mathcal{D}_l$ to include new queried labels};
				\path [line] (start) -- (point);
				\path [line] (point) -- (train);
				\path [line] (train) -- (new data);
				\path [line] (new data) -- node {no}(stop);
				\path [line] (new data) -- node {yes}(update u);
				\path [line, dashed, draw=black!50] (measured data) -- (update u);
				\path [line] (update u) -- (predict);
				\path [line] (predict) -- (EVPI);
				\path [line] (EVPI) -- (inspect);
				\path [line] (inspect) -- node {yes}(query);
				\path [line] (inspect) -- node {no}(new data);
				\path [line, dashed, draw=black!50] (annotate) -- (query);
				\path [line] (query) -- (update l);
				\path [line] (update l) |- (point);
			\end{footnotesize}
		\end{tikzpicture}
	}
	\caption{Flow chart to illustrate the risk-based active learning process for classifier development and inspection scheduling.}
	\label{fig:RBAL}
\end{figure*}

A flow chart detailing the risk-based active learning process presented in \cite{Hughes2022}, is shown in Figure \ref{fig:RBAL}.

Adopting a risk-based approach to active learning allows one to learn statistical classifiers with consideration for the decision support contexts in which they may be employed. In \cite{Hughes2022}, it is demonstrated that this provides a cost-efficient manner for classifier development.

\section{Effects of Sampling Bias}
\label{sec:Effects}

Although active learning has been found to be an effective way of constructing highly informative and valuable datasets when labels are costly to obtain, such algorithms are somewhat of a double-edged sword because of a phenomenon known as \textit{sampling bias} \cite{Dasgupta2011,Dasgupta2008}. Sampling bias is an issue associated with machine learning in general; bias is introduced into models via training data when certain subpopulations have a higher or lower chance of being represented in said data. Specifically, sampling bias arises in active learning because of the preferential nature of the querying schemes. In active learning, data labels are obtained for specific regions of a feature space, according to the specified query measures and heuristics. Consequently, unrepresentative training datasets are formed in which the data diverges from the underlying generative distribution. In some applications, sampling bias can degrade the performance of classifiers \cite{Dasgupta2011,Dasgupta2008}. Risk-based approaches to active learning are more susceptible to the effects of sampling bias than traditional information-based approaches because data with high-value of information are often a subset of data with high-information content \cite{Bull2019,Hughes2022}. This characteristic can be understood if one realises that not all data are equally informative from a classification perspective, and further, that not all information is equally valuable from decision-making perspective.

Thus far, active learning algorithms for SHM have centred around generative classifiers \cite{Bull2019,Hughes2022}; learning joint probability distributions across the input and label space. Specifically, mixtures of Gaussian distributions, or Gaussian mixture models (GMMs), were utilised in \cite{Bull2019,Hughes2022}. To highlight the effects of sampling bias, a case study on a representative synthetic dataset is presented. In particular, the case study focusses on the performance of a decision-making agent utilising a risk-based active learning algorithm for the development of a GMM.

\subsection{Case Study: Visual Example}\label{sec:VisualExample}

To draw attention to the effects of sampling bias in risk-based active learning algorithms, a modified version of the synthetic dataset used in \cite{Hughes2022} is adopted. The dataset consists of a two-dimensional input space $\mathbf{x}_t = \{ x_t^{1} , x_t^{2} \}$ and a four-class label space $y_t \in \{ 1,2,3,4 \}$. Framed as an SHM dataset, one can consider the inputs as features used to discriminate between classes corresponding to four health states of interest for a structure $S$ where:

\begin{itemize}
	\item Class 1 corresponds to the structure being undamaged and fully functional.
	\item Class 2 corresponds to the structure possessing minor damage whilst remaining fully functional.
	\item Class 3 corresponds to the structure possessing significant damage, resulting in a reduced operational capacity.
	\item Class 4 corresponds to the structure possessing critical damage, resulting in operational incapacity, i.e.\ total structural failure.
\end{itemize}

\begin{figure}[ht!]
	\begin{subfigure}{.5\textwidth}
		\centering
		\scalebox{0.4}{
			\includegraphics{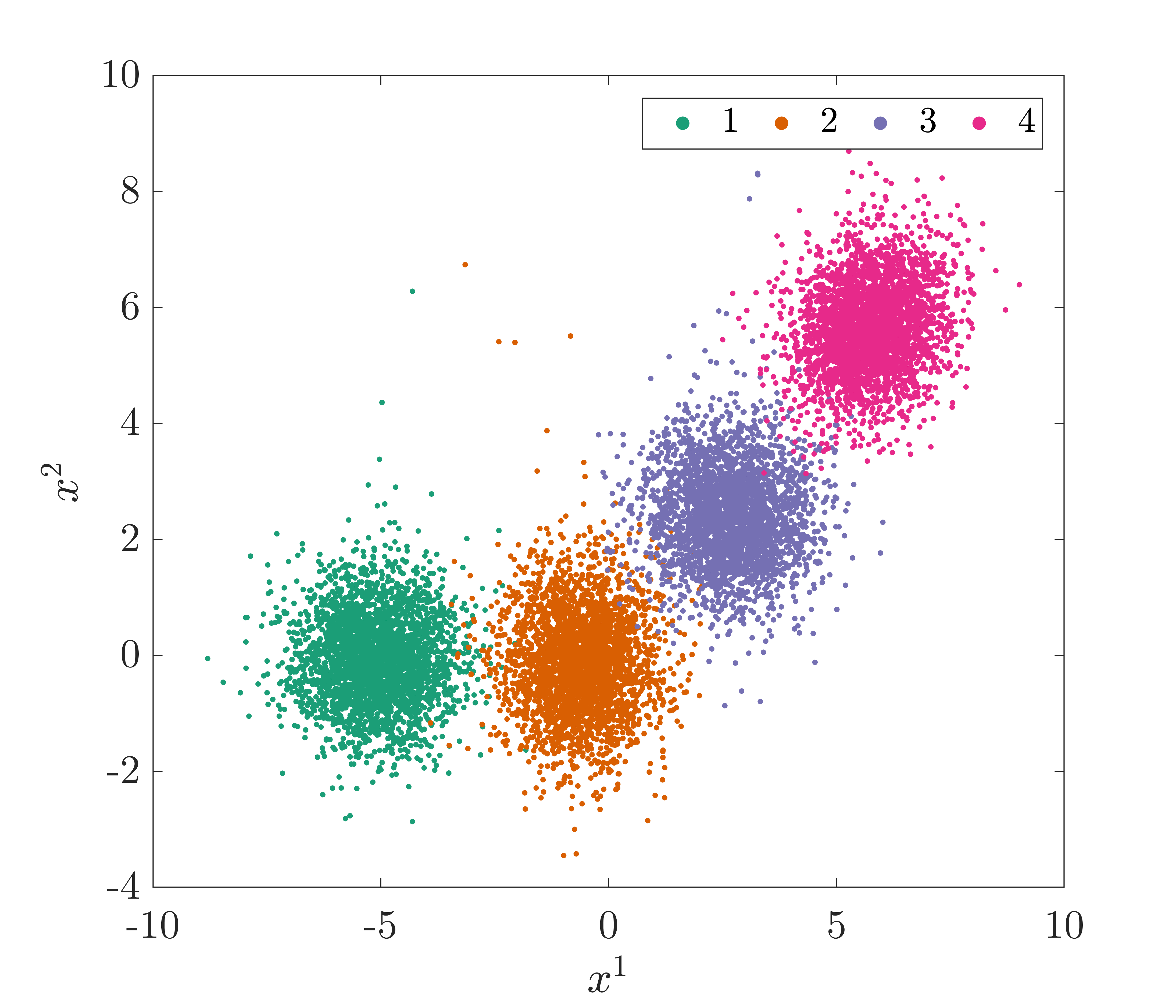}
		}
		\caption{ }
		\label{fig:scatter_ext}
	\end{subfigure}
	\begin{subfigure}{.5\textwidth}
		\centering
		\scalebox{0.4}{
			\includegraphics{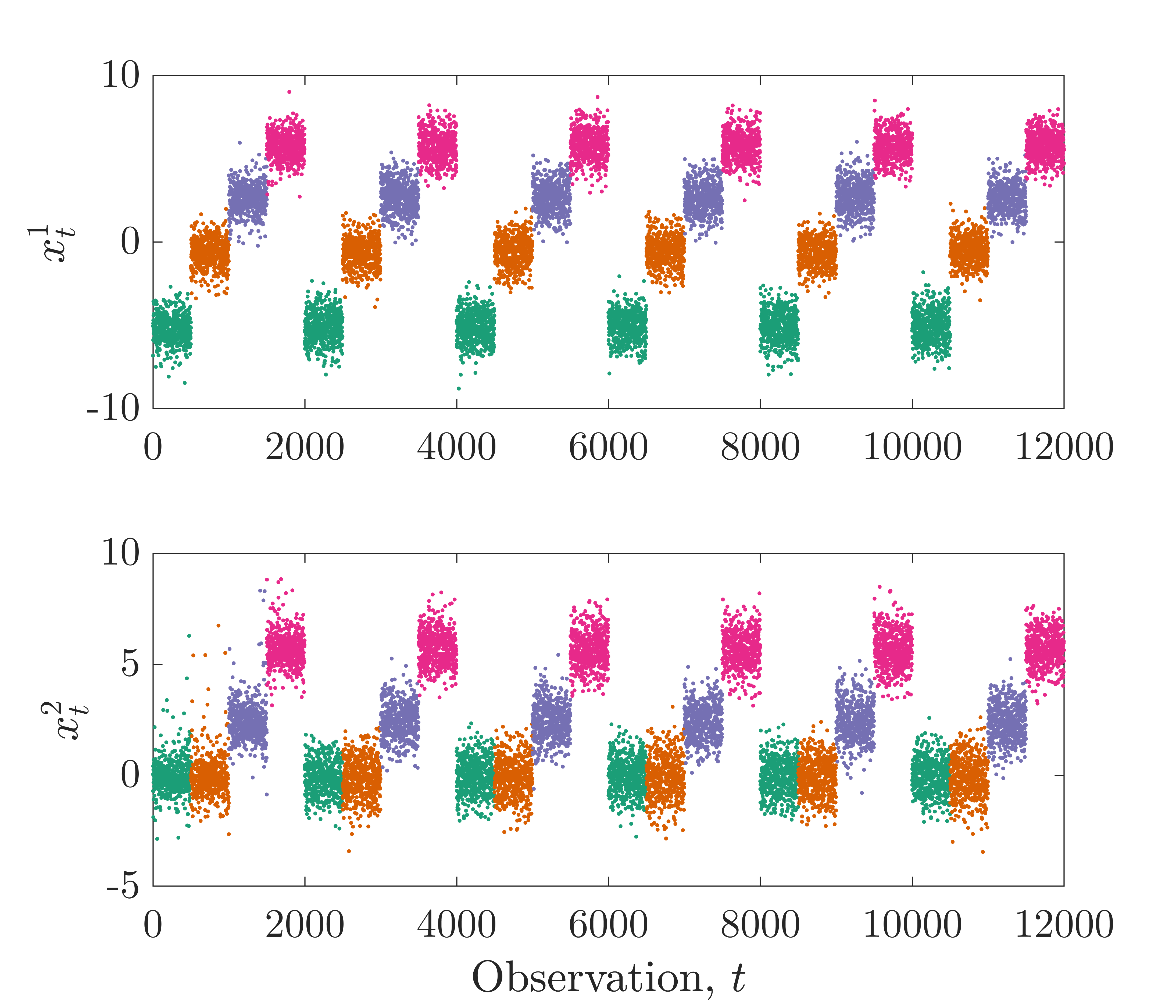}
		}
		\caption{ }
		\label{fig:overview_ext}
	\end{subfigure}
	\caption{Visualisation of the extended synthetic dataset in (a) the feature space and (b) discrete time $t$.}
	\label{fig:data_ext}
\end{figure}

The original dataset comprised of 1997 data points ordered according to health state -- from Class 1 through to Class 4. For the current case study, the dataset was extended by drawing additional independent samples from a generative distribution learned from the original dataset. This extension procedure was conducted a total of five times resulting in 11997 data points -- the original dataset contributing 1997 points, and each of the five extensions contributing an additional 2000 points. The data generated progresses repeatedly from Class 1 to Class 4 a total six times, thereby emulating the process of structural degradation and subsequent repair -- a pattern that could conceivably be experienced by the fictitious structure of interest $S$. A visualisation of the extended dataset is presented in Figure \ref{fig:data_ext}.

\subsubsection{Decision Process}\label{sec:DecProc1}

In order to employ risk-based active learning for the development of a statistical classifier, a decision process for the structure $S$ must first be specified. For consistency with \cite{Hughes2022}, an identical binary maintenance decision process is selected for the current study.

\begin{figure}[ht!]
	\centering
	\begin{tikzpicture}[x=1.7cm,y=1.8cm]
		
		\node[det] (uf2) {$U_{y_{t+1}}$} ;
		\node[latent, below=1cm of uf2] (x2) {$y_{t+1}$} ;
		\node[latent, left=2cm of x2] (x1) {$y_{t}$} ;
		\node[obs, below=1cm of x1] (y1) {$\mathbf{x}_{t}$} ;
		\node[rectangle,draw=black,minimum width=1cm,minimum height=1cm,below=1cm of y1] (d1) {$d_{t}$} ;
		\node[det, below=1cm of d1] (u1) {$U_{d_{t}}$} ;

		\edge {x2} {uf2} ; %
		\edge {x1} {x2} ; %
		\edge {x1} {y1} ; %
		\edge {d1} {x2} ; %
		\edge {d1} {u1} ; %
		\edge {y1} {d1} ; %

	\end{tikzpicture}
	\caption{An influence diagram representation of the decision process associated with structure $S$.}
	\label{fig:ID_ext}
\end{figure}
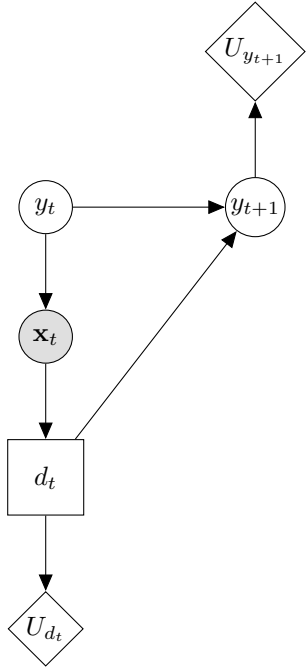

Consider an agent that, at discrete-time instances $t \in \mathbb{N}$, is tasked with making a binary decision $d_t$. The agent must select an action from $\text{dom}(d_t) = \{ 0 \text{ (do nothing)} , 1 \text{ (repair)} \}$, such that some degree of operational capacity is maintained for $S$ at instance $t+1$. It is assumed that the agent has access to the observed discriminative features $\mathbf{x}_t$ and must infer the current and future latent health states $y_t$ and $y_{t+1}$ via a statistical classifier and a transition model, respectively. An influence diagram for such a decision process is shown in Figure \ref{fig:ID_ext}.

Figure \ref{fig:ID_ext} indicates that the actions in the domain of $d_t$ have costs associated, specified by the utility function $U(d_t)$, and shown in Table \ref{tab:Ud}. It is assumed that $d_t = 0$ has no utility associated with it. On the other hand, $d_t = 1$ has negative utility because of the expenditure necessary to cover the material and labour costs associated with structural maintenance. It is worth noting that, in many practical applications, the specification of utility function is non-trivial and is an active topic of research outside the scope of the current paper. For further reading related to the specification of utility functions, the reader is directed to \cite{Valkonen2021}. Hence, for this case study, relative utility values are selected to be only somewhat representative of the SHM context.

For simplicity, and for more compact influence diagrams, within the current case study it is assumed that health states may be mapped directly to utilities. For information regarding how this may be achieved in a principled manner, via the specification of failure events as Bayesian network representations of fault trees, the reader is directed to \cite{Hughes2021}. Once again, the utility function used here reflects the relative utility values that may be expected in a typical SHM application. The utility function $U(y_{t+1})$ is provided in Table \ref{tab:Uy}. It can be seen from Table \ref{tab:Uy} that for health-states 1 and 2, in which the structure is at full operational capacity, positive utility has been assigned. For health-state 3, which corresponds to a reduced operational capacity, a lesser positive utility has been assigned. For health-state 4, which corresponds to critical damage,  a relatively-large negative utility has been assigned, to reflect a complete loss of operational capacity and an additional severe consequence associated with structural failure, e.g.\ environmental damage.

\begin{table}
	\begin{minipage}{.5\linewidth}
		\centering
		\caption{The utility function $U(d_t)$ where $d_t=0$ and $d_t=1$ denote the `do nothing' and `repair' actions, respectively.}
	  \label{tab:Ud}   
	  \begin{tabular}{cc}
		  \toprule
		  \midrule
		  $d_t$ & $U(d_t)$\\
		  \midrule
		  $0$ & $0$\\
		  $1$ & $-30$\\
		  \midrule
		  \bottomrule
	  \end{tabular}
	  \end{minipage}
	\begin{minipage}{.5\linewidth}
	  \centering
	  \caption{The utility function $U(y_{t+1})$.}
	\label{tab:Uy}   
	\begin{tabular}{cc}
		\toprule
		\midrule
		$y_{t+1}$ & $U(y_{t+1})$\\
		\midrule
		$1$ & $10$\\
		$2$ & $10$\\
		$3$ & $5$\\
		$4$ & $-75$\\
		\midrule
		\bottomrule
	\end{tabular}
	\end{minipage}%
  \end{table}

As aforementioned, the agent requires a transition model in order to obtain a forecast for the future health state $y_{t+1}$. For the current case study, it is assumed that this model is known \textit{a priori} so that focus can be given to the development of the statistical classifier. Examples of this transition model being developed from data and prior knowledge of physics are presented in \cite{Vega2020a} and \cite{Hughes2022b}, respectively. As shown in Figure \ref{fig:ID_ext}, the future health state $y_{t+1}$ is conditionally dependent on the current health state $y_{t}$ and the decision $d_t$. Given $d_t = 0$, it is assumed that the structure will monotonically degrade with a propensity to remain in its current health state. This assumption is reflected in the conditional probability distribution $P(y_{t+1}|y_t,d_t=0)$ presented in Table \ref{tab:P_y1_y0_d0}. Given $d_t = 1$, it is assumed that the structure is returned to its undamaged state with probability 0.99 and remains in its current state with probability 0.01. This assumption is reflected in the conditional probability distribution $P(y_{t+1}|y_t,d_t=1)$ presented in Table \ref{tab:P_y1_y0_d1}.

\begin{table}[ht!]
	\centering
	\caption{The conditional probability table $P(y_{t+1}|y_t, d_t)$ for $d_t = 0$.}
	\label{tab:P_y1_y0_d0}       
\begin{tabular}{c c c c c c}
	\toprule
	\midrule
	& & \multicolumn{4}{c}{$y_{t+1}$}\\
	&& 1  & 2 & 3 & 4  \\ \cmidrule{3-6}
	\multicolumn{1}{c}{\multirow{4}{*}{\begin{sideways}\parbox{1.5cm}{\centering $y_t$}\end{sideways}}}   &
	\multicolumn{1}{l}{1}& 0.8 & 0.18 & 0.015 & 0.005 \\
	\multicolumn{1}{c}{}    &
	\multicolumn{1}{l}{2}& 0 & 0.8 & 0.15 & 0.05  \\
	\multicolumn{1}{c}{}    &
	\multicolumn{1}{l}{3} & 0 & 0 & 0.8 & 0.2  \\
	\multicolumn{1}{c}{}    &   
	\multicolumn{1}{l}{4} & 0 & 0 & 0 & 1  \\
	\midrule
	\bottomrule
\end{tabular}
\end{table}

\begin{table}[ht!]
	\centering
	\caption{The conditional probability table $P(y_{t+1}|y_t, d_t)$ for $d_t = 1$.}
	\label{tab:P_y1_y0_d1}       
	\begin{tabular}{c c c c c c}
		\toprule
		\midrule
		& & \multicolumn{4}{c}{$y_{t+1}$}\\
		&& 1  & 2 & 3 & 4  \\ \cmidrule{3-6}
		\multicolumn{1}{c}{\multirow{4}{*}{\begin{sideways}\parbox{1.5cm}{\centering $y_t$}\end{sideways}}}   &
		\multicolumn{1}{l}{1}& 1 & 0 & 0 & 0 \\
		\multicolumn{1}{c}{}    &
		\multicolumn{1}{l}{2}& 0.99 & 0.01 & 0 & 0  \\
		\multicolumn{1}{c}{}    &
		\multicolumn{1}{l}{3} & 0.99 & 0 & 0.01 & 0  \\
		\multicolumn{1}{c}{}    &   
		\multicolumn{1}{l}{4} & 0.99 & 0 & 0 & 0.01  \\
		\midrule
		\bottomrule
	\end{tabular}
\end{table}

As alluded to previously, the current health state $y_t$ is inferred from the observed features $\mathbf{x}_t$ via a statistical classifier subject to risk-based active learning. To fully specify the contextual parameters for risk-based active learning, it is assumed that the ground-truth health state at discrete-time instance $t$ can be obtained via inspection at the cost of $C_{\text{ins}} = 7$. Here, cost can be interpreted as the magnitude of a strictly non-positive utility.

\subsubsection{Statistical Classifier}\label{sec:GMM}

As previously discussed, generative classifiers, specifically GMMs, have garnered the vast proportion of attention with regard to active learning for SHM \cite{Bull2019,Hughes2022}. As such, the current case study adopts a classifier specified by a mixture of four multivariate Gaussian distributions learned in a supervised manner from a labelled dataset $\mathcal{D}_l$. Each Gaussian component, specifies a generative model for the observable features $\mathbf{x}_t$ and is conditionally dependent on one of the four possible health states in $\text{dom}(y_t)$,

\begin{equation}
	p(\mathbf{x}_t|y_t = k) = \mathcal{N}(\bm{\mu}_k,\Sigma_k)
\end{equation}

\noindent
 for $k = 1,\ldots,4$ and where $\bm{\mu}_k$ and $\Sigma_k$ are the parameters of the multivariate Gaussian distribution corresponding to the mean and covariance, respectively. Further details of the GMM, including specification of the \textit{mixing proportions} $\mathbf{\lambda}$ and steps required for Bayesian inference of the distribution parameters, are provided in Appendix \ref{app:GMMs}.

Concisely, a Gaussian mixture model was trained in a supervised Bayesian manner on $\mathcal{D}_l$. Within the risk-based active learning algorithm, the learning procedure presented in Appendix \ref{app:GMMs} is reapplied each time the labelled dataset $\mathcal{D}_l$ is extended following the inspection of structure $S$ that is mandated when $\text{EVPI}(d_t|y_t) > C_{ins}$.

\subsubsection{Results}\label{sec:Results1}

In order to assess the effects of sampling bias, the risk-based active learning approach was used to learn a GMM within the decision process outlined in Section \ref{sec:DecProc1}. This process was repeated 1000 times. For each repetition, the dataset was randomly halved into a test set and a training set $\mathcal{D}$. From the training set, a small ($\sim$0.2\%) random subset retain their corresponding ground-truth labels. These data form the initialised labelled dataset $\mathcal{D}_l$. The remaining majority of data from $\mathcal{D}$ have their ground-truth labels hidden, forming the unlabelled dataset $\mathcal{D}_u$.

\begin{figure}[ht!]
	\begin{subfigure}{.5\textwidth}
		\centering
		\scalebox{0.4}{
			\includegraphics{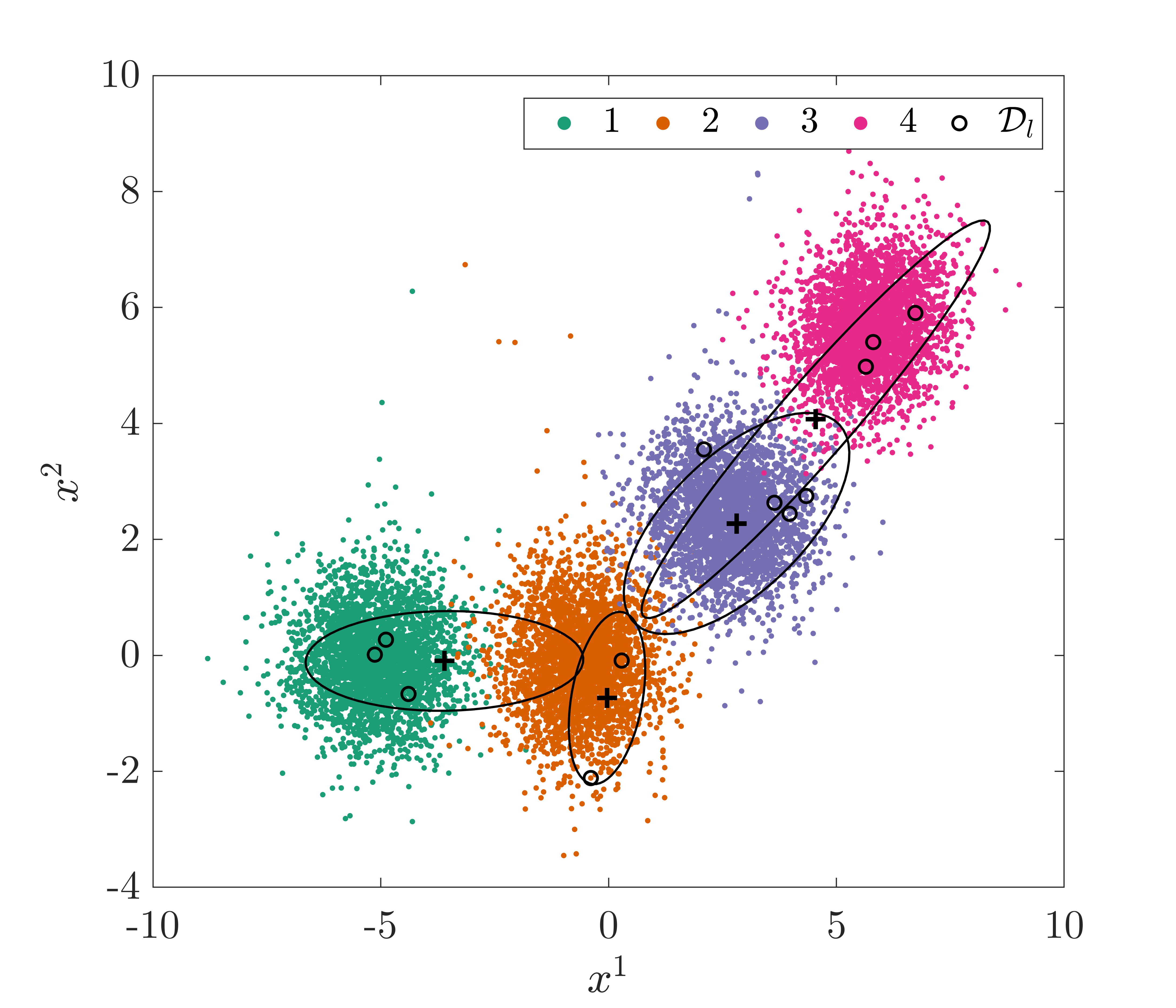}
		}
		\caption{ }
		\label{fig:initialScatter}
	\end{subfigure}
	\begin{subfigure}{.5\textwidth}
		\centering
		\scalebox{0.4}{
			\includegraphics{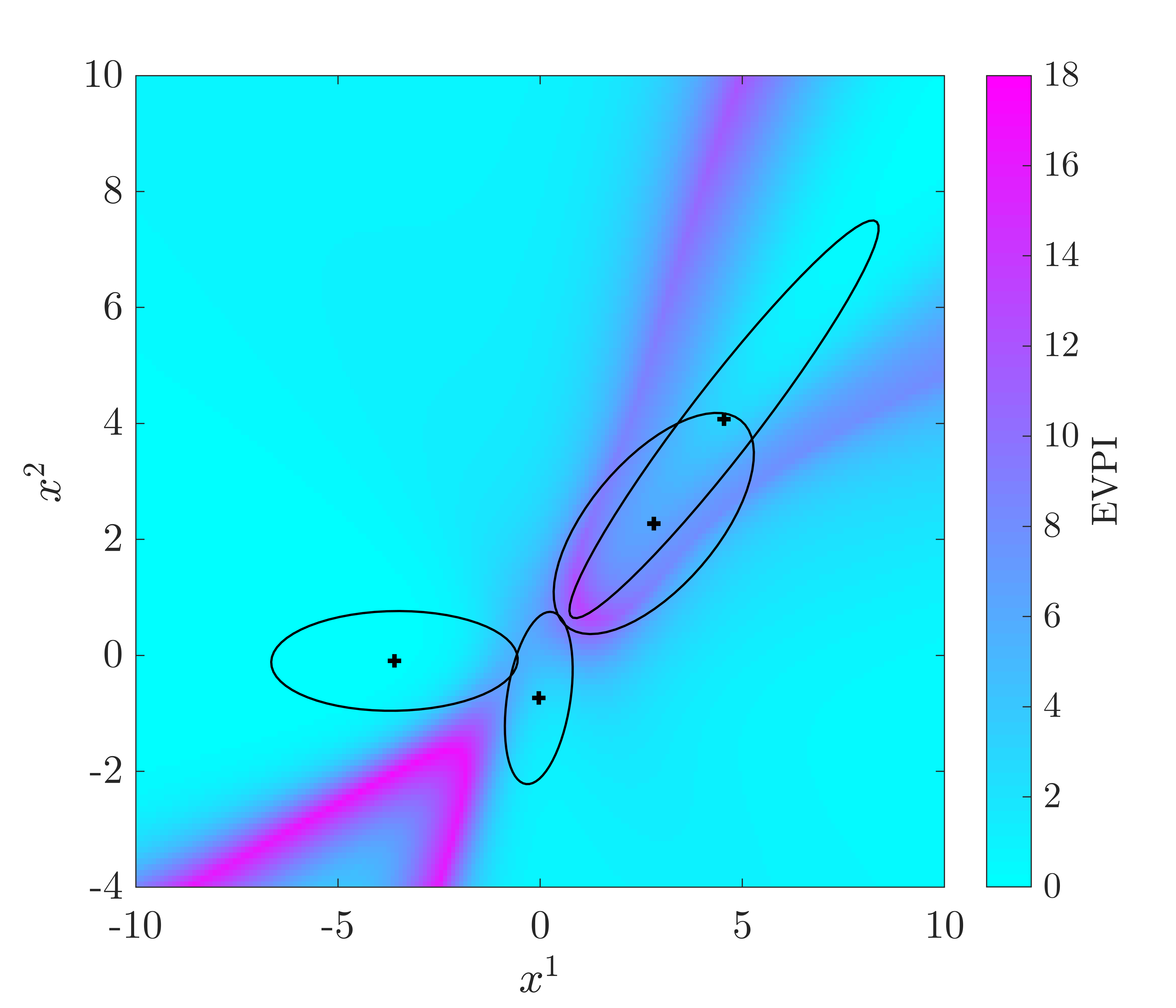}
		}
		\caption{ }
		\label{fig:initialVOI}
	\end{subfigure}
	\caption{A statistical classifier $p(y_t,\mathbf{x}_t,\bm{\Theta})$ prior to risk-based active learning; \textit{maximum a posteriori} (MAP) estimates of the mean (+) and covariance (ellipses represent 2$\sigma$) are shown. (a) shows the initial model overlaid onto the data with labelled data $\mathcal{D}_l$ encircled and (b) shows the resulting EVPI over the feature space.}
	\label{fig:initialModel}
\end{figure}

\begin{figure}[ht!]
	\begin{subfigure}{.5\textwidth}
		\centering
		\scalebox{0.4}{
			\includegraphics{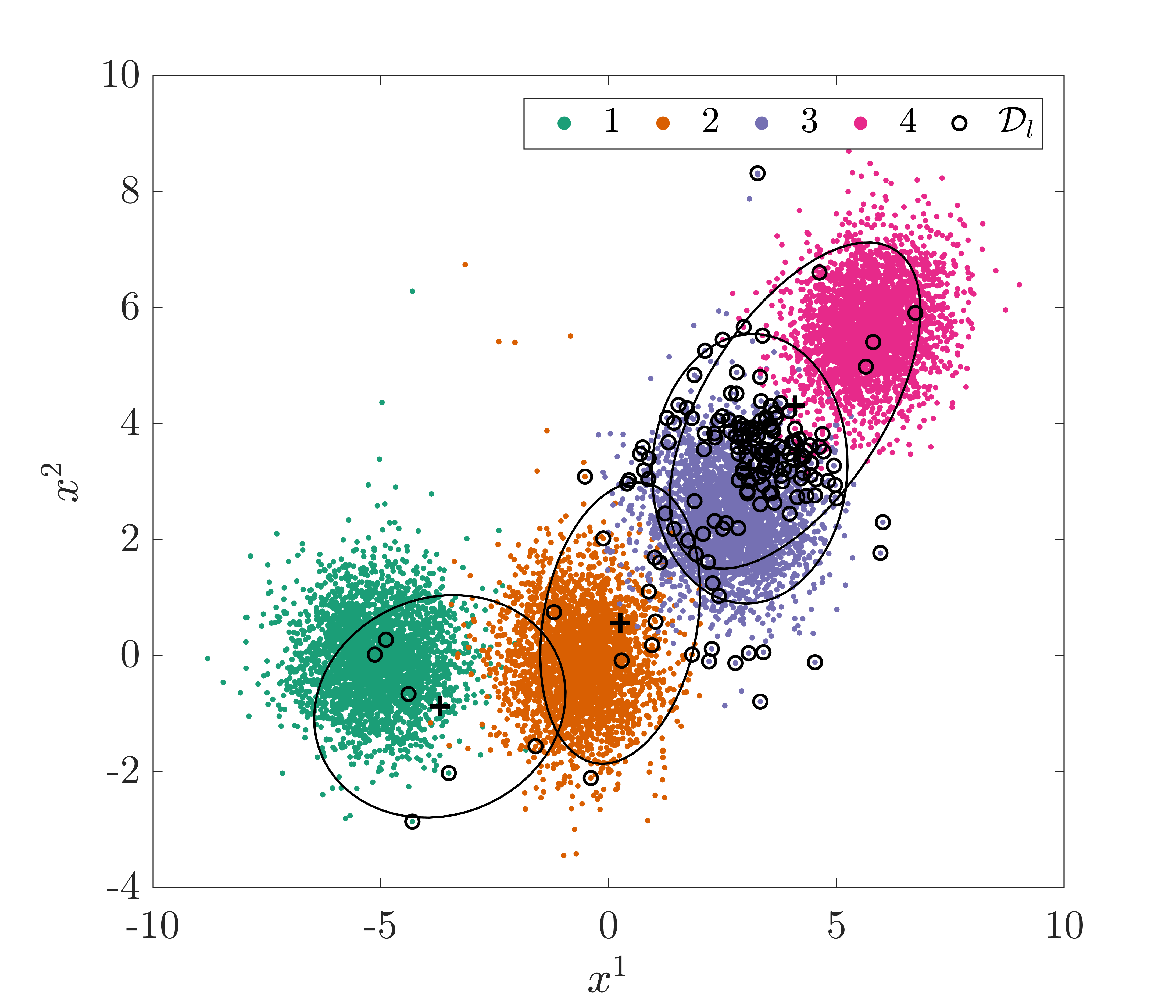}
		}
		\caption{ }
		\label{fig:finalScatter}
	\end{subfigure}
	\begin{subfigure}{.5\textwidth}
		\centering
		\scalebox{0.4}{
			\includegraphics{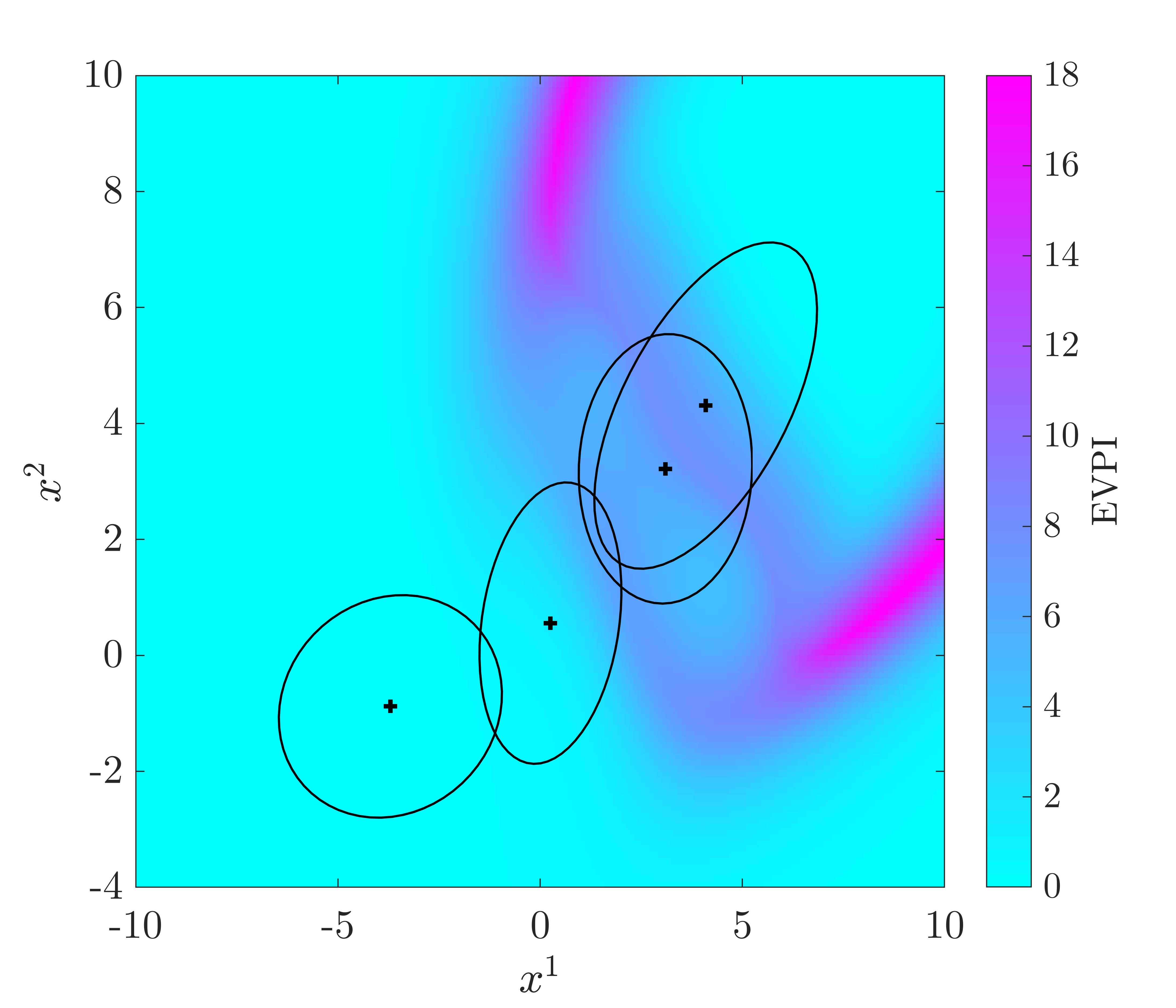}
		}
		\caption{ }
		\label{fig:finalVOI}
	\end{subfigure}
	\caption{A statistical classifier $p(y_t,\mathbf{x}_t,\bm{\Theta})$ following risk-based active learning; \textit{maximum a posteriori} (MAP) estimates of the mean (+) and covariance (ellipses represent 2$\sigma$) are shown. (a) shows the final model overlaid onto the data with labelled data $\mathcal{D}_l$ encircled and (b) shows the resulting EVPI over the feature space.}
	\label{fig:finalModel}
\end{figure}

Figures \ref{fig:initialModel} and \ref{fig:finalModel} show a GMM from one of the 1000 repetitions, before and after the risk-based active learning process, respectively.

It can be seen from Figure \ref{fig:initialScatter} that, initially, the model fits poorly; the means and covariances for each class being heavily influenced by the zero-mean and unit-variance priors because of the lack of data. Figure \ref{fig:initialVOI} shows the EVPI over the feature space induced by considering the initial model within the context of the decision process. The near-symmetric `sharp' regions of high EVPI (pink) can be attributed to the cluster for Class 4; in particular, the major axis of the covariance ellipse (i.e.\ the dominant eigenpair of the covariance matrix). Intuitively, the regions of high value of information occur between the clusters for classes with severe and milder consequences.

From Figure \ref{fig:finalScatter}, one can see that during the active learning process, querying is concentrated in highly localised regions of the feature space. Specifically, regions between the clusters for Class 3 (moderate damage) and Class 4 (severe damage), have been queried preferentially. It is clear from Figure \ref{fig:finalScatter} that the subset of data $\mathcal{D}_l$ is not representative of the underlying distribution, indicating that sampling bias is present. Nonetheless, the queried data have been somewhat successful in learning a decision boundary - as can be deduced by observing the region of high EVPI between the means of clusters for Class 3 and Class 4 in Figure \ref{fig:finalVOI}.

\begin{figure}[ht!]
	\begin{subfigure}{.5\textwidth}
		\centering
		\scalebox{0.4}{
			\includegraphics{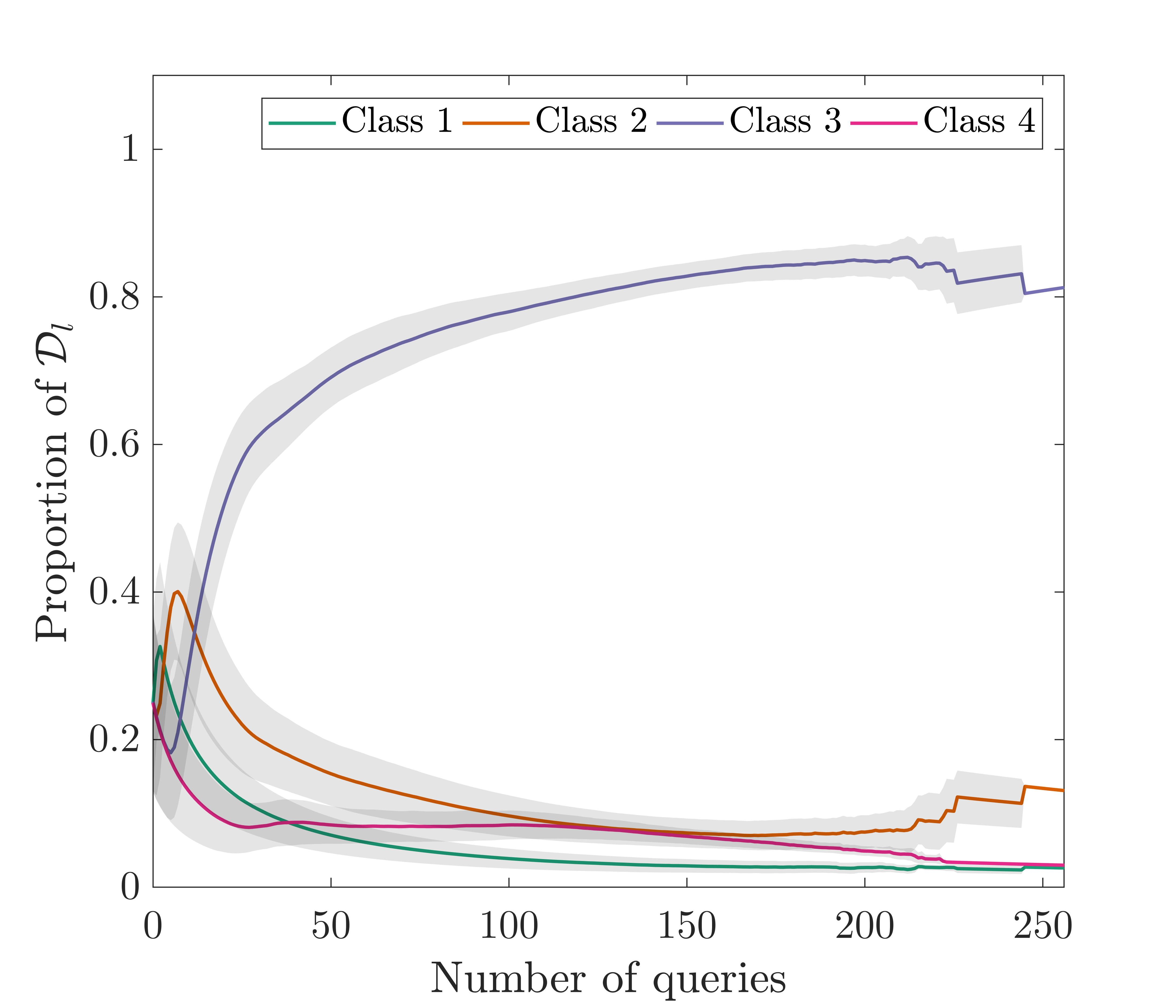}
		}
		\caption{ }
		\label{fig:cpropral_ext}
	\end{subfigure}
	\begin{subfigure}{.5\textwidth}
		\centering
		\scalebox{0.4}{
			\includegraphics{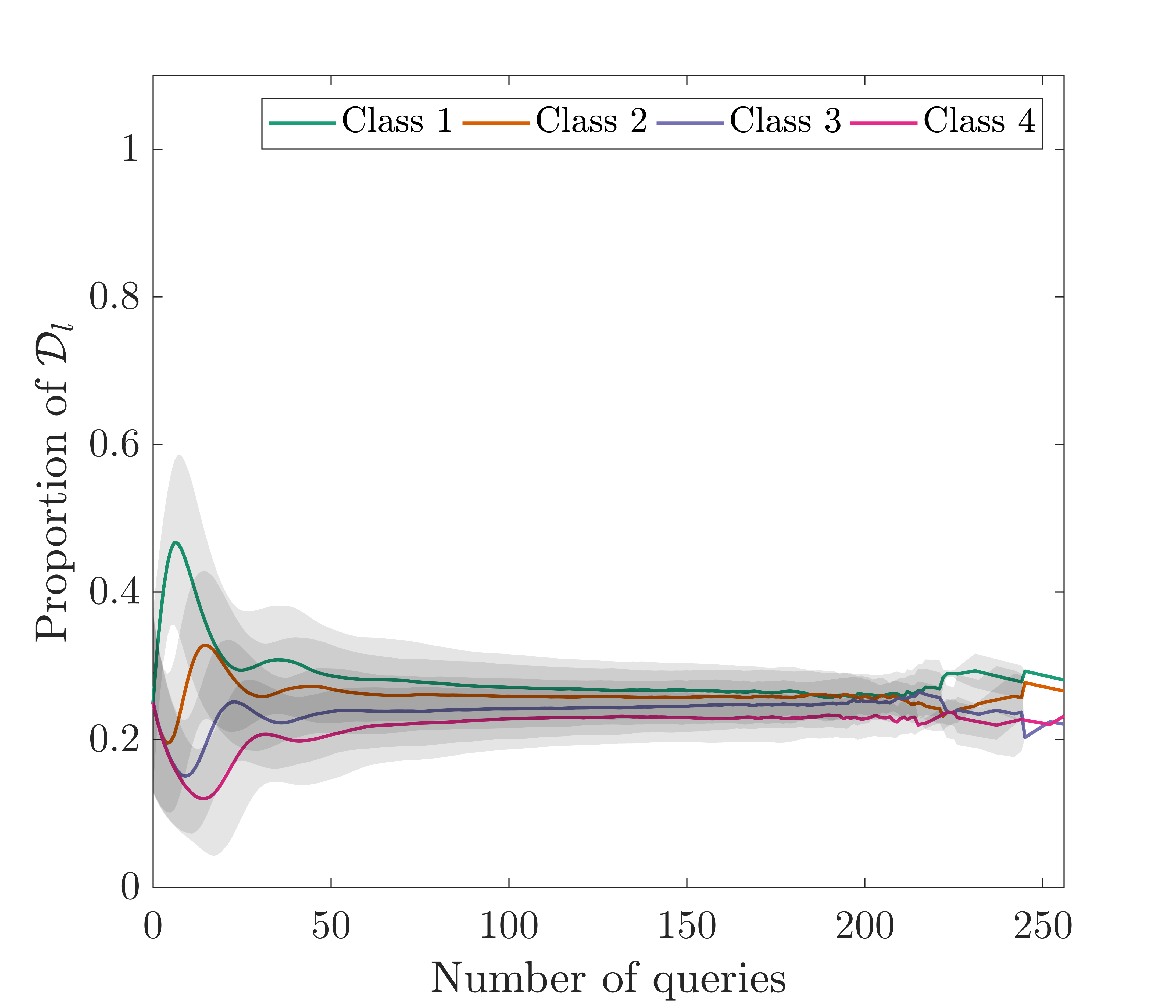}
		}
		\caption{ }
		\label{fig:cproprand_ext}
	\end{subfigure}
	\caption{Variation in class proportions within $\mathcal{D}_l$ with number of label queries for an agent utilising a GMM learned from $\mathcal{D}_l$ extended via (a) risk-based active querying and (b) random sampling. Shaded area shows $\pm1\sigma$.}
	\label{fig:cprop_ext}
\end{figure}

\begin{figure}[ht!]
	\centering
		\scalebox{0.4}{
			\includegraphics{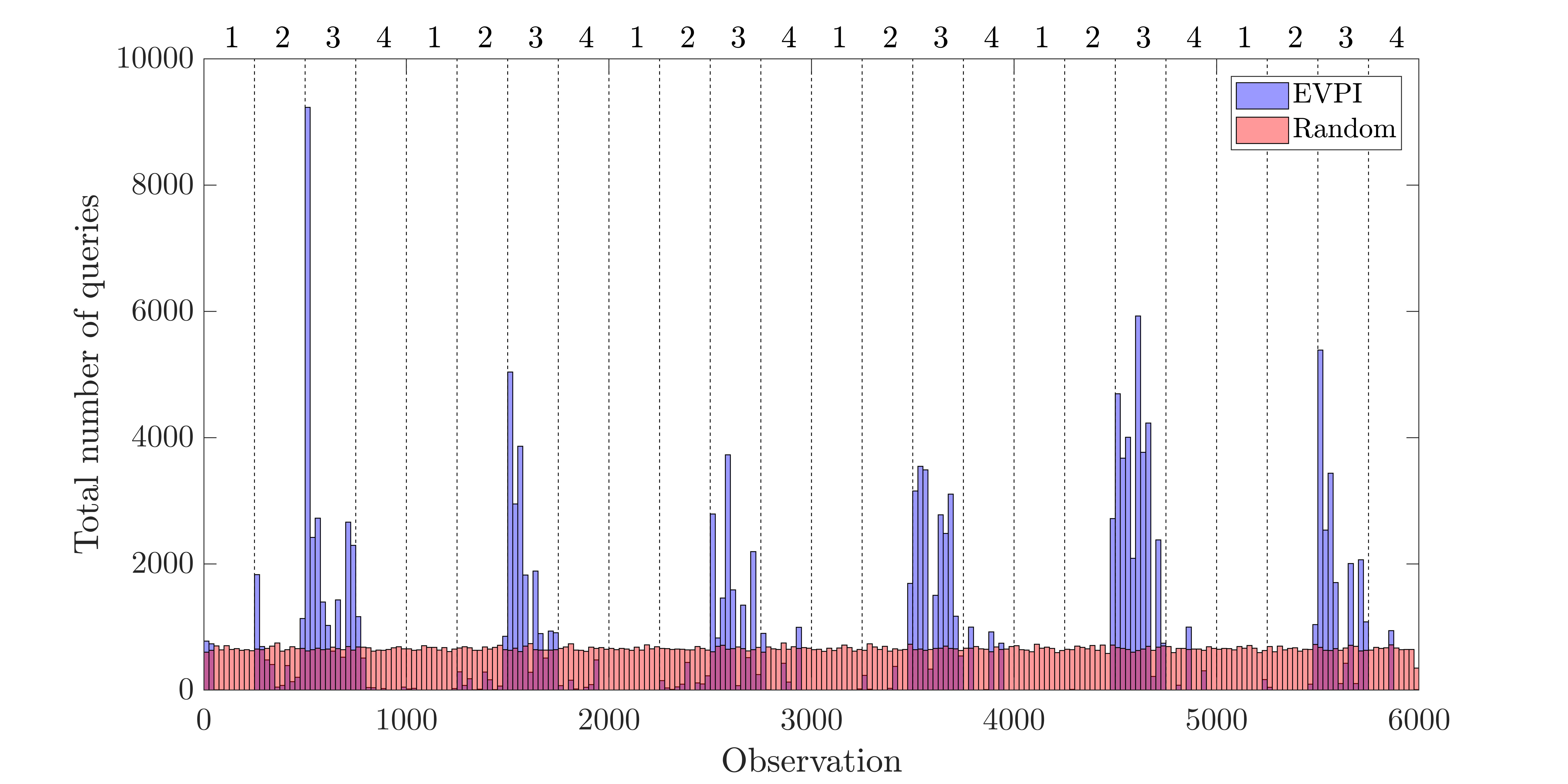}
		}
		\caption{Histograms showing the distribution of the number of queries for each observation in $\mathcal{D}_u$ over 1000 runs when adopting (a) risk-based active learning (EVPI) and (b) random sampling (Random) in order to learn a GMM. The average location of classes within $\mathcal{D}_u$ are numbered on the upper horizontal axis and transitions between classes are denoted as a dashed line.}
		\label{fig:all_queries_ext}
\end{figure}

Figures \ref{fig:cprop_ext} and \ref{fig:all_queries_ext} provide further evidence of sampling bias in the risk-based active learning approach.

Figure \ref{fig:cprop_ext} shows the means and standard deviations, calculated from the 1000 repetitions, of the relative representation of each class within $\mathcal{D}_l$. Figure \ref{fig:cpropral_ext} shows the class proportions for $\mathcal{D}_l$ subject to risk-based active learning whereas Figure \ref{fig:cproprand_ext} shows the class proportions for $\mathcal{D}_l$ subject to an equivalent number of random queries. As one would expect, the random querying is initially biased, as the learning algorithm is presented with ordered data. However, as more random queries are made, the proportion for each class converges to approximately 25\%. In stark contrast, under querying guided according to EVPI, the class proportions diverge early in the querying process. Figure \ref{fig:cpropral_ext} shows that, on average, Class 3 garners a majority representation after fewer than 50 queries. This disparity in class representation is maintained throughout the risk-based active learning process with class 3 reaching a peak representation of approximately 80\%. 

Figure \ref{fig:all_queries_ext} compares the total number of queries for data point indices in $D_u$ over the 1000 repetitions for EVPI-based and random querying. Although individual data points within the training data $D$ are not fixed within $D_u$ because of the random selection of $D_l$, only a small proportion of data are in $D_l$ and therefore the index of a datapoint will differ by no more than 12 between $D$ and $D_u$. Furthermore, the histogram `bins' represent groups containing 25 indices each, thereby minimising the amount of leakage from differing indices. As one would expect, random sampling results in each data point being queried approximately by an equal amount. As is indicated by the class proportions shown in Figure \ref{fig:cprop_ext}, Figure \ref{fig:all_queries_ext} shows risk-based active learning results in data corresponding to Class 3 being queried preferentially which, of course, results in a biased labelled dataset on which the classifier is trained.

\begin{figure}[ht!]
	\begin{subfigure}{.5\textwidth}
		\centering
		\scalebox{0.4}{
			\includegraphics{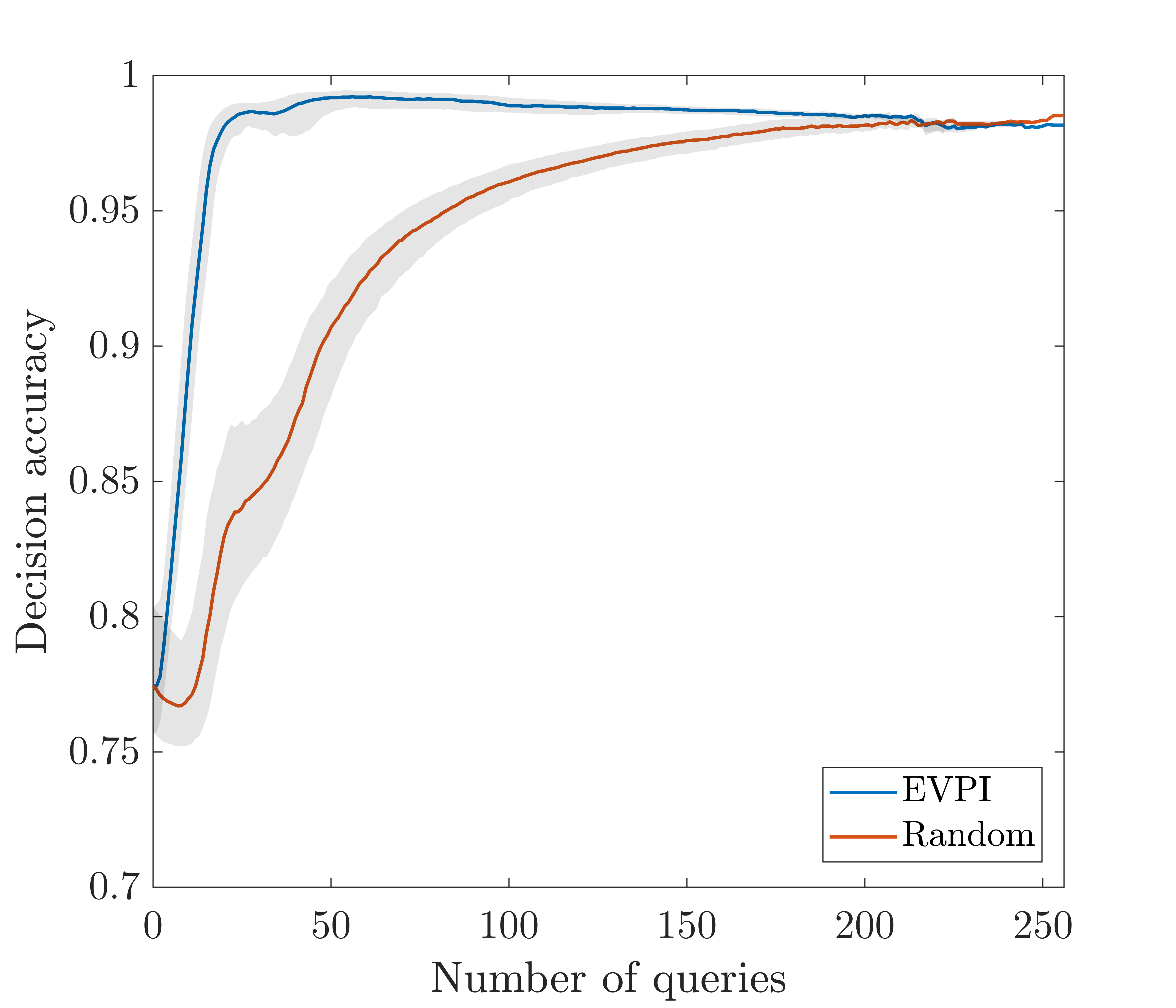}
		}
		\caption{ }
		\label{fig:dacc_ext}
	\end{subfigure}
	\begin{subfigure}{.5\textwidth}
		\centering
		\scalebox{0.4}{
			\includegraphics{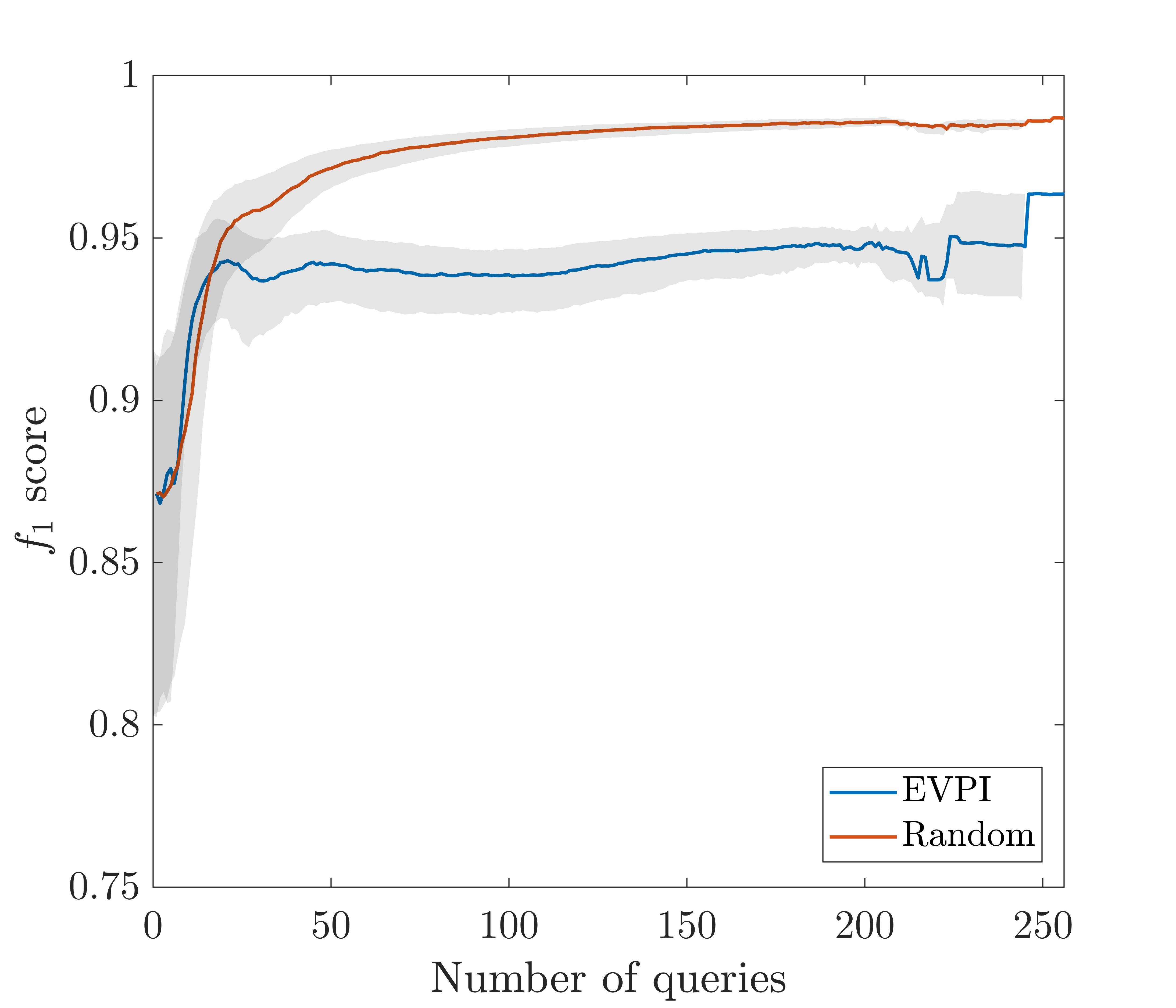}
		}
		\caption{ }
		\label{fig:f1score_ext}
	\end{subfigure}
	\caption{Variation in median (a) decision accuracy and (b) $f_{1}$ score with number of label queries for an agent utilising a GMM learned from $\mathcal{D}_l$ extended via (i) risk-based active querying (EVPI) and (ii) random sampling (Random). Shaded area shows the interquartile range.}
	\label{fig:performance_ext}
\end{figure}

Both the positive and negative consequences of sampling bias can be realised by considering Figure \ref{fig:performance_ext}. Figure \ref{fig:dacc_ext} shows the median and interquartile range (calculated from 1000 repetitions) of the decision accuracy for an agent utilising the learned GMMs, evaluated throughout the risk-based active-learning process. For comparison, also shown in Figure \ref{fig:dacc_ext}, is the same performance measure for an agent utilising GMMs learned from a labelled dataset comprised of an equal number of randomly-queried data points. Here, `decision accuracy' is a measure of decision-making performance \cite{Hughes2021}. The decision accuracy is a comparison between the actions selected by an agent using the classifier being evaluated, and the optimal (correct) actions selected by an agent in possession of perfect information. Quantitatively, decision accuracy is defined as,

\begin{equation}
	\text{decision accuracy} = \frac{\text{number of correct decisions}}{\text{total number of decisions}}
\end{equation}

It can be seen from Figure \ref{fig:dacc_ext}, that a high level of decision-making performance can be achieved with fewer queries by adopting a risk-based approach to active learning, when compared to random sampling. This result is achieved because of the preferential querying process that induces sampling bias in the labelled dataset. That being said, the performance of the agent utilising risk-based active learning gradually degrades as more queries are made. While randomly obtaining queries improves decision accuracy at a lesser rate, a decline in performance is not observed and, in fact, random sampling eventually surpasses guided querying in terms of decision-making performance. These results indicate that the decline in decision accuracy is a result of the sampling bias that is exacerbated by the latter queries in the active learning process. This proposition is further supported by Figure \ref{fig:f1score_ext}.

Figure \ref{fig:f1score_ext} shows the average classification performances ($f_{1}$-score \cite{Fawcett2006}) of agents utilising EVPI-guided and random approaches to querying. It can be seen from Figure \ref{fig:f1score_ext}, that active querying and random querying increase classification performance at similar rates initially. However, the classification performance for risk-based active learning plateaus at a lower value than that of random sampling. This observation provides supplementary evidence that classifiers developed via risk-based active learning represent the underlying distribution poorly\footnote{In the limit of infinite queries, random sampling will achieve classification performance equivalent to the `gold standard' of fully-supervised learning.}. Once again, this observation is to be expected when considering the disparity in class representation within $\mathcal{D}_l$ for risk-based active learning. The drastic difference between the decision accuracy and classification accuracy demonstrates that risk-based active learning algorithms prioritise decision-making performance over classification performance.

\subsubsection{Further Comments}\label{sec:FurtherComments1}

As an archetypal example of a probabilistic generative model, one can rationalise the effects of sampling bias on decision-making agents utilising such models by examining the posterior parameter estimates for the GMM, given in equations \ref{eq:studentT} and \ref{eq:mixpropupdate}. Here, a reminder is provided that, as a result of the Bayesian inference on distribution parameters, components of the posterior GMM are Student-\textit{t} distributed, rather than Gaussian as the name would imply.

One way in which bias is introduced to posterior model is via the estimation of the posterior means and covariances (equations (\ref{eq:posterior_m}) and (\ref{eq:posterior_S}), respectively). These updates depend on weighted averages of the prior mean and sample mean. The sample mean $\bar{\mathbf{x}}_k$ becomes biased because, for a given label, regions of the feature space that are shared with or close to another label are sampled preferentially. Whilst these parameters do introduce bias, this bias facilitates the rapid improvement in decision-making performance achieved via risk-based active learning. These biased estimates cause the distributions corresponding to classes of differing optimal strategies to drift closer together, thereby narrowing the decision boundary region inferred via the model.

The bias introduced via the posterior mixing proportions (equation (\ref{eq:mixpropupdate})) can likely be ascribed to the deterioration in decision performance. This posterior update is simply a summation of the prior weight for a given class and the number of samples corresponding to that class in $\mathcal{D}_l$. As is demonstrated in Figure \ref{fig:cprop_ext}, the risk-based active learning process causes the class proportions in $\mathcal{D}_l$ to greatly deviate from those in the underlying distribution. This effect undoubtedly results in misclassifications, and therefore erroneous decisions, particularly for data points close to decision boundaries.

Finally, an additional effect of sampling bias,not immediately obvious from the case study presented above, arises via the posterior estimates for the Student-\textit{t} distribution's degrees of freedom parameter. As previously mentioned, this parameter governs the weightedness of the distribution tails; distributions with fewer degrees of freedom possess tails that are weighted more heavily and thus hold a greater probability mass away from the mean. In contrast, distributions with a greater number of degrees of freedom have more probability mass concentrated around the mean. From equations (\ref{eq:posterior_v}) and (\ref{eq:studentT}), it can be seen that the degrees of freedom are a monotonically-increasing function of the number of samples for a given class. This characteristic has implications for how an agent utilising an actively-learned model responds when presented with outlying observations. As some classes are seldom queried during risk-based active learning, a classifier trained accordingly will make overconfident predictions on outlying data in favour of classes that are under-represented in the training data. This result has particularly concerning consequences for structural maintenance decision processes if benign classes are under-represented as extreme structural damage states may go ignored. Further discussion around the treatment of outlying data for classifiers used in an SHM decision-support context is provided in Section \ref{sec:Conclusion}.

In essence, risk-based active learning provides a useful methodology for developing statistical classifiers for decision-support applications because of its ability to consider the value of information with respect to the cost of obtaining said information and thereby improve decision-making performance in a cost-effective manner. Nevertheless, in some scenarios, because of inherent issues with sampling bias, the use of generative models alone may not serve as the most appropriate foundation for risk-based active learning algorithms. Although the decline in decision performance observed in the previous subsection is small, ultimately, it corresponds to a loss in resource/utility and therefore it is desirable to rectify this issue. The following sections outline and demonstrate approaches to curtail the detrimental effects of sampling bias in such algorithms.

\section{Approaches to Address Sampling Bias}
\label{sec:Approaches}

As discussed previously, sampling bias can occur in generative models learned by risk-based active learning. The problem is from posterior estimates of distribution parameters and a result of imbalanced class representation and non-uniform coverage of the feature space within $\mathcal{D}_l$.

The current section aims to address these issues by considering two main approaches; semi-supervised learning, and discriminative classifiers. Two methods for semi-supervised learning are considered; expectation-maximisation, and latent-state smoothing. The multiclass relevance vector machine (mRVM) is considered as the discriminative classifier. Two formulations of the mRVM are considered; $\text{mRVM}_1$ and $\text{mRVM}_2$ \cite{Psorakis2010,Damoulas2008}.

\subsection{Semi-supervised Learning}\label{sec:semisupervised}

Alongside active learning, semi-supervised learning is a form of partially-supervised learning -- utilising both $\mathcal{D}_l$ and $\mathcal{D}_u$ to inform the classification mapping. The fundamental principle of semi-supervised learning that distinguishes it from active learning is as follows; data in the unlabelled dataset $\mathcal{D}_u$ can be given \textit{pseudo-labels} that are informed by the ground-truth labels available in $\mathcal{D}_l$. By incorporating pseudo-labels for unqueried data into the risk-based active learning algorithm, class imbalance and inadequate coverage of the feature space can be rectified.

There are several approaches to semi-supervised learning. The simplest approach, \textit{self-labelling}, involves training a classifier using $\mathcal{D}_l$ in a supervised manner \cite{Chapelle2006}. Pseudo-labels can then be assigned to the unlabelled data according to the predictions of this classifier. The model can subsequently be retrained incorporating both ground-truth labels and pseudo-labels. The effectiveness of self-labelling is strongly conditioned on the implementation and the underlying supervised-learning algorithm. More advanced approaches that utilise \textit{low-density-separation} \cite{Chapelle2006} and \textit{graph-based} learners \cite{Chen2014} are available.

Semi-supervised learning has been applied to pattern recognition problems for SHM \cite{Bull2020}. This method of learning brings several benefits, such as an increased utilisation of information obtained via costly structural inspections.

\subsubsection{Expectation-Maximisation}

The first of the approaches to semi-supervised learning considered here, aims to exploit the generative mixture model form of the statistical classifier presented in Section \ref{sec:GMM}. Generative models can conveniently account for labelled and unlabelled data; this is achieved by modifying the \textit{expectation-maximisation} (EM) algorithm \cite{Dempster1977}, typically used for unsupervised density estimation, such that the log-likelihood of the model is maximised over both unlabelled and labelled data. Details of this EM algorithm are provided in Appendix \ref{app:EM_GMMs}.

The EM algorithm iterates between E-steps and M-steps, resulting in a hill-climbing search that terminates when the log-likelihood (equation (\ref{eq:EMloglik})) converges to a local maximum in the parameter space. The EM algorithm is sensitive to initial conditions and in many applications requires multiple random initialisations. For the current application, however, the model is initialised using the labelled dataset; this additional information mitigates the need to re-initialise. Examination of the parameter updates provided in equations (\ref{eq:EMupdate1}) to (\ref{eq:EMupdate9}) reveals that the information introduced by the incorporation of unlabelled data via EM learning has the ability to reduce the bias in parameter estimates by considering a broader span of the feature space.

\begin{figure*}[ht]
	\centering
	\scalebox{0.8}{
		\begin{tikzpicture}[auto]
			\begin{footnotesize}
				\tikzstyle{decision} = [diamond, draw, text width=4em, text badly centered, inner sep=2pt]
				\tikzstyle{block} = [rectangle, draw, text width=10em, text centered, rounded corners, minimum height=4em]
				\tikzstyle{block2} = [rectangle, draw, text width=10em, text centered, rounded corners, minimum height=4em, fill=black!5]
				\tikzstyle{line} = [draw, -latex']
				\tikzstyle{cloud} = [draw=black!50, circle, node distance=3cm, minimum height=2em, text width=4em,text centered]
				\tikzstyle{point}=[draw, circle]
				\node [block2, node distance=4em] (start) {start:\\ initial training-set, $\mathcal{D} = \mathcal{D}_l \cup \mathcal{D}_u$, $\mathcal{D}_u = \emptyset$};
				\node [point, below of=start, node distance=12mm] (point) {};
				\node [block, below of=point, node distance=12mm] (train) {train model\\ $p(\mathbf{x}, y | \mathcal{D}_l)$};
				\node [decision, below of=train, node distance=25mm] (empty) {$\mathcal{D}_u = \emptyset$?};
				\node [block, left of=empty, node distance=40mm] (EM update) {EM update $p(\mathbf{x}, y | \mathcal{D}_l,\mathcal{D}_u)$};
				\node [decision, below of=empty, node distance=25mm] (new data) {new data?};
				\node [block2, left of=new data, node distance=40mm] (stop) {stop};
				\node [block, below of=new data, node distance=25mm] (update u) {update unlabelled set, $\mathcal{D}_u$};
				\node [cloud, left of=update u, node distance=40mm] (measured data) {new observation, $\tilde{\mathbf{x}}_t$};
				\node [block, below of=update u, node distance=25mm] (predict) {predict \\ $p(\tilde{y}_t | \tilde{\mathbf{x}}_t,\mathcal{D}_l)$};
				\node [block, right of=predict, node distance=42mm] (EVPI) {compute \\ $\text{EVPI}(d_t | y_t)$};
				\node [decision, above of=EVPI, node distance=25mm] (inspect) {$\text{EVPI} > C_{\text{ins}}$?};
				\node [block, above of=inspect, node distance=25mm] (query) {query health-state information for $\mathbf{x}_t$};
				\node [cloud, right of=query, node distance=40mm] (annotate) {$y_t$ provided via inspection by engineer};
				\node [block, above of=query, node distance=25mm] (update l) {update $\mathcal{D}_l$ to include new queried labels};
				\path [line] (start) -- (point);
				\path [line] (point) -- (train);
				\path [line] (train) -- (empty);
				\path [line] (empty) -- node {yes}(new data);
				\path [line] (empty) -- node {no}(EM update);
				\path [line] (EM update) -- (new data);
				\path [line] (new data) -- node {no}(stop);
				\path [line] (new data) -- node {yes}(update u);
				\path [line, dashed, draw=black!50] (measured data) -- (update u);
				\path [line] (update u) -- (predict);
				\path [line] (predict) -- (EVPI);
				\path [line] (EVPI) -- (inspect);
				\path [line] (inspect) -- node {yes}(query);
				\path [line] (inspect) -- node {no}(new data);
				\path [line, dashed, draw=black!50] (annotate) -- (query);
				\path [line] (query) -- (update l);
				\path [line] (update l) |- (point);
			\end{footnotesize}
		\end{tikzpicture}
	}
	\caption{Flow chart to illustrate the risk-based active learning process incorporating expectation-maximisation.}
	\label{fig:RBAL_EM}
\end{figure*}
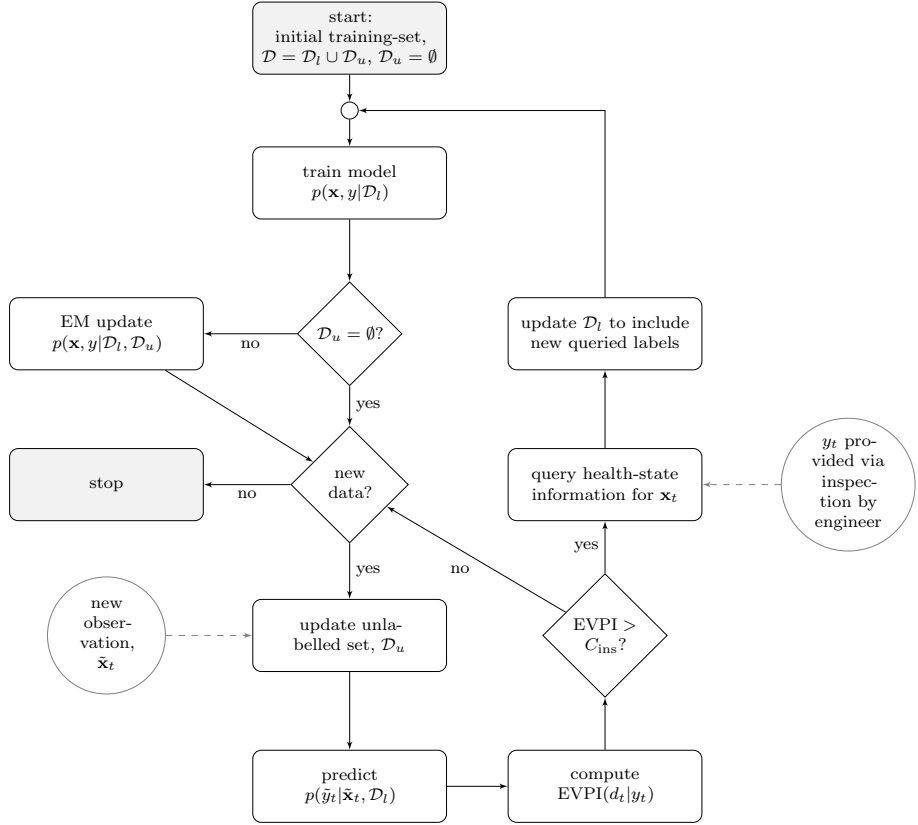

Within the risk-based active learning algorithm, it is possible to apply EM every time a new unlabelled observation is acquired; however, this would be computationally expensive. To limit the computational cost of the modified active-learning algorithm, the EM update was only applied following the retraining of the model subsequent to the acquisition of a new ground-truth label obtained via inspection. A flow chart detailing the risk-based active-learning algorithm modified to incorporate EM is shown in Figure \ref{fig:RBAL_EM}.

\subsubsection{Latent Health-State Smoothing}
 
The second approach to semi-supervised learning aims to exploit the Markovian model underpinning the decision process outlined in Section \ref{sec:DecProc1}. The Markov property asserts that future states of a stochastic process are conditionally independent of past states, given the present states, i.e.\ $p(y_{T+1}|y_{1:T}) = p(y_{T+1}|y_T)$. Here, the notation $1$:$T$ is employed to refer to all instances from $1$ to $T$, inclusive.

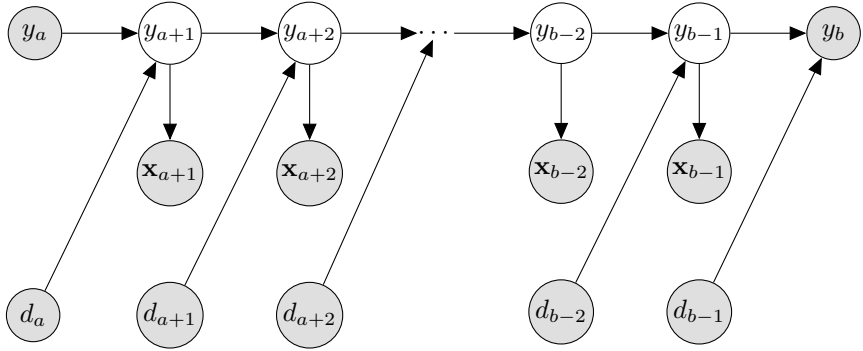
\begin{figure}[ht!]
	\centering
	\begin{tikzpicture}[x=1.7cm,y=1.8cm]
		
		\node[obs] (ya) {$y_a$} ;
		\node[latent, right=1cm of ya] (ya1) {$y_{a+1}$} ;
		\node[obs, below=1cm of ya1] (xa1) {$\mathbf{x}_{a+1}$} ;
		\node[obs,below=1cm of xa1] (da1) {$d_{a+1}$} ;
		\node[obs,left=1cm of da1] (da) {$d_{a}$} ;
		\node[latent, right=1cm of ya1] (ya2) {$y_{a+2}$} ;
		\node[obs, below=1cm of ya2] (xa2) {$\mathbf{x}_{a+2}$} ;
		\node[obs,below=1cm of xa2] (da2) {$d_{a+2}$} ;
		\node[const,right=1cm of ya2] (dots) {$\cdots$} ;
		\node[latent, right=1cm of dots] (yb2) {$y_{b-2}$} ;
		\node[obs, below=1cm of yb2] (xb2) {$\mathbf{x}_{b-2}$} ;
		\node[obs,below=1cm of xb2] (db2) {$d_{b-2}$} ;
		\node[latent, right=1cm of yb2] (yb1) {$y_{b-1}$} ;
		\node[obs, below=1cm of yb1] (xb1) {$\mathbf{x}_{b-1}$} ;
		\node[obs,below=1cm of xb1] (db1) {$d_{b-1}$} ;
		\node[obs,right=1cm of yb1] (yb) {$y_{b}$} ;

		\edge {ya} {ya1} ; %
		\edge {da} {ya1} ; %
		\edge {ya1} {xa1} ; %
		\edge {ya1} {ya2} ; %
		\edge {da1} {ya2} ; %
		\edge {ya2} {xa2} ; %
		\edge {ya2} {dots} ; %
		\edge {da2} {dots} ; %
		\edge {dots} {yb2} ; %
		\edge {yb2} {xb2} ; %
		\edge {db2} {yb1} ; %
		\edge {yb2} {yb1} ; %
		\edge {yb1} {xb1} ; %
		\edge {db1} {yb} ; %
		\edge {yb1} {yb} ; %

	\end{tikzpicture}
	\caption{A Bayesian network representation of a modified HMM with latent states conditionally dependent on known historical actions.}
	\label{fig:HMM1}
\end{figure}

Hidden Markov models (HMMs) are a statistical state-space representation of stochastic processes for which the Markov property holds. Within HMMs, it is typically assumed that latent states are partially-observable and must be inferred via indirect observations. For the current application, one can form a HMM for latent states between inspections at times $t=a$ and $t=b$. As they have occurred in the past, modifications can be made to the HMM such that decisions between $a$ and $b$ are treated as observed exogenous inputs that influence the subsequent latent state. A Bayesian network of this process can be seen in Figure \ref{fig:HMM1}.

Given that a modified HMM can be formed for states between two inspections, posterior marginal distributions conditioned on both inspections and intermediate observations can be obtained for the latent states via a smoothing algorithm. For this application, the \textit{forward-backward} algorithm was employed to infer the smoothed distributions \cite{Einicke2012}. The forward-backward algorithm is a message-passing algorithm and involves three key steps: (1) computing forward messages, (2) computing backward messages, and (3) computing smoothed distributions.

The forward messages $\phi_t$ relate to the joint probability distribution $p(\tilde{\mathbf{x}}_{a+1:t},\tilde{y}_{t}|y_a)$ $\forall t \in (a,b)$, and can be computed recursively by invoking the Markov property, 

\begin{equation}
	\phi_t = p(\tilde{\mathbf{x}}_t|y_t) \sum_{y_{t-1}} \phi_{t-1} \cdot p(y_t|y_{t-1},d_{t-1})
\end{equation}

\noindent where the message $\phi_a$ is initialised as $\delta_{k,y_a}$. Here, $p(y_t|y_{t-1},d_{t-1})$ is equivalent to the transition matrix $p(y_{t+1}|y_{t},d_{t})$ and $p(\tilde{\mathbf{x}}_t|y_t)$ is the posterior predictive distribution of the statistical classifier.

The backward messages $\psi_t$ relate to the conditional distribution $p(\tilde{\mathbf{x}}_{t:b-1}|\tilde{y}_{t:b-1}, y_b)$ $\forall t \in (a,b)$, and can again be computed recursively as follows,

\begin{equation}
	\psi_t = \sum_{y_{t+1}} \psi_{t+1} \cdot p(\tilde{\mathbf{x}}_{t+1}|y_{t+1}) \cdot p(y_t|y_{t-1},d_{t-1})^{\top}
\end{equation}

\noindent where the message $\psi_b$ is initialised as $\delta_{k,y_b}$. Here, it is worth recalling that $p(y_{t+1}|y_{t},d_{t})$ can be considered to be a square $K \times K$ matrix for a given $d_t$.

To obtain the posterior smoothed distributions $p(\tilde{y}_t|y_a,y_b,\tilde{\mathbf{x}}_{a+1:b-1},d_{a:b-1})$, one can simply multiply together the forward and backward messages and normalise to unity \cite{Binder1997},

\begin{equation}
	p(\tilde{y}_t|y_a,y_b,\tilde{\mathbf{x}}_{a+1:b-1},d_{a:b-1}) \propto \phi_t \psi_t
\end{equation}

By taking the MAP of the smoothed distribution, pseudo-labels $\hat{y}_t$ may be assigned to $\tilde{\mathbf{x}}$ such that,

\begin{equation}
	\mathcal{D}_u = \{ (\tilde{\mathbf{x}}_t,\hat{y}_t)|\tilde{\mathbf{x}}_t \in X, \hat{y}_t \in Y \}_{t=1}^{m}
\end{equation}

\noindent and where,

\begin{equation}
	\hat{y}_t = \text{argmax}_{\tilde{y}_t}p(\tilde{y}_t|y_a,y_b,\tilde{\mathbf{x}}_{a+1:b-1},d_{a:b-1})
\end{equation}

Using the pseudo-labels provided for $\mathcal{D}_u$, in addition to $\mathcal{D}_l$, updated estimates for $\bm{\Theta}$ can be computed in a supervised manner via equations (\ref{eq:posterior_m}) to (\ref{eq:posterior_S}) and (\ref{eq:mixpropupdate}).

As with the decision process underpinning risk-based active learning, latent-state smoothing relies on the specified observation and transition models. If poor models are chosen, erroneous pseudo-labels may be provided for unlabelled data.

Similar to the EM approach, the smoothing approach to semi-supervised learning addresses sampling bias by reducing the class imbalance and generating a training dataset that is more representative of the underlying distribution. Unlike the EM approach, latent-state smoothing utilises information encoded within the decision process transition model which can yield some powerful inferences. One can imagine a scenario where it is determined via inspection that the structure is undamaged at time $a$, and time $b$, and furthermore, it is known that no interventions were performed on the structure inside the interval $(a,b)$. Under these conditions, and given that the relevant transition model allows only monotonic degradation, it can be inferred that the structure is undamaged with unit probability for all instances in $(a,b)$.

\begin{figure*}[ht]
	\centering
	\scalebox{0.8}{
		\begin{tikzpicture}[auto]
			\begin{footnotesize}
				\tikzstyle{decision} = [diamond, draw, text width=4em, text badly centered, inner sep=2pt]
				\tikzstyle{block} = [rectangle, draw, text width=10em, text centered, rounded corners, minimum height=4em]
				\tikzstyle{block2} = [rectangle, draw, text width=10em, text centered, rounded corners, minimum height=4em, fill=black!5]
				\tikzstyle{line} = [draw, -latex']
				\tikzstyle{cloud} = [draw=black!50, circle, node distance=3cm, minimum height=2em, text width=4em,text centered]
				\tikzstyle{point}=[draw, circle]
				\node [block2, node distance=4em] (start) {start:\\ initial training-set, $\mathcal{D} = \mathcal{D}_l \cup \mathcal{D}_u$, $\mathcal{D}_u = \emptyset$};
				\node [point, below of=start, node distance=12mm] (point) {};
				\node [block, below of=point, node distance=12mm] (train) {train model\\ $p(\mathbf{x}, y | \mathcal{D}_l, \mathcal{D}_u)$};
				\node [decision, below of=train, node distance=25mm] (new data) {new data?};
				\node [block2, left of=new data, node distance=40mm] (stop) {stop};
				\node [block, below of=new data, node distance=25mm] (update u1) {update $\mathcal{D}_u$ with $\tilde{\mathbf{x}}_t$};
				\node [cloud, left of=update u1, node distance=40mm] (measured data) {new observation, $\tilde{\mathbf{x}}_t$};
				\node [block, below of=update u1, node distance=25mm] (predict) {predict \\ $p(\tilde{y}_t | \tilde{\mathbf{x}}_t,\mathcal{D}_l)$};
				\node [block, below of=predict, node distance=25mm] (EVPI) {compute \\ $\text{EVPI}(d_t | y_t)$};
				\node [decision, right of=EVPI, node distance=60mm] (inspect) {$\text{EVPI} > C_{\text{ins}}$?};
				\node [block, above of=inspect, node distance=25mm] (query) {query health-state information for $\mathbf{x}_t$};
				\node [cloud, right of=query, node distance=40mm] (annotate) {$y_t$ provided via inspection by engineer};
				\node [block, above of=query, node distance=25mm] (update l) {update $\mathcal{D}_l$ to include new queried labels};
				\node [block, above of=update l, node distance=25mm] (smooth) {smooth $p(\tilde{y}_t|y_a,y_b,\ldots$ $\tilde{\mathbf{x}}_{a+1:b-1},d_{a:b-1}),$ $\forall t \in (a,b)$};
				\node [block, above of=smooth, node distance=25mm] (update u2) {update $\mathcal{D}_u$ with  $\hat{y}_t$};
				\path [line] (start) -- (point);
				\path [line] (point) -- (train);
				\path [line] (train) -- (new data);
				\path [line] (new data) -- node {no}(stop);
				\path [line] (new data) -- node {yes}(update u1);
				\path [line, dashed, draw=black!50] (measured data) -- (update u1);
				\path [line] (update u1) -- (predict);
				\path [line] (predict) -- (EVPI);
				\path [line] (EVPI) -- (inspect);
				\path [line] (inspect) -- node {yes}(query);
				\path [line] (inspect) -- node {no}(new data);
				\path [line, dashed, draw=black!50] (annotate) -- (query);
				\path [line] (query) -- (update l);
				\path [line] (update l) -- (smooth);
				\path [line] (smooth) -- (update u2);
				\path [line] (update u2) |- (point);
			\end{footnotesize}
		\end{tikzpicture}
	}
	\caption{Flow chart to illustrate the risk-based active learning process incorporating latent-state smoothing.}
	\label{fig:RBAL_smooth}
\end{figure*}
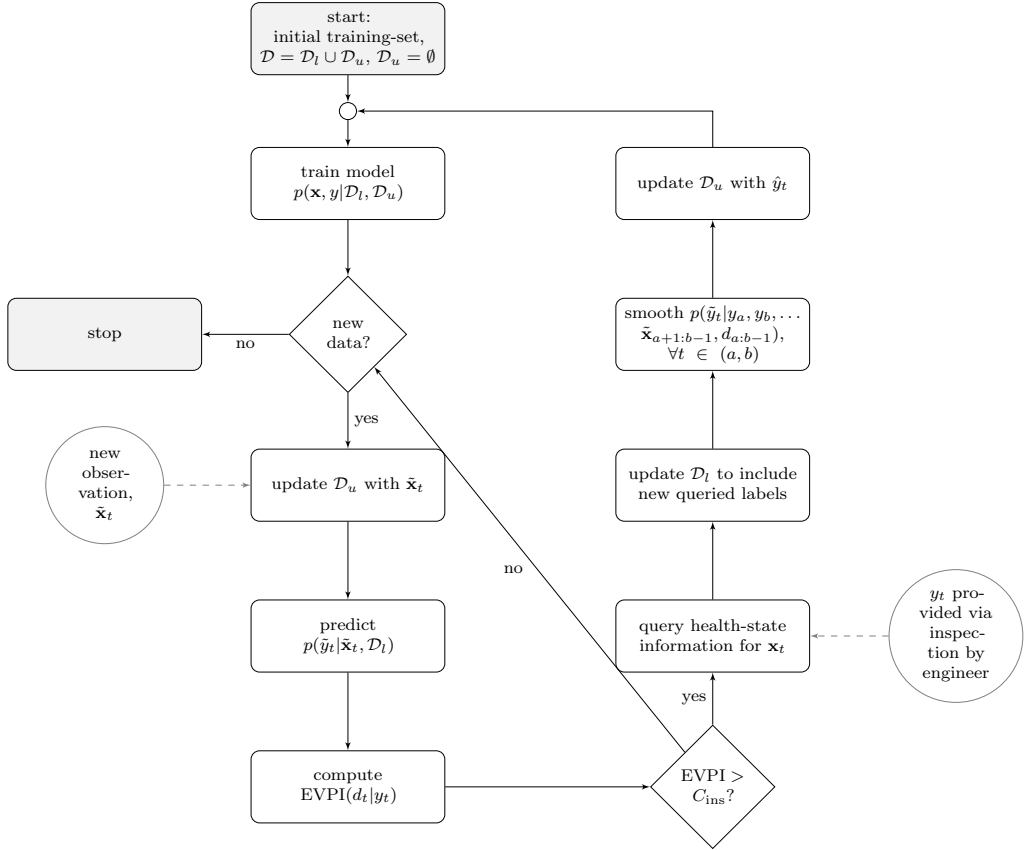

Latent-state smoothing fits naturally within the risk-based active learning framework and can be applied after each inspection. When applied immediately following the first inspection, it is assumed that at the onset of the active learning algorithm, the structure was in its undamaged state. For the current paper, latent-states are smoothed only once, using the version of the classifier available up to the most recent inspection. An alternative approach in which all historical latent-states are smoothed after every inspection using the most up-to-date classifier available may yield better performance, however, would be more computationally expensive. A flow chart detailing the risk-based active learning algorithm modified to incorporate smoothing is shown in Figure \ref{fig:RBAL_smooth}.

\subsubsection{Results}\label{sec:results_ss_ext}

The EM and smoothing approach to semi-supervised learning were each incorporated into the risk-based active learning process and applied to the case study presented in Section {\ref{sec:Results1}. Once again, 1000 repetitions were conducted, each with randomly-selected training and test datasets and with $\mathcal{D}_l$ randomly initialised as a small subset of the training data. For the smoothing approach, decisions were specified according to the hidden labels associated with $\mathcal{D}_u$, such that the state transitions present in the dataset were consistent with the transition models specified in Tables \ref{tab:P_y1_y0_d0} and \ref{tab:P_y1_y0_d1}.

Figures \ref{fig:finalModel_em} and \ref{fig:finalModel_smooth} show a GMM for one of the 1000 runs after the risk-based active-learning process incorporating EM and smoothing, respectively. The initial model prior to risk-based active learning is unchanged from that shown in Figure \ref{fig:initialModel}.

It can be seen from Figures \ref{fig:finalScatter_EM} and \ref{fig:finalScatter_smooth} that, similar to the GMM without semi-supervised learning, risk-based active learning with semi-supervised learning results in labels being obtained for localised regions of the feature space, with Class 3 (moderate damage) being preferentially queried. Nonetheless, these figures show that the clusters learned in a semi-supervised manner fit the data very well. Furthermore, examination of Figures \ref{fig:finalVOI_EM} and \ref{fig:finalVOI_smooth}, reveals that the resulting EVPI distributions over the feature spaces distinctly differ from that in Figure \ref{fig:finalVOI}. For both the EM and smoothing approaches, the introduction of semi-supervised learning into the risk-based active learning process has enabled the inference of a well-defined decision-boundary indicated by the band of high EVPI separating the $2\sigma$ clusters for Class 3 and Class 4.

\begin{figure}[ht!]
	\begin{subfigure}{.5\textwidth}
		\centering
		\scalebox{0.4}{
			\includegraphics{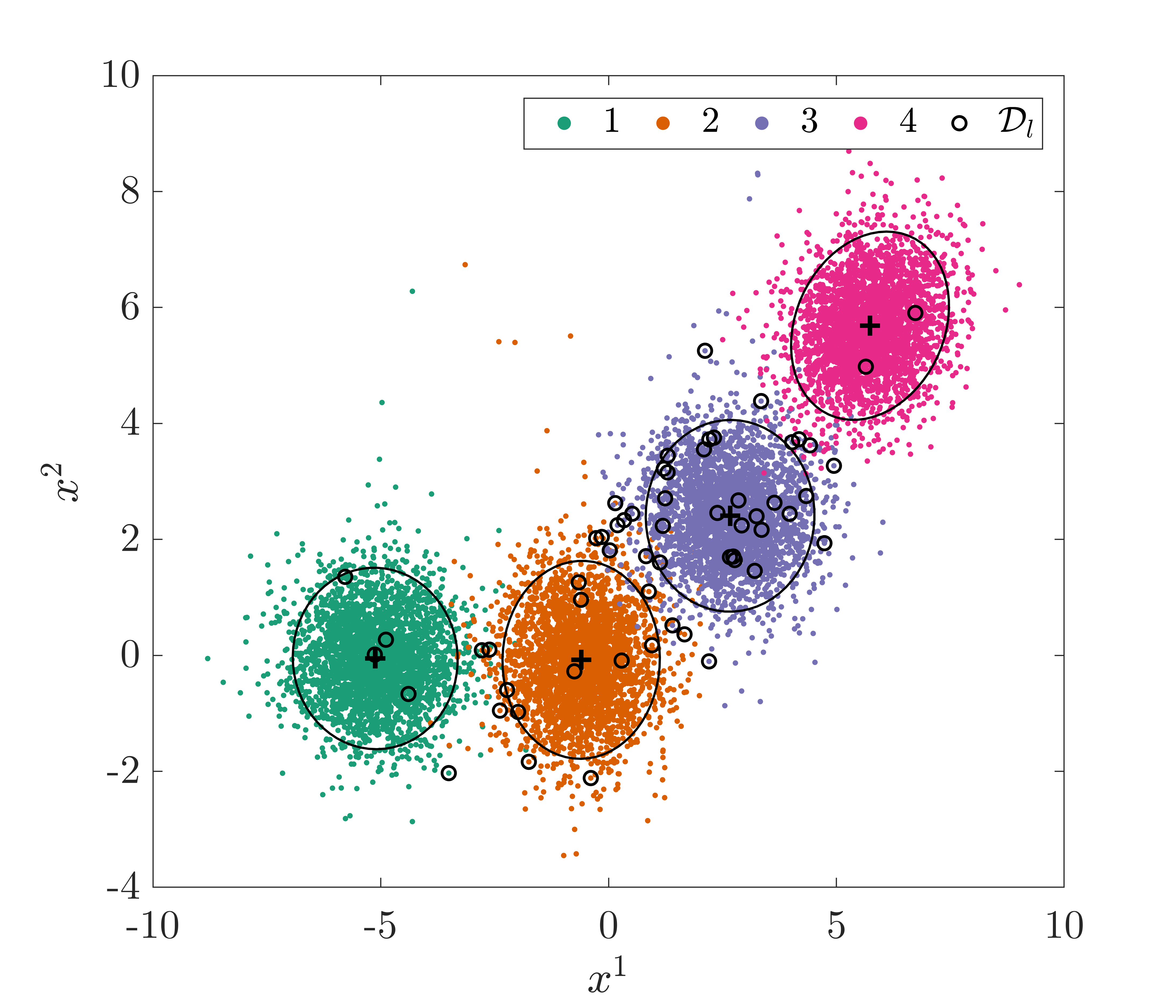}
		}
		\caption{ }
		\label{fig:finalScatter_EM}
	\end{subfigure}
	\begin{subfigure}{.5\textwidth}
		\centering
		\scalebox{0.4}{
			\includegraphics{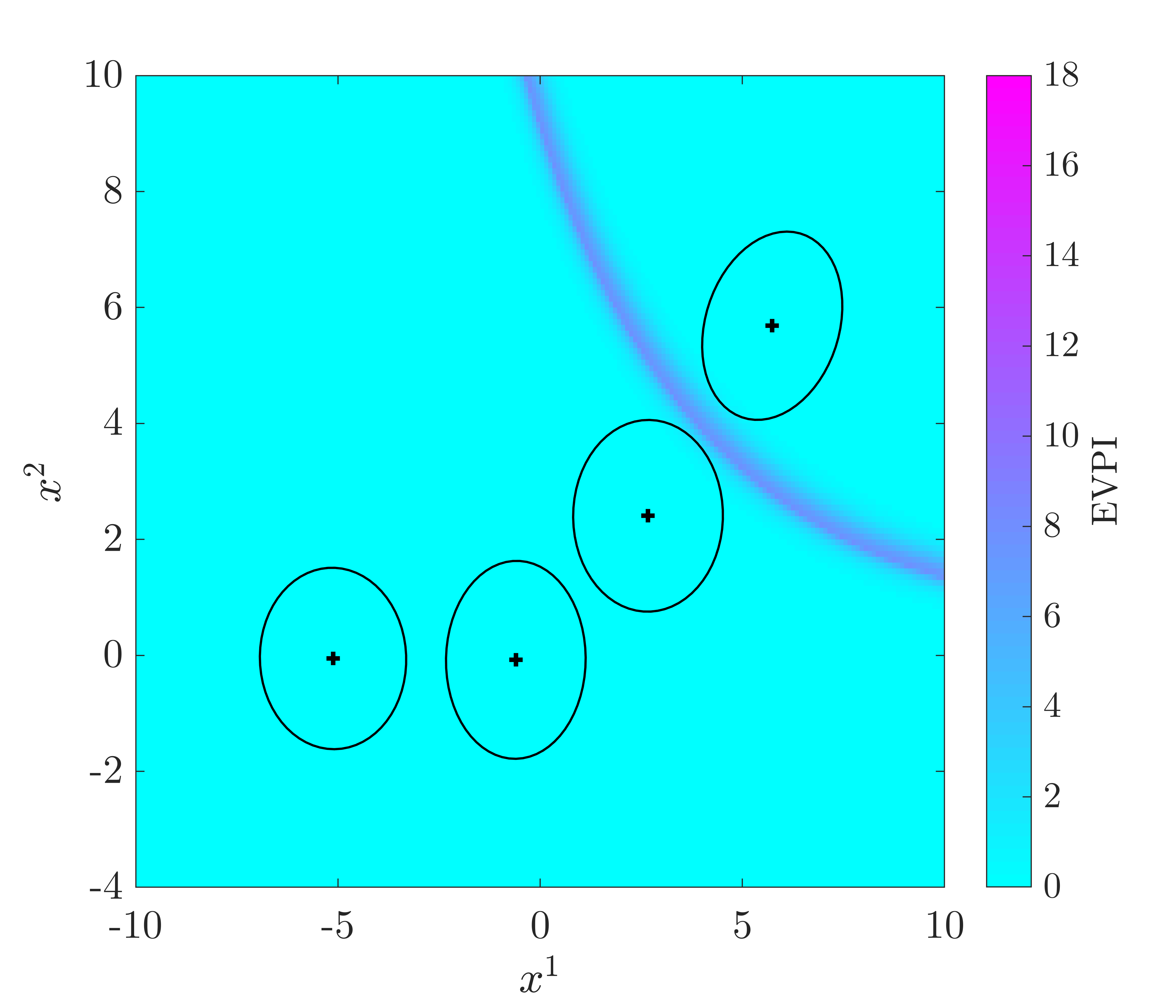}
		}
		\caption{ }
		\label{fig:finalVOI_EM}
	\end{subfigure}
	\caption{A statistical classifier $p(y_t,\mathbf{x}_t,\bm{\Theta})$ following risk-based active learning with semi-supervised learning via EM; \textit{maximum a posteriori} (MAP) estimates of the mean (+) and covariance (ellipses represent 2$\sigma$) are shown. (a) shows the final model overlaid onto the data with labelled data $\mathcal{D}_l$ encircled and (b) shows the resulting EVPI over the feature space.}
	\label{fig:finalModel_em}
\end{figure}

\begin{figure}[ht!]
	\begin{subfigure}{.5\textwidth}
		\centering
		\scalebox{0.4}{
			\includegraphics{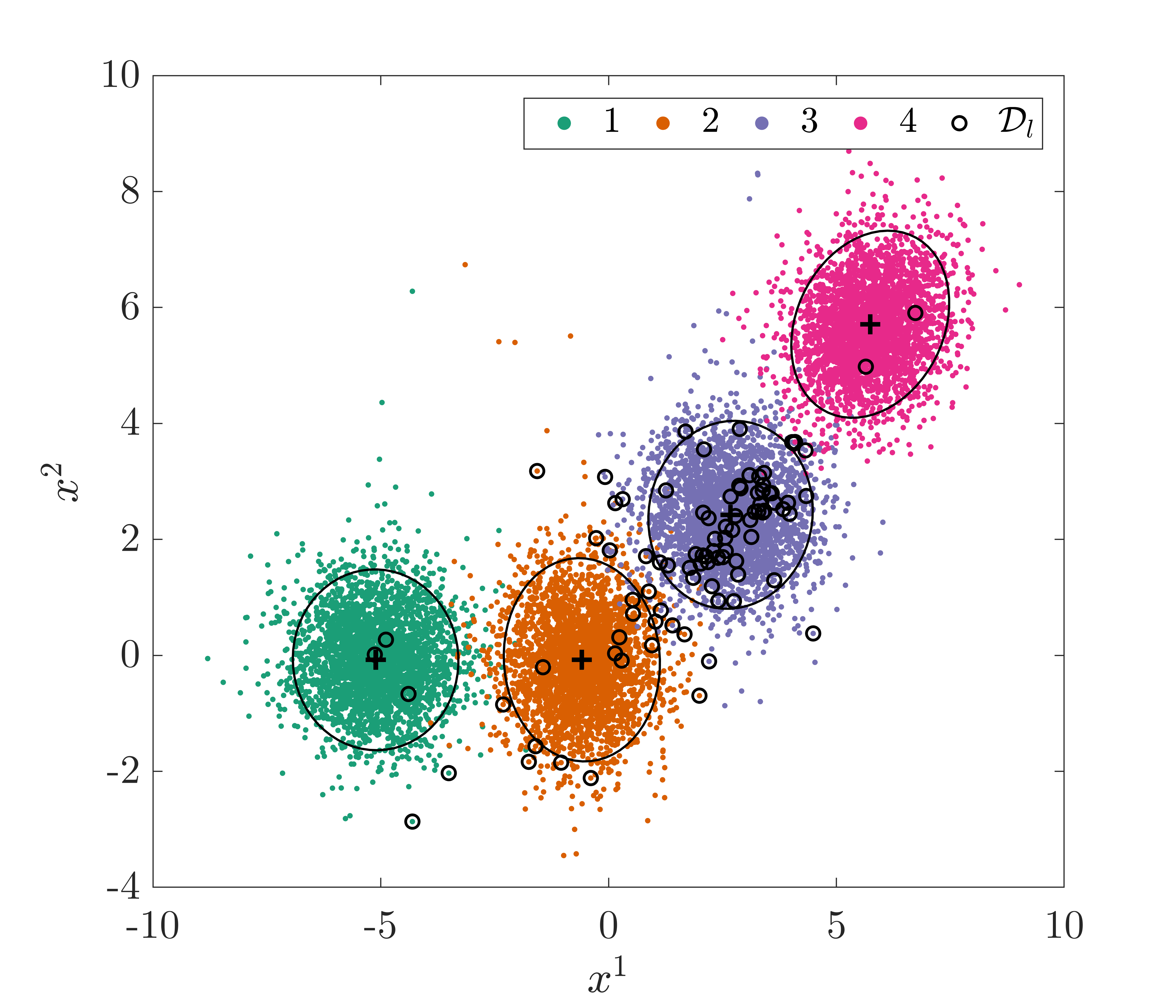}
		}
		\caption{ }
		\label{fig:finalScatter_smooth}
	\end{subfigure}
	\begin{subfigure}{.5\textwidth}
		\centering
		\scalebox{0.4}{
			\includegraphics{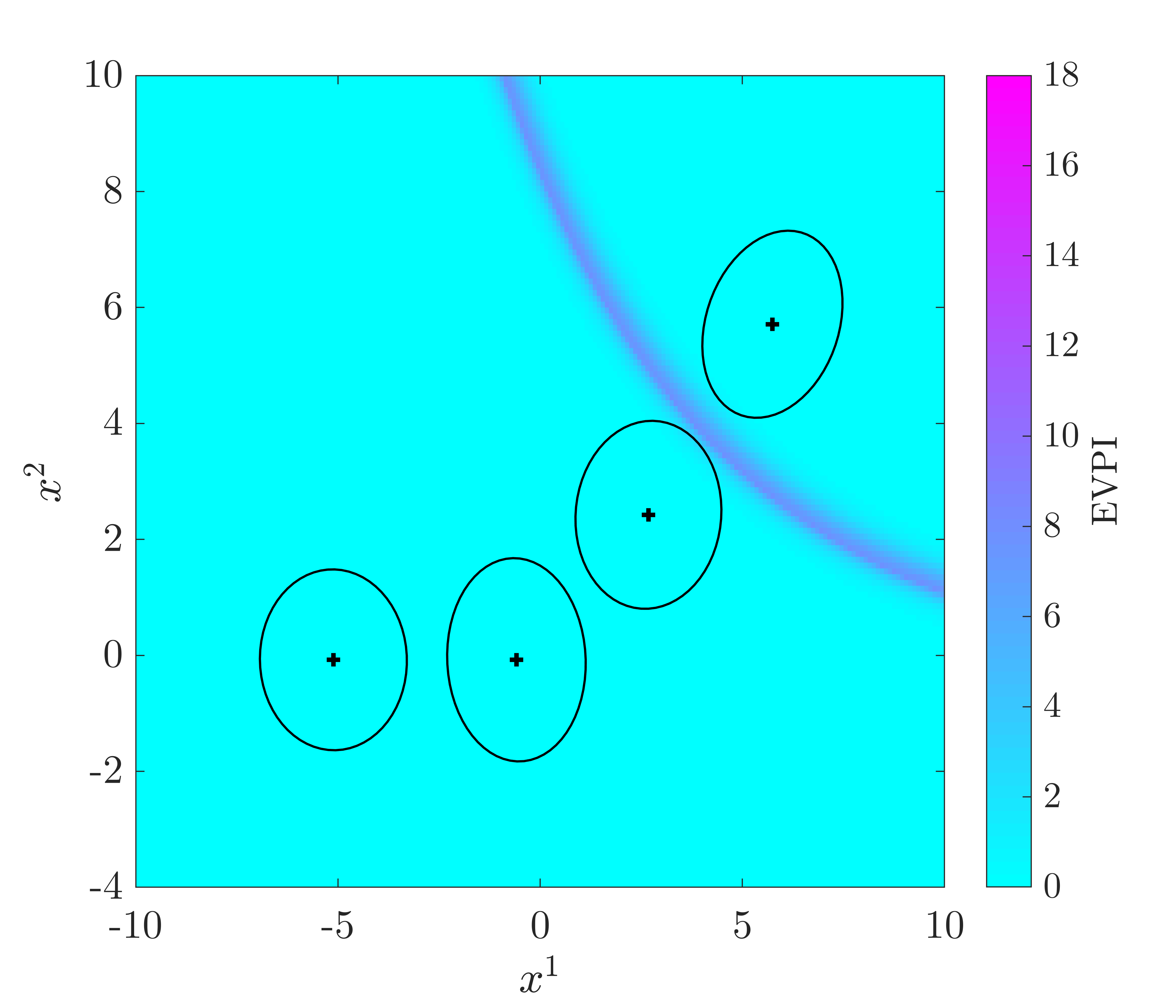}
		}
		\caption{ }
		\label{fig:finalVOI_smooth}
	\end{subfigure}
	\caption{A statistical classifier $p(y_t,\mathbf{x}_t,\bm{\Theta})$ following risk-based active learning with semi-supervised learning via smoothing; \textit{maximum a posteriori} (MAP) estimates of the mean (+) and covariance (ellipses represent 2$\sigma$) are shown. (a) shows the final model overlaid onto the data with labelled data $\mathcal{D}_l$ encircled and (b) shows the resulting EVPI over the feature space.}
	\label{fig:finalModel_smooth}
\end{figure}

Figure \ref{fig:cprop_ss} shows the effective class proportions in the training data $\mathcal{D}$ averaged over the 1000 runs for both the EM and smoothing approaches to active learning. These figures, when considered in the context of the equations specifying the relevant model parameter updates, provide some explanation as to why incorporating semi-supervised learning yields improved model fits over standard risk-based active learning. It can be seen from Figures \ref{fig:cprop_em} and \ref{fig:cprop_smooth} that the effective class proportions in the latter stages of the querying process are much more representative of the underlying data distribution when semi-supervised learning is employed - approximately 25\% of the dataset represented by each of the four classes. As postulated in Section \ref{sec:semisupervised}, having a more representative training dataset results in better-fitting generative distribution, and consequently, a better-defined decision boundary.

\begin{figure}[ht!]
	\begin{subfigure}{.5\textwidth}
		\centering
		\scalebox{0.4}{
			\includegraphics{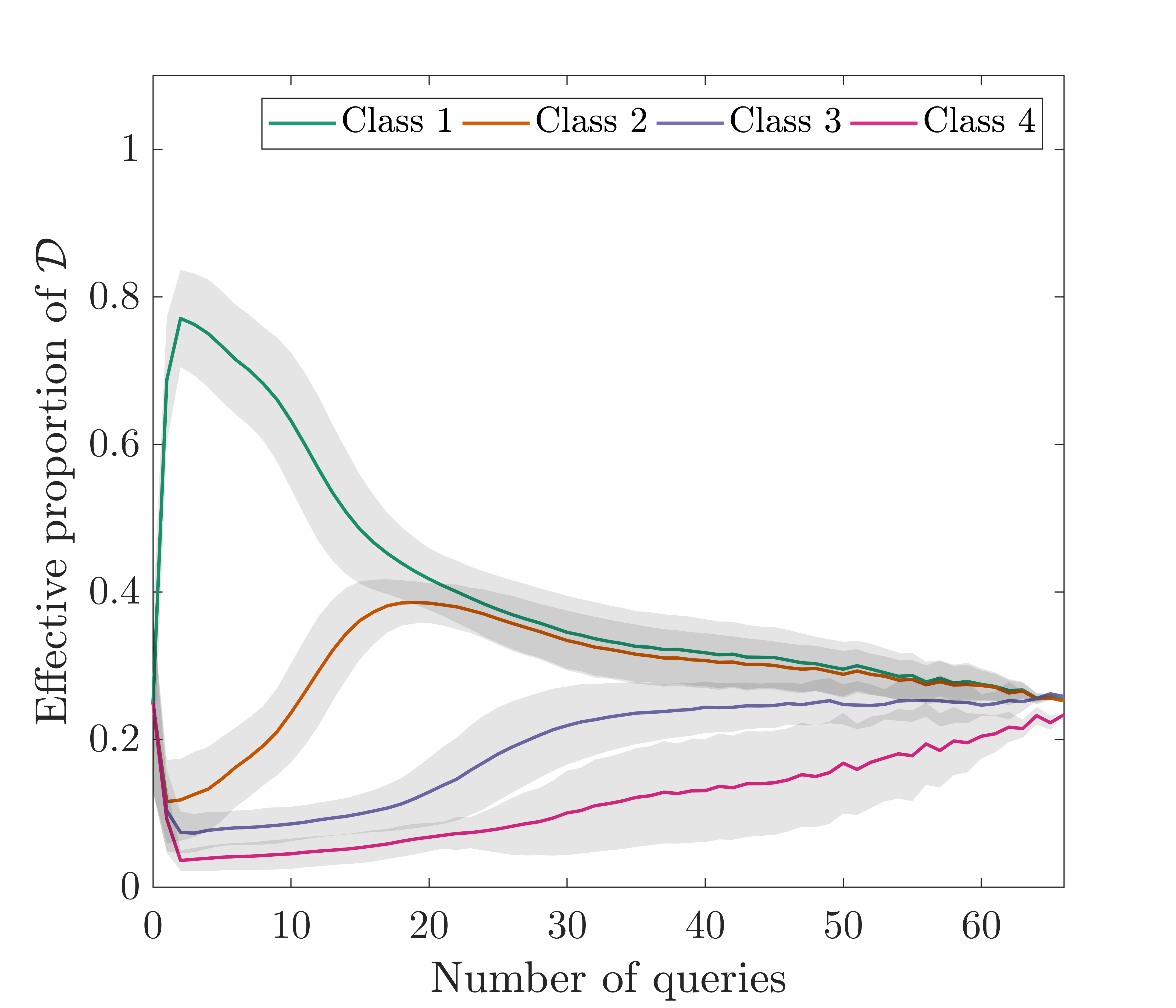}
		}
		\caption{ }
		\label{fig:cprop_em}
	\end{subfigure}
	\begin{subfigure}{.5\textwidth}
		\centering
		\scalebox{0.4}{
			\includegraphics{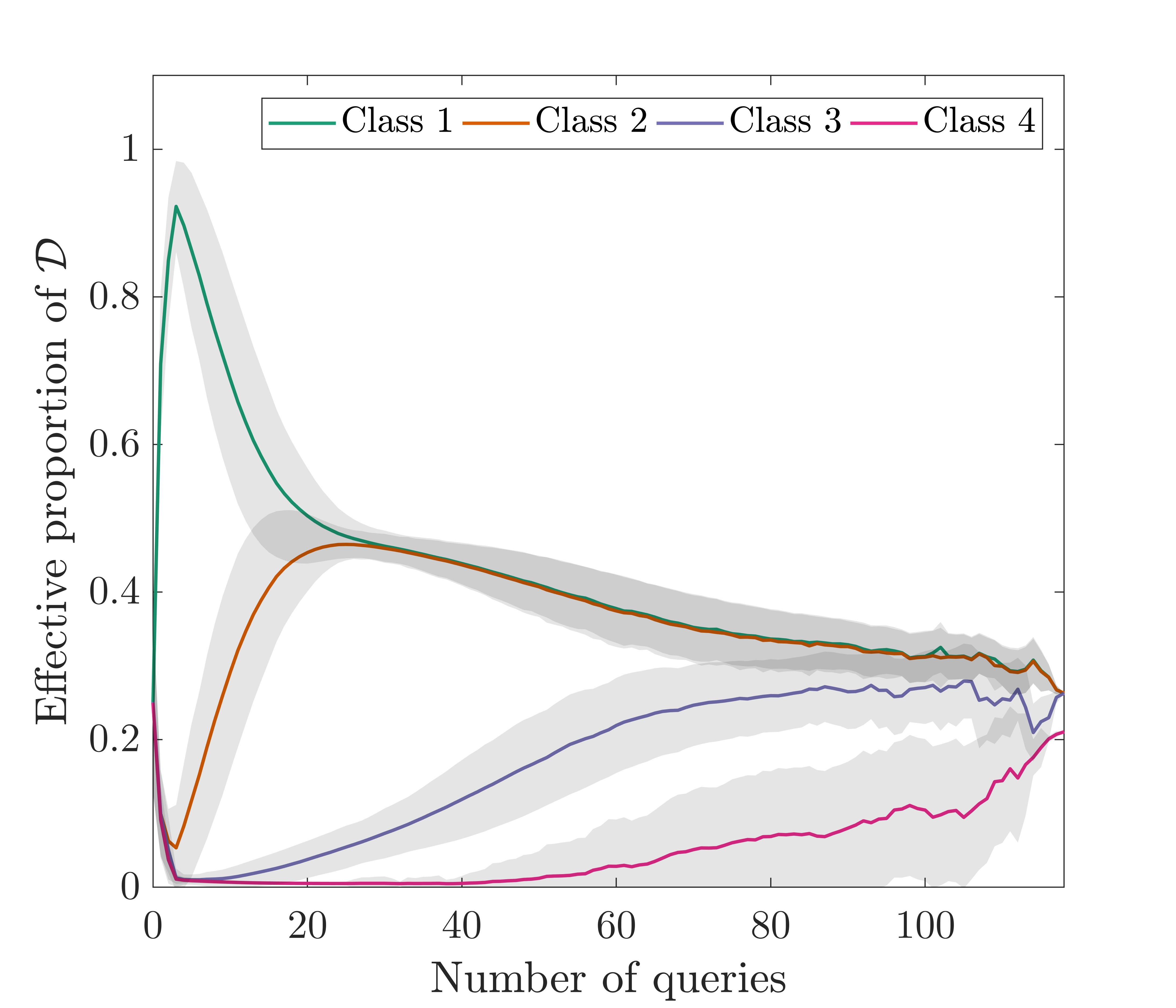}
		}
		\caption{ }
		\label{fig:cprop_smooth}
	\end{subfigure}
	\caption{Variation in effective class proportions within $\mathcal{D}_l$ with number of label queries for an agent utilising a GMM learned from $\mathcal{D}_l$ extended risk-based active learning with semi-supervised updating via (a) expectation-maximisation and (b) latent-state smoothing. Shaded area shows $\pm1\sigma$.}
	\label{fig:cprop_ss}
\end{figure}

Figure \ref{fig:hist_ss} shows how the number of queries varies between each approach to risk-based active learning. It is immediately obvious from Figure \ref{fig:hist_ss}, that incorporating semi-supervised learning, via either EM or smoothing, substantially reduces the number of queries made. The significance of this result becomes most apparent if one recalls that, in the proposed SHM decision context, the number of queries can be mapped directly onto inspection expenditure. This result is to be expected as, when employing semi-supervised learning, supplementary information from the unlabelled dataset is utilised to specify the model each time a query is made. This characteristic allows clusters to become well-defined  more quickly, reducing the area of high-EVPI regions (as is visible in Figures \ref{fig:finalModel_em} and \ref{fig:finalModel_smooth}) meaning fewer queries are made later in the dataset. This result is evident from Figure \ref{fig:all_queries_ss}.

\begin{figure}[ht!]
	\centering
		\scalebox{0.4}{
			\includegraphics{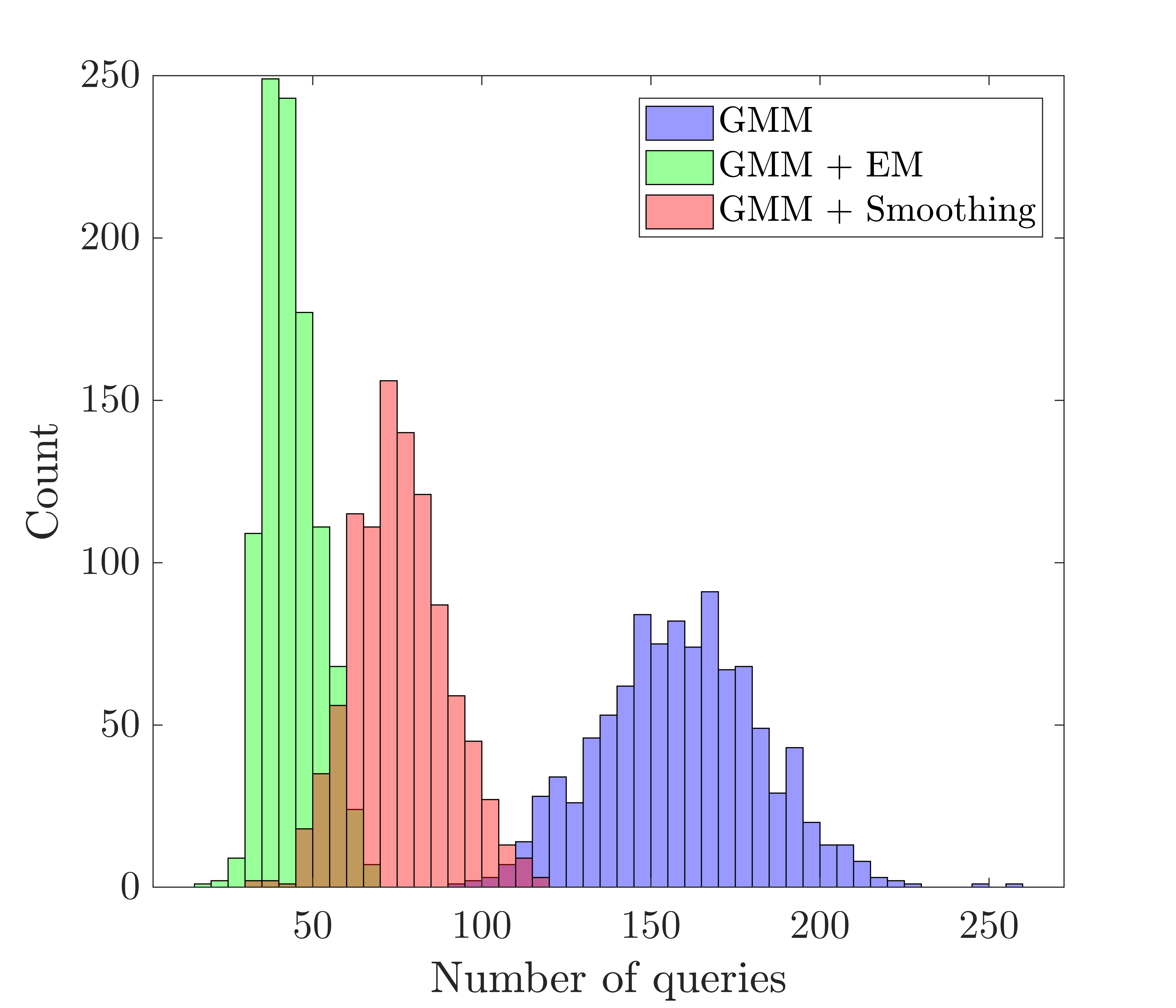}
		}
		\caption{Histograms showing the distribution of the number of queries from 1000 runs of the risk-based active learning of (i) a GMM (blue) (ii) a GMM semi-supervised via expectation-maximisation (green) and (iii) a GMM semi-supervised via latent-state smoothing (red).}
		\label{fig:hist_ss}
\end{figure}

Figure \ref{fig:all_queries_ss} compares the total number of queries for each index in $\mathcal{D}_u$ over the 1000 repetitions of risk-based active learning conducted with a GMM, a GMM with EM and a GMM with smoothing. It can be seen from Figure \ref{fig:all_queries_ss}, that the incorporation of semi-supervised learning into the risk-based active approach results in relatively more queries being obtained during the first occurrence of each class, with relatively fewer queries being made at later occurrences. It can be seen that Class 3 is heavily investigated at each occurrence when semi-supervised methods are not employed. This phenomenon is to be expected when one considers the differences between the high-EVPI regions shown in Figure \ref{fig:finalVOI} and Figures \ref{fig:finalVOI_EM} and \ref{fig:finalVOI_smooth}; as previously discussed semi-supervised learning results in a well-defined decision boundary thereby reducing the likelihood that data will have high value of information.

\begin{figure}[ht!]
	\centering
		\scalebox{0.4}{
			\includegraphics{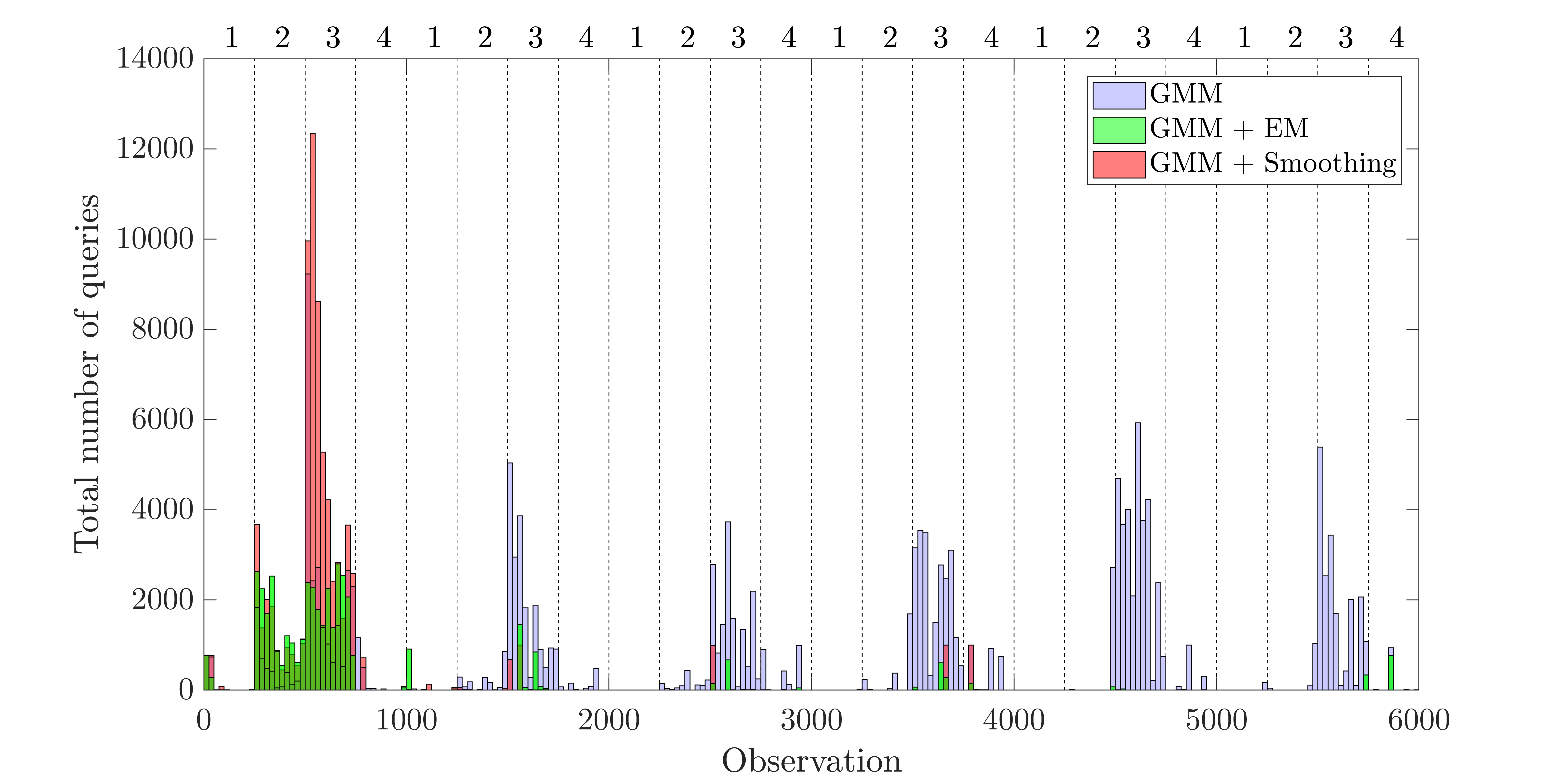}
		}
		\caption{Histograms showing the distribution of the number of queries for each observation in $\mathcal{D}_u$ from 1000 runs of risk-based active learning for (i) a GMM (blue) (ii) a GMM with EM (green) and (iii) a GMM with smoothing (red). The average locations of classes within $\mathcal{D}_u$ are numbered on the upper horizontal axis and transitions are denoted as a dashed line.}
		\label{fig:all_queries_ss}
\end{figure}

Figure \ref{fig:performance_ss} shows median decision accuracies and $f_1$-scores throughout the query process, and compares those measures for risk-based active learning with, and without, semi-supervised learning. From Figure \ref{fig:dacc_ss}, one may be inclined to deduce that the introduction of semi-supervised learning has been detrimental to the performance of risk-based active learning, as both EM and smoothing result in a delayed increase in decision accuracy. However, when considered alongside Figure \ref{fig:all_queries_ss}, one can realise that, because semi-supervised learning results in increased querying early in the dataset, decision accuracy over the whole dataset is improved. Furthermore, a decline in decision performance is not observed for the algorithms incorporating semi-supervised learning; this result is because of the reduction in sampling bias obtained via the inclusion of unlabelled data. Figure \ref{fig:f1score_ss} shows the $f_1$-score classification performance; these results provide further indication that the models learned via semi-supervised risk-based active learning better represent the underlying distribution of data and that the detrimental effects of sampling bias have been reduced.

\begin{figure}[ht!]
	\begin{subfigure}{.5\textwidth}
		\centering
		\scalebox{0.4}{
			\includegraphics{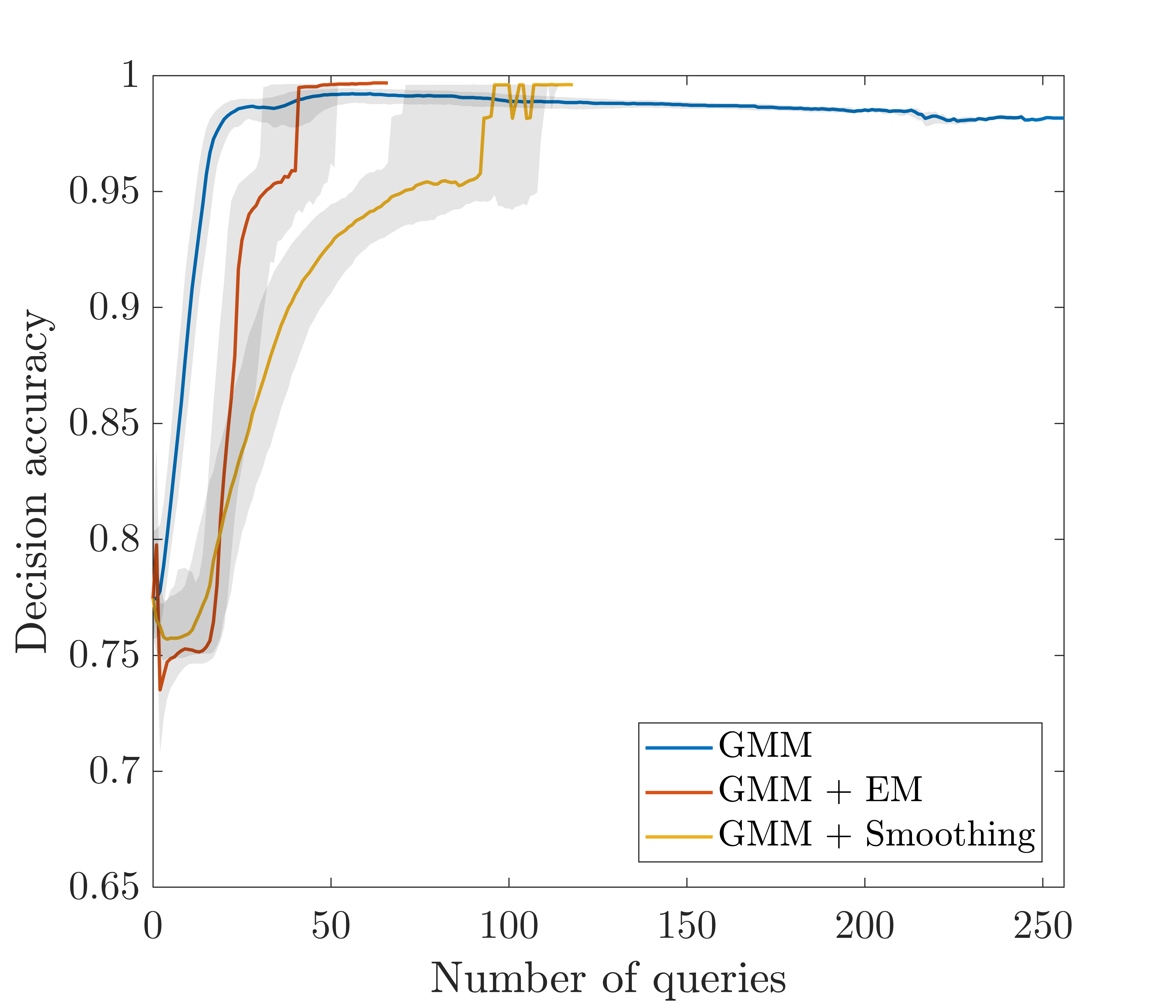}
		}
		\caption{ }
		\label{fig:dacc_ss} 
	\end{subfigure}
	\begin{subfigure}{.5\textwidth}
		\centering
		\scalebox{0.4}{
			\includegraphics{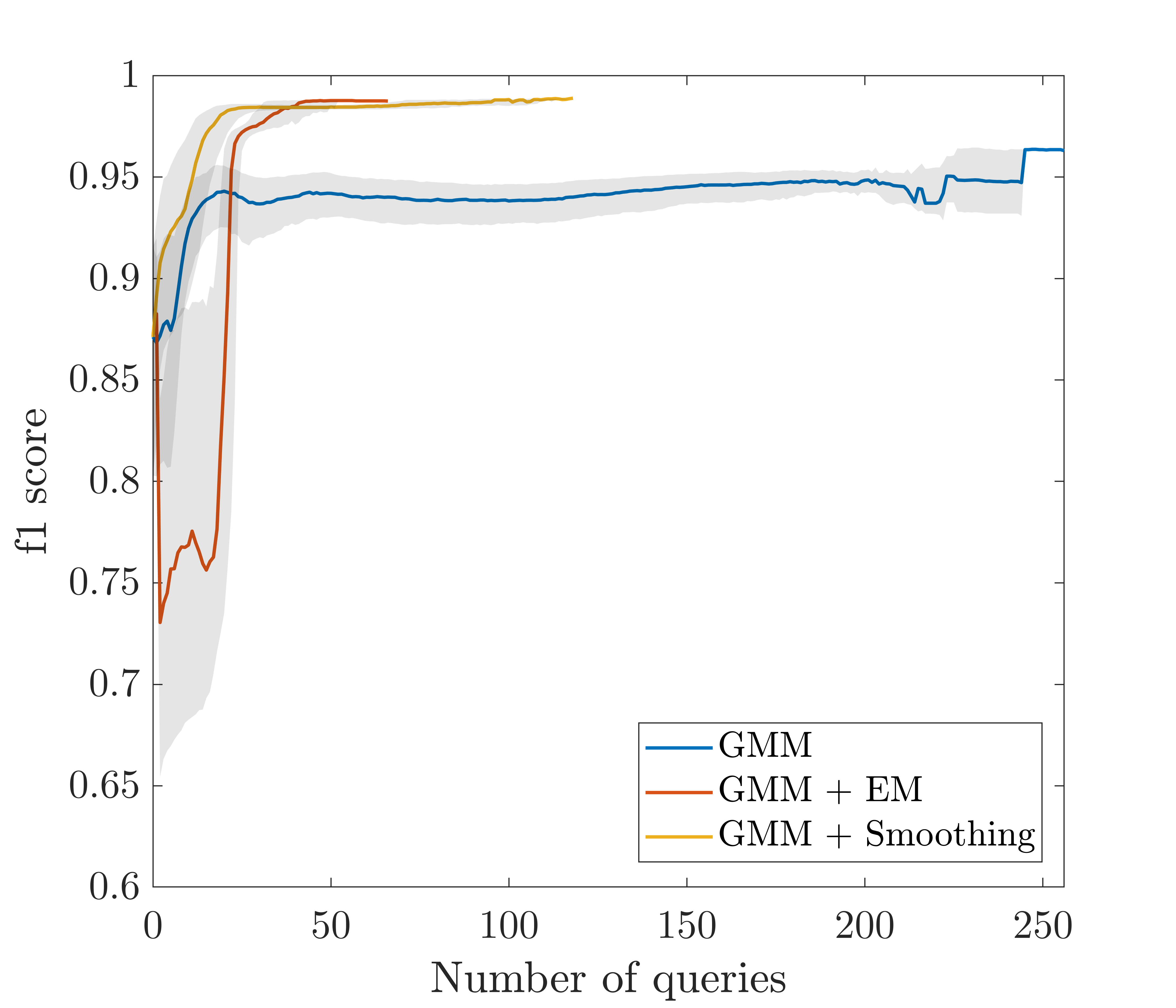}
		}
		\caption{ }
		\label{fig:f1score_ss}
	\end{subfigure}
	\caption{Variation in median (a) decision accuracy and (b) $f_{1}$ score with number of label queries for an agent utilising a GMM learning via risk-based active learning (i) without semi-supervised updates, (ii) with semi-supervised updates via EM and (iii) with semi-supervised updated via smoothing. Shaded area shows the interquartile range.}
	\label{fig:performance_ss}
\end{figure}

In summary, the case study has shown that, for some applications, semi-supervised learning provides a suitable approach to reducing the effects of sampling bias. The generative distributions obtained via these semi-supervised risk-based active learning approaches better fit the underlying data distributions, whilst also establishing well-defined decision boundaries in a cost-effective manner. The semi-supervised learning approach to reducing the negative impact of sampling bias on decision-making performance is re-examined with respect to an experimental dataset in Section \ref{sec:Z24_Results_ss}.

\subsection{Discriminative Classifiers}

Probabilistic discriminative models (sometimes referred to as conditional models) provide an alternative approach to generative models for developing statistical classifiers. Whereas generative models seek to first represent the underlying joint probability distribution $p(\mathbf{x},y)$, discriminative models seek to learn the predictive conditional distribution $p(y | \mathbf{x})$ directly. For discriminative classifiers, the mapping $f: X \rightarrow Y$ is typically specified via boundaries that partition the feature space $X$ according to $Y$. In probabilistic discriminative models, these classification boundaries are `soft', allowing for nondeterministic classification. For example, data lying on a binary classification boundary will be attributed equal probabilities of belonging to each class. Unlike with generative models, discriminative models do not require assumptions to be made regarding the underlying distribution of data (e.g.\ it was assumed earlier that data were Gaussian distributed), as only separating boundaries are learned. Thus, discriminative models do not necessarily require training datasets to be representative of the base distribution -- rather, they rely on data being representative of the classification boundaries. Because of this characteristic, and considering the proximity of queries to decision-boundaries, it is believed that discriminative classifiers may provide improved robustness to the type of sampling bias prevalent in risk-based active learning.

There are numerous approaches to probabilistic discriminative modelling. Arguably, \textit{logistic regression} is the simplest example of such a model. For logistic regression, a separating plane (or hyperplane) between two classes is learned. A sigmoid function then maps the distance of a data point from the plane to the probability of class membership. Support vector machines (SVMs) are also a popular formulation of discriminative classifier that have found application in many domains including engineering \cite{Cortes1995}. SVMs seek to find (hyper)planes with maximal margins between classes such that classification errors in the training dataset are minimised; a principal in statistical learning theory known as \textit{empirical risk minimisation}\footnote{Here, `risk' is distinct from the definition used elsewhere in the current paper. A mapping between the two could be formulated by considering the costs associated with misclassifications, however, this is outside the scope of the current paper.}. Furthermore, SVMs are capable of determining nonlinear classification boundaries by utilising nonlinear kernels to project data into higher dimensions in which the data are more separable. Fundamentally, SVMs are deterministic classifiers; however, the outputs can be modified to have a probabilistic interpretation via post-processing in the form of Platt scaling \cite{Platt1999}.

For its robust uncertainty quantification, a Bayesian extension of the SVM, known as the \textit{relevance vector machine} (RVM) is selected as the probabilistic discriminative classifier for the current paper. Additionally, as RVMs are sparse models utilising a small subset of training data, they potentially have enhanced robustness to sampling bias as problematic data can be ignored. Details of RVMs and their multiclass extension (mRVMs) are provided in the following subsection.

\subsubsection{Relevance Vector Machines}\label{sec:RVMs}

Originally introduced by Tipping in \cite{Tipping2001}, the RVM is a computationally-efficient Bayesian model capable of achieving high accuracies for both regression and classification tasks via the use of basis functions specified via a kernel function and a sparse subset of the training data. In \cite{Damoulas2008,Damoulas2009}, the RVM is extended from binary classification to multiclass classification via the multinomial probit link and multinomial probit likelihood. As previously mentioned, RVMs are able to achieve high computational efficiency. This is achieved by using a subset of $n^{\ast}$ `relevant' samples from $\mathcal{D}_l$ such that $n^{\ast} << n$. This reduced subset is denoted $\mathcal{A}$. Two approaches for achieving this sparsity are presented in \cite{Psorakis2010} and termed \mRVMa{} and \mRVMb{}. Details of the mRVM, and specifically the formulations presented in \cite{Psorakis2010}, are provided in Appendix \ref{app:mRVMs}.

Succinctly, \mRVMa{} utilises a constructive approach to build $\mathcal{A}$. $\mathcal{A}$ is initialised as an empty set with samples subsequently added or removed from $\mathcal{A}$ based upon their contribution to the objective function given by the decomposed form of the marginal log-likelihood detailed in \cite{Psorakis2010}. Whereas \mRVMa{} provides a constructive approach to the formation of $\mathcal{A}$ of relevance vectors, the approach termed \mRVMb{} sculpts $\mathcal{A}$ from $\mathcal{D}_l$ by iteratively discarding samples with weights approximately equal to zero for all classes. Here, it is worth noting that, within the \mRVMb{} approach, once a sample has been removed from $\mathcal{A}$ it cannot be reintroduced to the model. This characteristic is in contrast to \mRVMa{}, which allows a previously pruned sample to be reintroduced if, during a later iteration, the sample is deemed to have positive contribution. In the context of risk-based active learning, this constraint of \mRVMb{} should prove to be insignificant as, within the active learning process, the model is repeatedly retrained, with all data in $\mathcal{D}_l$ considered.

By using the sparse subset $\mathcal{A}$ to form the basis functions for the mRVM classification model, improvements in computational efficiency are achieved. Furthermore, in the context of risk-based active learning, it is hypothesised that probabilistic discriminative classifiers will show robustness to sampling bias over the generative classifiers -- in part because detrimental or superfluous data are excluded from the model by virtue of its sparsity, and because discriminative classifiers in general, do not rely upon assumptions regarding the underlying distribution of the data.

\subsubsection{Results}

Both \mRVMa{} and \mRVMb{} were incorporated into the risk-based active-learning process and applied to the case study presented in Section {\ref{sec:Results1}. As it is well-studied and flexible, both models were formed using a Gaussian kernel. The Gaussian kernel function has the following form,

\begin{equation}
	k(\mathbf{x}_t,\mathbf{x}_i) = \exp (-\gamma \parallel \mathbf{x}_t - \mathbf{x}_i \parallel ^{2})
\end{equation}

\noindent where $\gamma$ is introduced as a hyperparameter and is fixed at $\frac{1}{D}$ in accordance with \cite{Manocha2007,Psorakis2010}.

As with the previous statistical classifiers, the risk-based active-learning process was applied to each model 1000 times with randomly-selected initial labelled datasets $\mathcal{D}_l$.

Figures \ref{fig:initModel_mRVM1} and \ref{fig:finalModel_mRVM1} show, from one of the 1000 repetitions, an \mRVMa{} classifier before and after the risk-based active learning process, respectively.

Figure \ref{fig:initModel_mRVM1} shows that the \mRVMa{} model, subject to the convergence criteria presented in \cite{Psorakis2010}, was unable to effectively discriminate between classes when learned from the initial limited subset of training data. In Figure \ref{fig:initScatter_mRVM1}, the absence of contours corresponding to $p(y_t = k|\mathbf{x}_t) = 0.5$ indicates that, if using the initial model shown, one would be unable to classify data with a high degree of confidence. This result is corroborated by the approximately uniform EVPI over the feature space, presented in Figure \ref{fig:initVOI_mRVM1}. 

Figure \ref{fig:finalModel_mRVM1} indicates that, following the risk-based active learning procedure, the updated \mRVMa{} classifier has established classification boundaries -- the contours corresponding to $p(y_t = k|\mathbf{x}_t) = 0.5$ are now visible and, in general, are a good fit to the data. In Figure \ref{fig:finalScatter_mRVM1}, similar to the GMM, the risk-based active learning algorithm preferentially queries data close to the boundary between Class 3 (significant damage) and Class 4 (critical damage). From Figure \ref{fig:finalVOI_mRVM1}, it can be seen that a band of high EVPI has been established close to the classification boundary between classes 3 and 4. Again, this band corresponds to a decision boundary and indicates the region of the feature space where a decision-making agent should inspect the structure; below this band the agent can be confident that the optimal decision is `do nothing', and above this band the agent can be confident that the optimal decision is `perform maintenance'.

\begin{figure}[ht!]
	\begin{subfigure}{.5\textwidth}
		\centering
		\scalebox{0.4}{
			\includegraphics{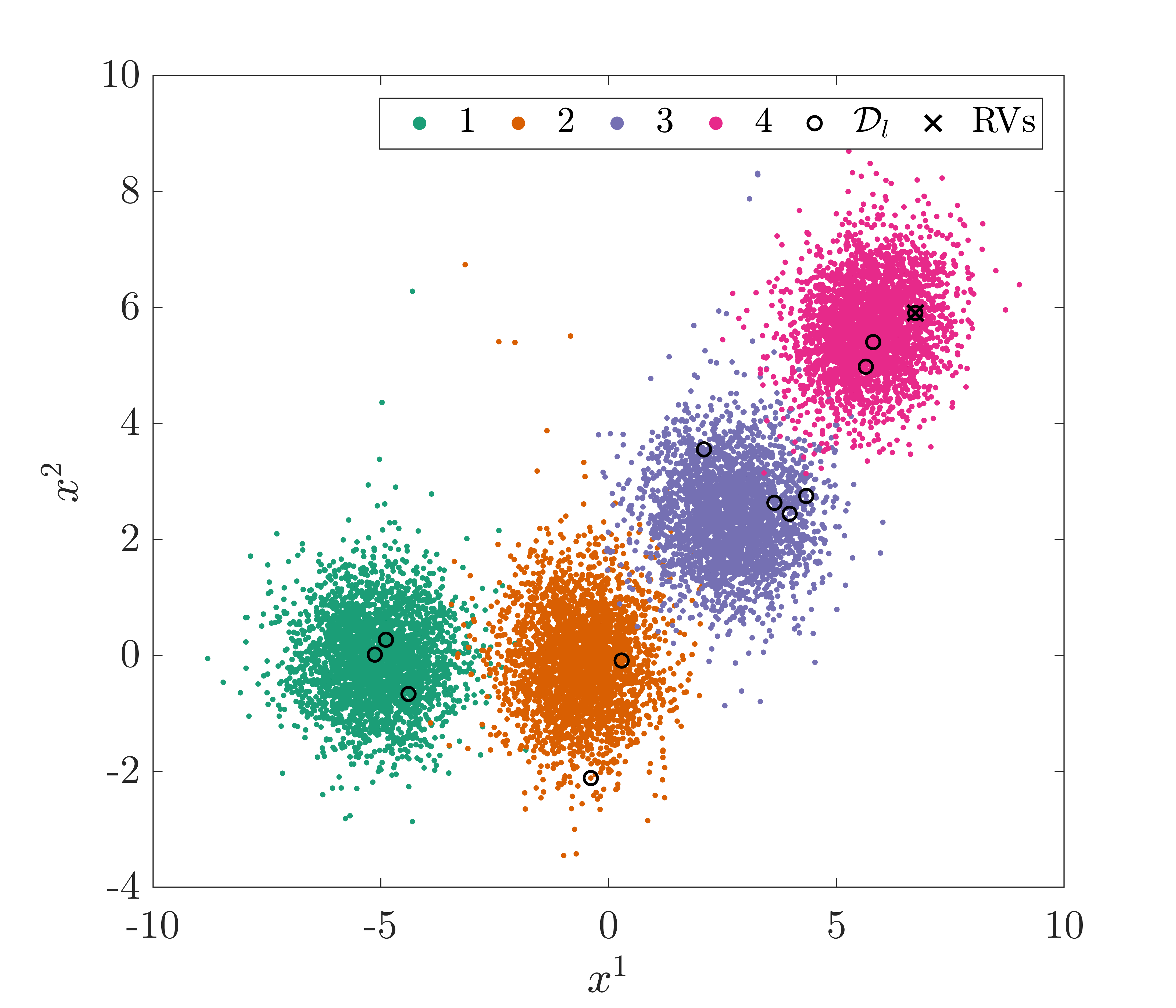}
		}
		\caption{ }
		\label{fig:initScatter_mRVM1}
	\end{subfigure}
	\begin{subfigure}{.5\textwidth}
		\centering
		\scalebox{0.4}{
			\includegraphics{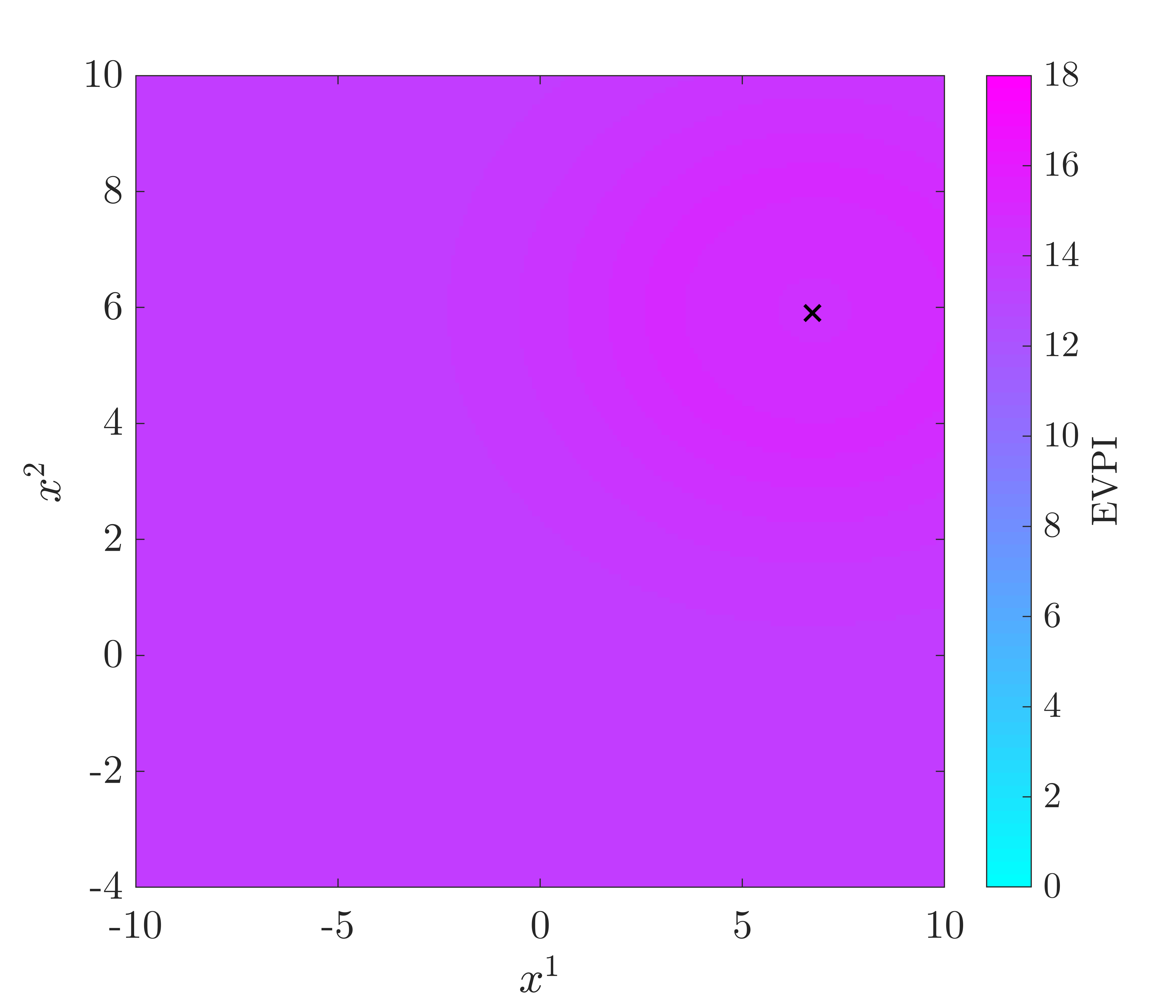}
		}
		\caption{ }
		\label{fig:initVOI_mRVM1}
	\end{subfigure}
	\caption{An mRVM$_1$ statistical classifier $p(y_t|\mathbf{x}_t;\mathbf{W})$ prior to risk-based active learning; relevance vectors are shown ($\times$). (a) shows the final model overlaid onto the data with labelled data $\mathcal{D}_l$ encircled and (b) shows the resulting EVPI over the feature space.}
	\label{fig:initModel_mRVM1}
\end{figure}

\begin{figure}[ht!]
	\begin{subfigure}{.5\textwidth}
		\centering
		\scalebox{0.4}{
			\includegraphics{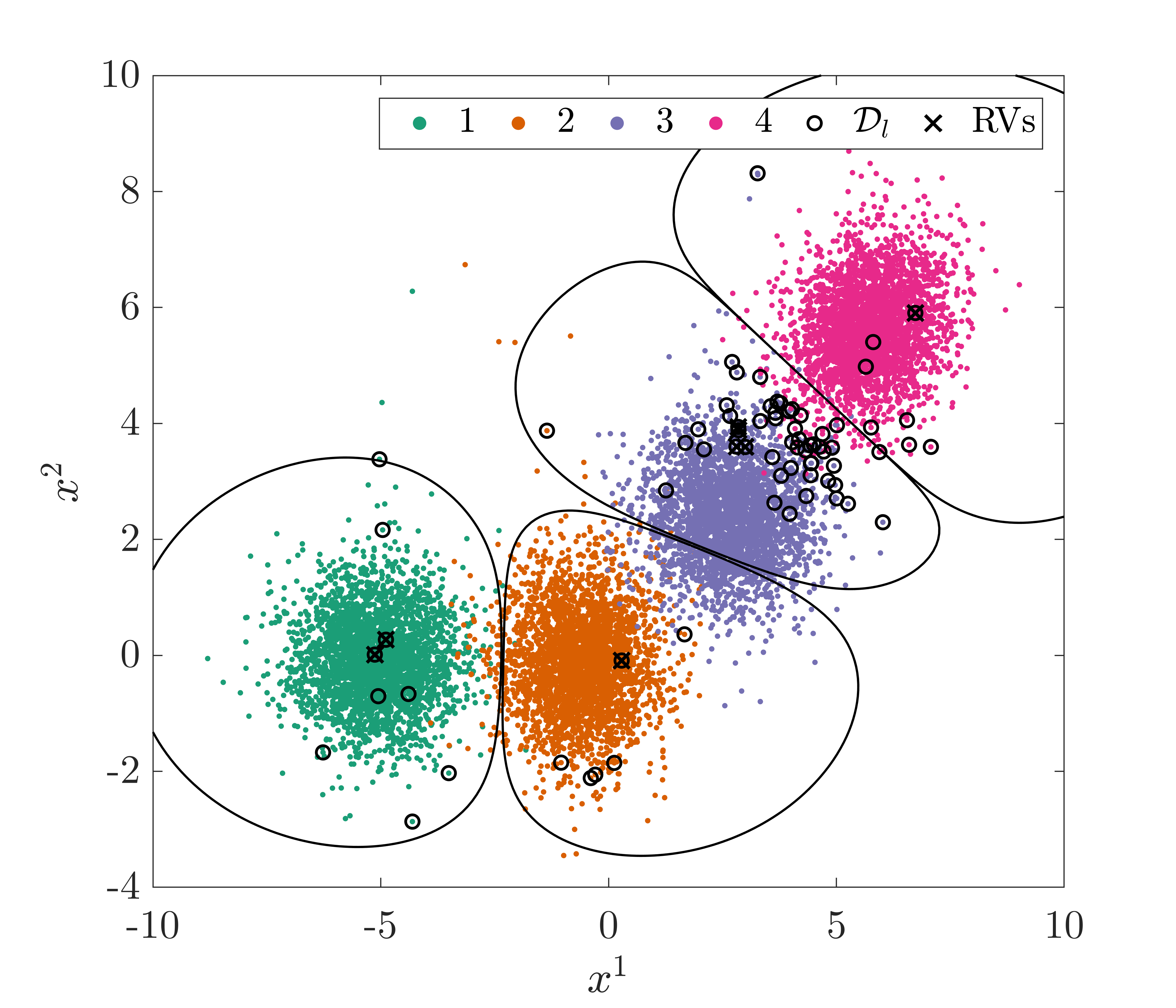}
		}
		\caption{ }
		\label{fig:finalScatter_mRVM1}
	\end{subfigure}
	\begin{subfigure}{.5\textwidth}
		\centering
		\scalebox{0.4}{
			\includegraphics{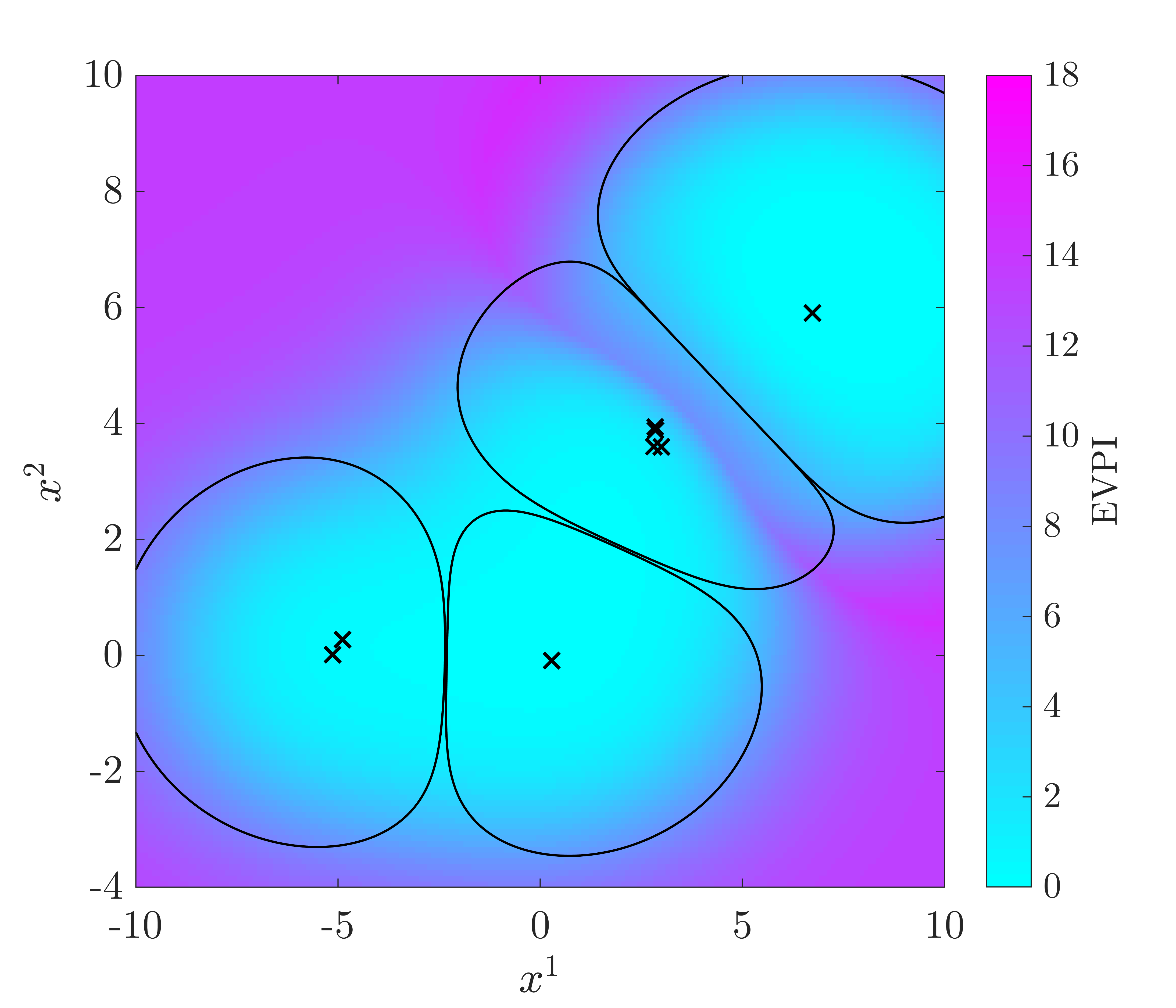}
		}
		\caption{ }
		\label{fig:finalVOI_mRVM1}
	\end{subfigure}
	\caption{An mRVM$_1$ statistical classifier $p(y_t|\mathbf{x}_t;\mathbf{W})$ after risk-based active learning; relevance vectors ($\times$) and contours (lines) denoting $p(y_t=k|\mathbf{x}_t)=0.5$ are shown. (a) shows the final model overlaid onto the data with labelled data $\mathcal{D}_l$ encircled and (b) shows the resulting EVPI over the feature space.}
	\label{fig:finalModel_mRVM1}
\end{figure}

Similar to Figures \ref{fig:initModel_mRVM1} and \ref{fig:finalModel_mRVM1}, Figures \ref{fig:initModel_mRVM2} and \ref{fig:finalModel_mRVM2} show, from one of the 1000 repetitions, an \mRVMb{} classifier before and after the risk-based active-learning process.

Unlike \mRVMa{}, the \mRVMb{} approach to relevance vector selection resulted in an initial model capable of discriminating between the four classes -- as demonstrated by the contours corresponding to $p(y_t = k|\mathbf{x}_t) = 0.5$ visible in Figure \ref{fig:initModel_mRVM2}. In general, the discriminative boundaries fit the data well, except for that corresponding to Class 2 (minor damage). The resulting EVPI distribution over the feature space, shown in Figure \ref{fig:initVOI_mRVM2}, indicates that a decision boundary (albeit somewhat nebulous) has been established, even from the very limited labelled data. 

\begin{figure}[ht!]
	\begin{subfigure}{.5\textwidth}
		\centering
		\scalebox{0.4}{
			\includegraphics{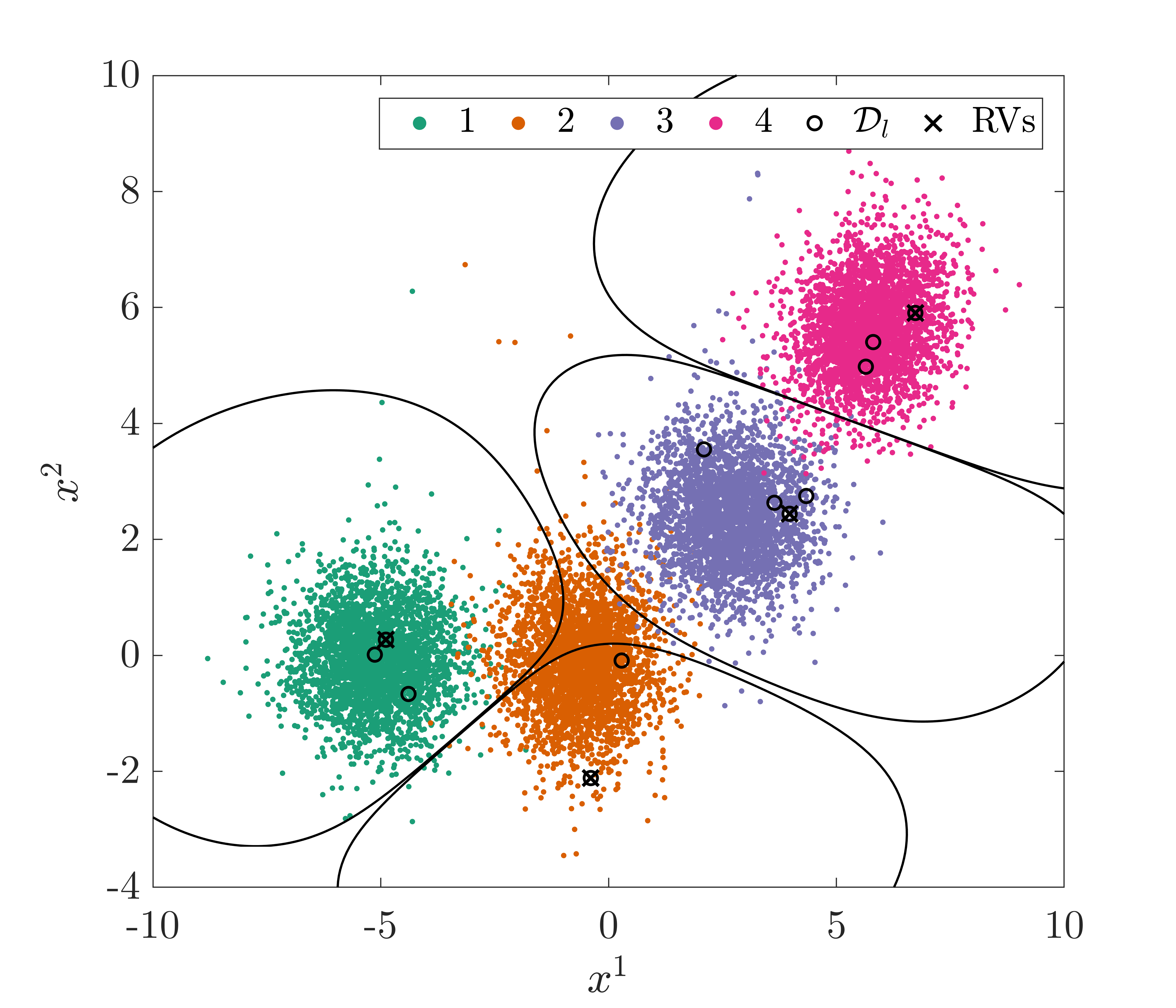}
		}
		\caption{ }
		\label{fig:initScatter_mRVM2}
	\end{subfigure}
	\begin{subfigure}{.5\textwidth}
		\centering
		\scalebox{0.4}{
			\includegraphics{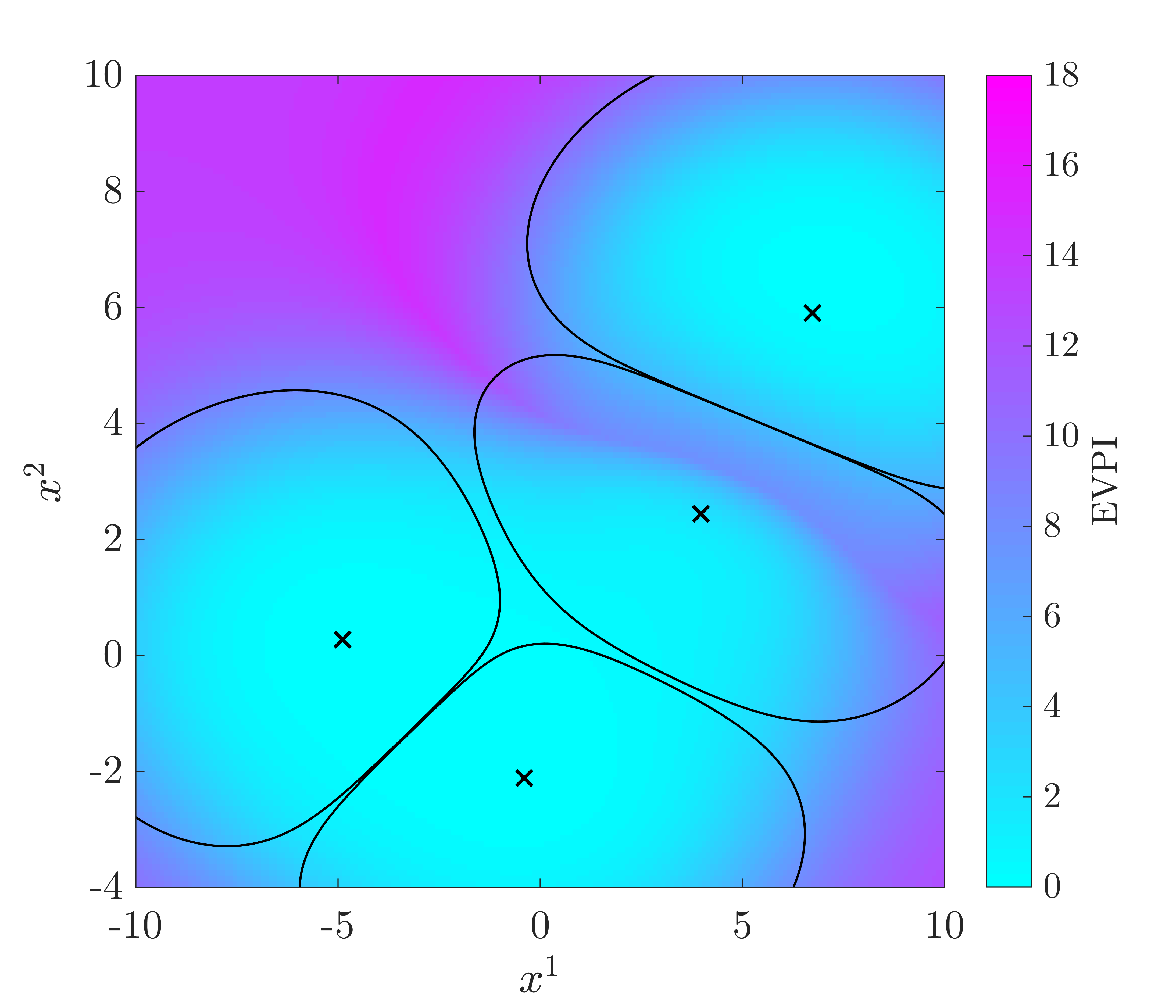}
		}
		\caption{ }
		\label{fig:initVOI_mRVM2}
	\end{subfigure}
	\caption{An mRVM$_2$ statistical classifier $p(y_t|\mathbf{x}_t;\mathbf{W})$ prior to risk-based active learning; relevance vectors ($\times$) and contours (lines) denoting $p(y_t=k|\mathbf{x}_t)=0.5$ are shown. (a) shows the final model overlaid onto the data with labelled data $\mathcal{D}_l$ encircled and (b) shows the resulting EVPI over the feature space.}
	\label{fig:initModel_mRVM2}
\end{figure}

\begin{figure}[ht!]
	\begin{subfigure}{.5\textwidth}
		\centering
		\scalebox{0.4}{
			\includegraphics{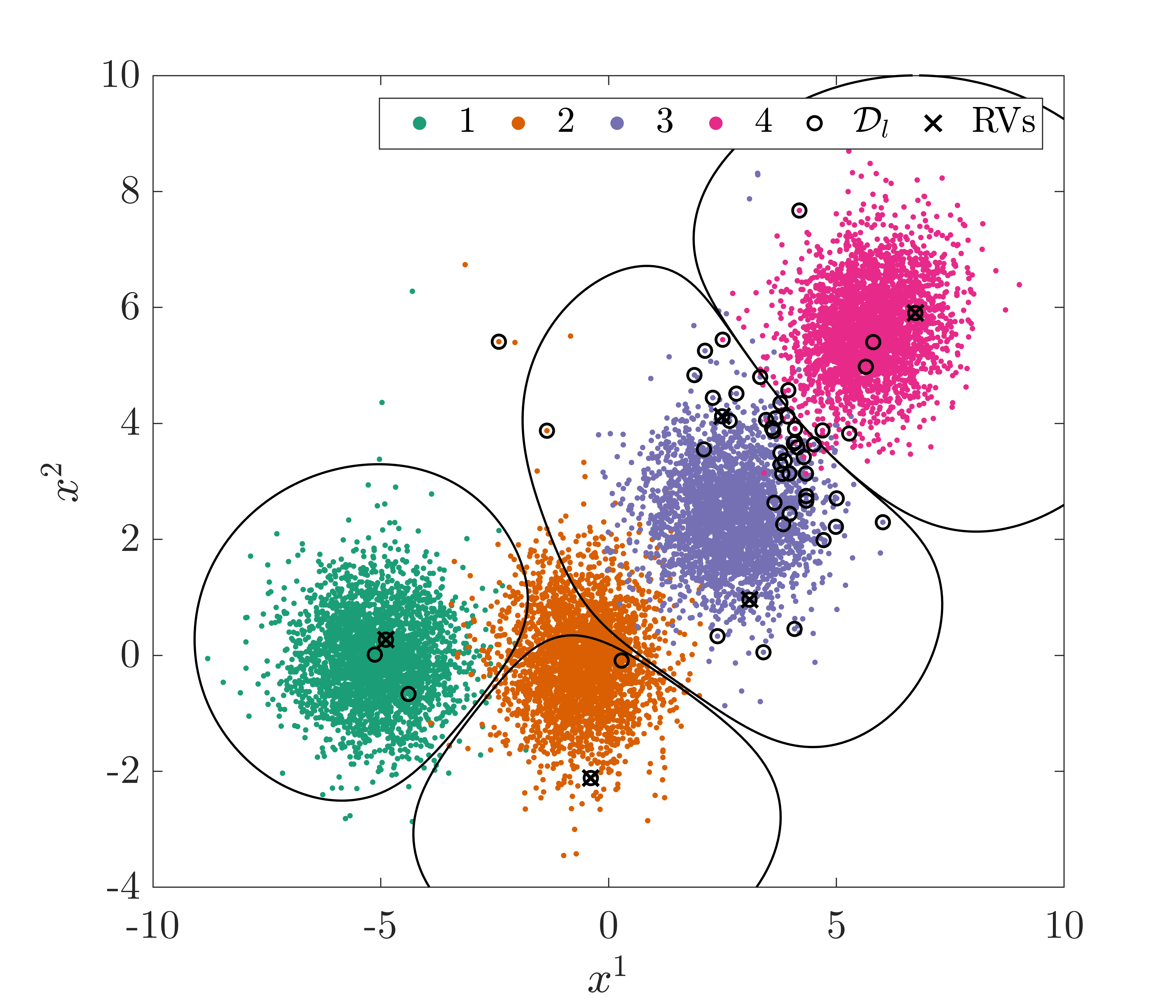}
		}
		\caption{ }
		\label{fig:finalScatter_mRVM2}
	\end{subfigure}
	\begin{subfigure}{.5\textwidth}
		\centering
		\scalebox{0.4}{
			\includegraphics{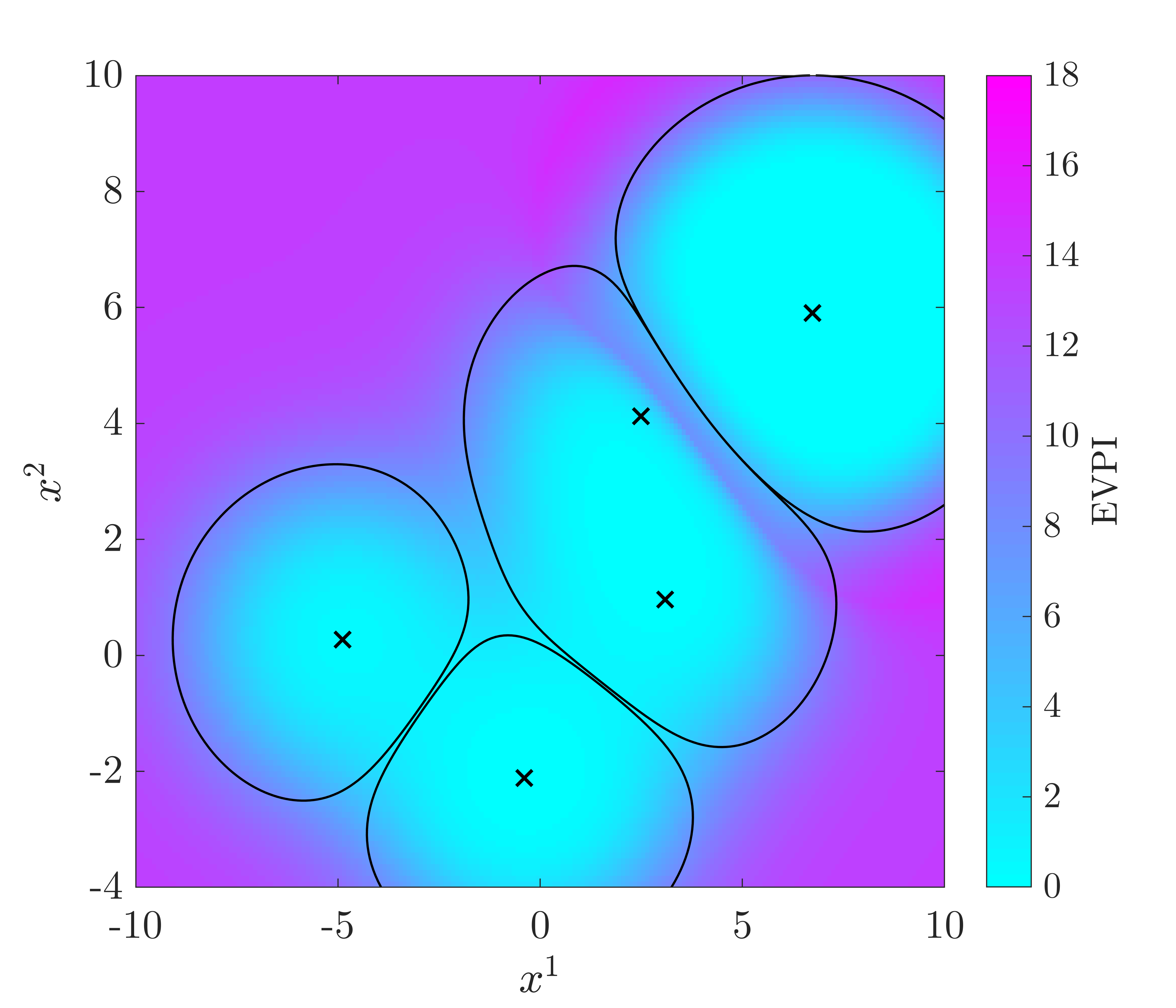}
		}
		\caption{ }
		\label{fig:finalVOI_mRVM2}
	\end{subfigure}
	\caption{An mRVM$_2$ statistical classifier $p(y_t|\mathbf{x}_t;\mathbf{W})$ after risk-based active learning; relevance vectors ($\times$) and contours (lines) denoting $p(y_t=k|\mathbf{x}_t)=0.5$ are shown. (a) shows the final model overlaid onto the data with labelled data $\mathcal{D}_l$ encircled and (b) shows the resulting EVPI over the feature space.}
	\label{fig:finalModel_mRVM2}
\end{figure}

It can be seen from Figure \ref{fig:finalScatter_mRVM2} that, once again, the active learning algorithm results in data being queried preferentially close to the boundary between classes 3 and 4. It can be seen from the probability contours that the updated model fits the data marginally better than the initial model. From Figure \ref{fig:finalVOI_mRVM2}, it can be seen that the inferred decision boundary is now more well-defined, represented by the narrow band of high EVPI close to the classification boundary between classes 3 and 4.

From the EVPI surfaces shown in Figures \ref{fig:initVOI_mRVM1}, \ref{fig:finalVOI_mRVM1}, \ref{fig:initVOI_mRVM2}, and \ref{fig:finalVOI_mRVM2}, it is apparent that, when using an mRVM model, data far from those already observed is classified with high uncertainty, meaning that high EVPI would be associated with any outlying observations. This result is in contrast to EVPI surfaces obtained when utilising generative models, where, as discussed in Section \ref{sec:FurtherComments1}, sampling bias resulted in over-confident predictions for outlying data. The decision support implications for the differing `attitudes' towards outliers, induced by the choice of a generative versus discriminative classifier, are discussed further in Section \ref{sec:Discussions}.

Figures \ref{fig:cprop_mRVM1} and \ref{fig:cprop_mRVM2} show the class proportions in the labelled dataset $\mathcal{D}_l$, averaged over the 1000 runs of risk-based active learning applied to \mRVMa{} and \mRVMb{}, respectively. Both figures show similar trends in class proportions as the original GMM, with Class 3 disproportionately represented ($\sim$60\%) in the final iteration of $\mathcal{D}_l$, above classes 1, 2, and 4. Here, it is worth recognising from Figures \ref{fig:finalModel_mRVM1} and \ref{fig:finalModel_mRVM2}, that within the sparse subset $\mathcal{A}$, the classes are more equally represented. For \mRVMa{} Class 1 corresponds to two of the eight relevance vectors, classes 2 and 4 each correspond to one, and Class 3 corresponds to 4 of the eight relevance vectors. Similarly, for \mRVMb{}, classes 1, 2 and 4 are each represented by 1 relevance vector, with Class 3 represented by two out of the five relevance vectors. This result suggests that utilising a sparse subset of data somewhat mitigates the prevalence of sampling bias in the training data.

\begin{figure}[ht!]
	\begin{subfigure}{.5\textwidth}
		\centering
		\scalebox{0.4}{
			\includegraphics{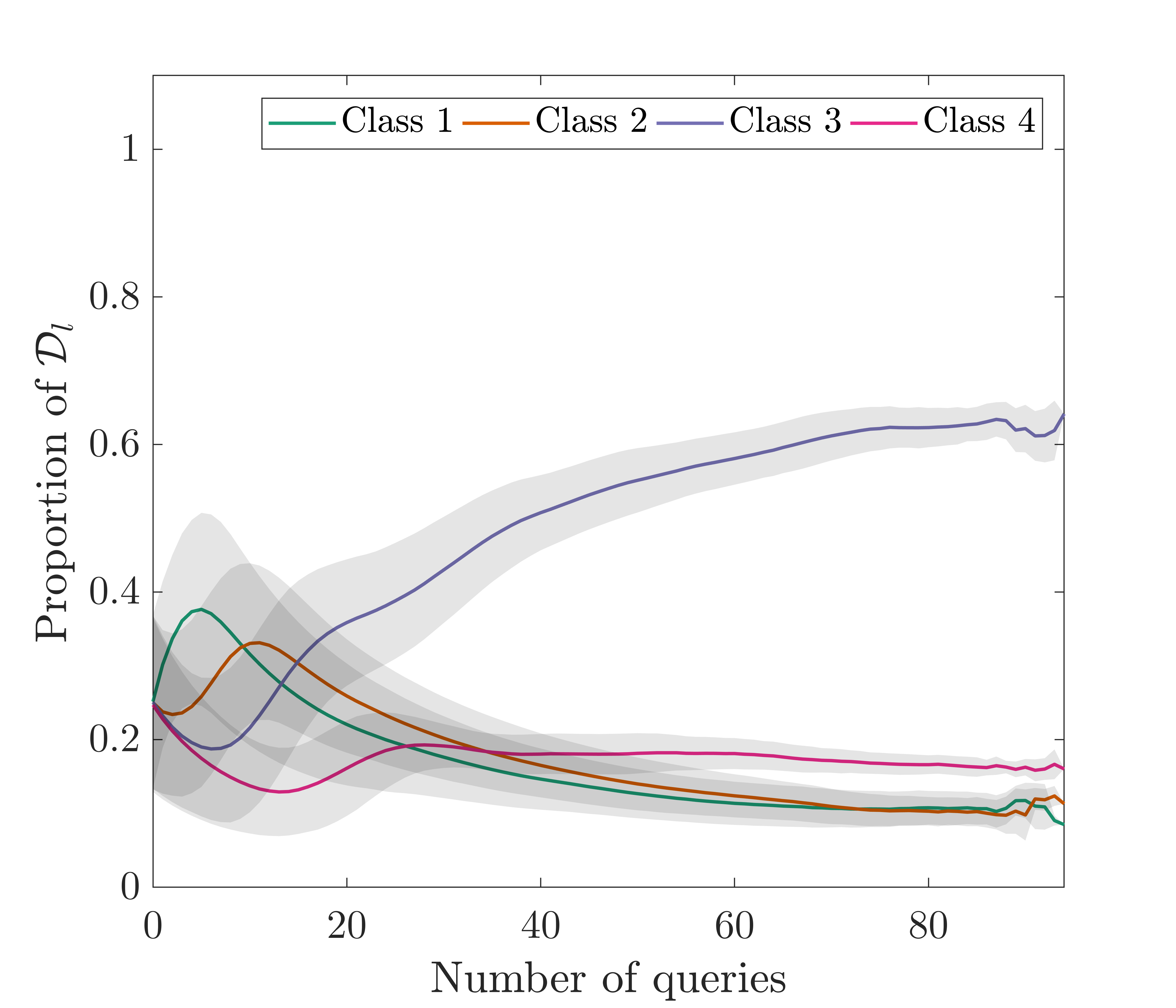}
		}
		\caption{ }
		\label{fig:cprop_mRVM1}
	\end{subfigure}
	\begin{subfigure}{.5\textwidth}
		\centering
		\scalebox{0.4}{
			\includegraphics{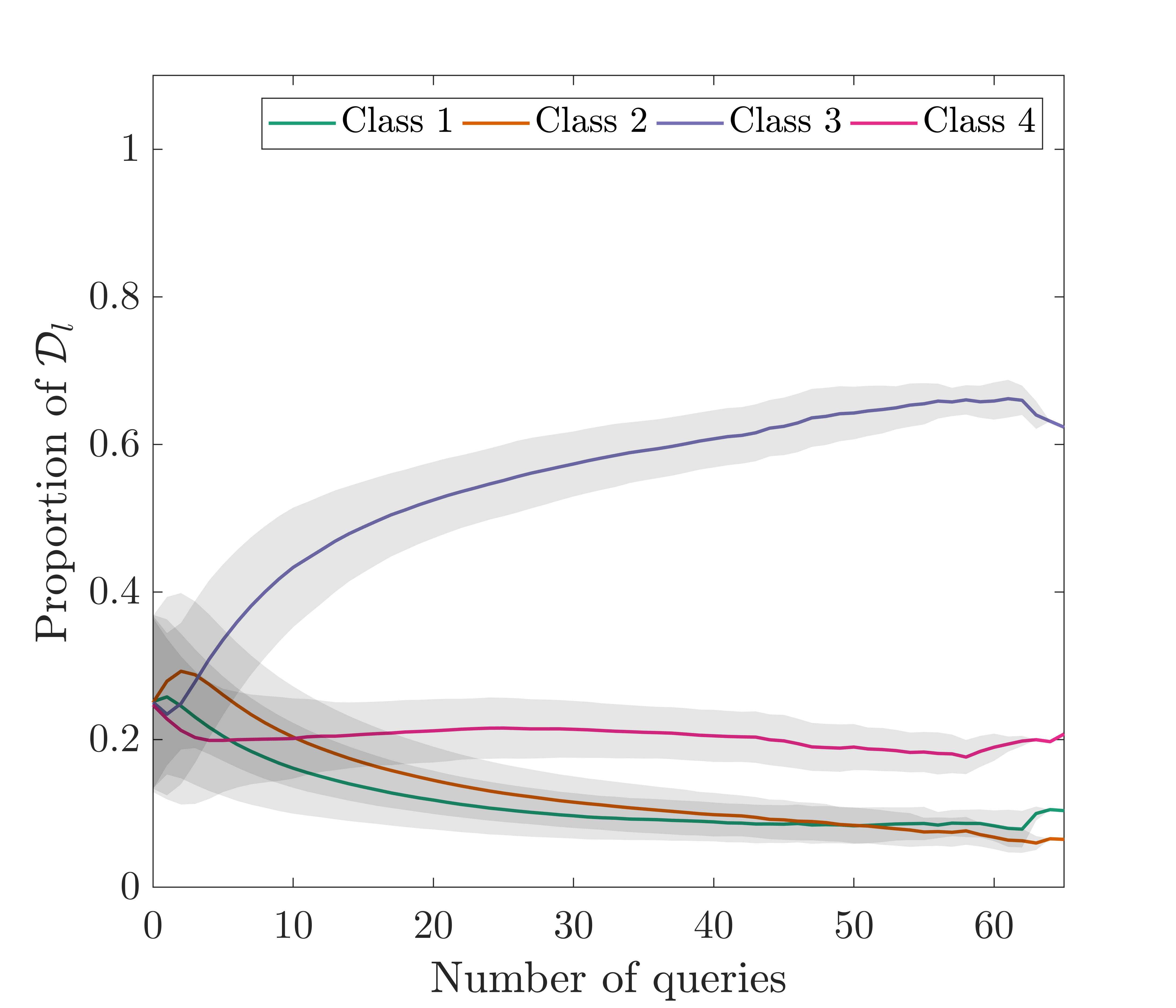}
		}
		\caption{ }
		\label{fig:cprop_mRVM2}
	\end{subfigure}
	\caption{Variation in class proportions within $\mathcal{D}_l$ with number of label queries for an agent utilising an (a) mRVM$_1$ and (b) mRVM$_2$ statistical classifier learned from $\mathcal{D}_l$ extended via risk-based active learning. Shaded area shows $\pm1\sigma$.}
	\label{fig:cprop_rvm}
\end{figure}

The histograms presented in Figure \ref{fig:hist_rvm} show the distributions of the number of queries made by the risk-based active learning algorithm as applied to \mRVMa{} and \mRVMb{} for each of the 1000 runs. Both \mRVMa{} and \mRVMb{} result in fewer queries than the GMM. Again, this result is significant, as it indicates that expenditure on queries has been reduced. Furthermore, the variance of the distributions corresponding to the mRVM algorithms is reduced compared to that for the GMM, indicating that relatively few queries are made, consistently.

\begin{figure}[ht!]
	\centering
		\scalebox{0.4}{
			\includegraphics{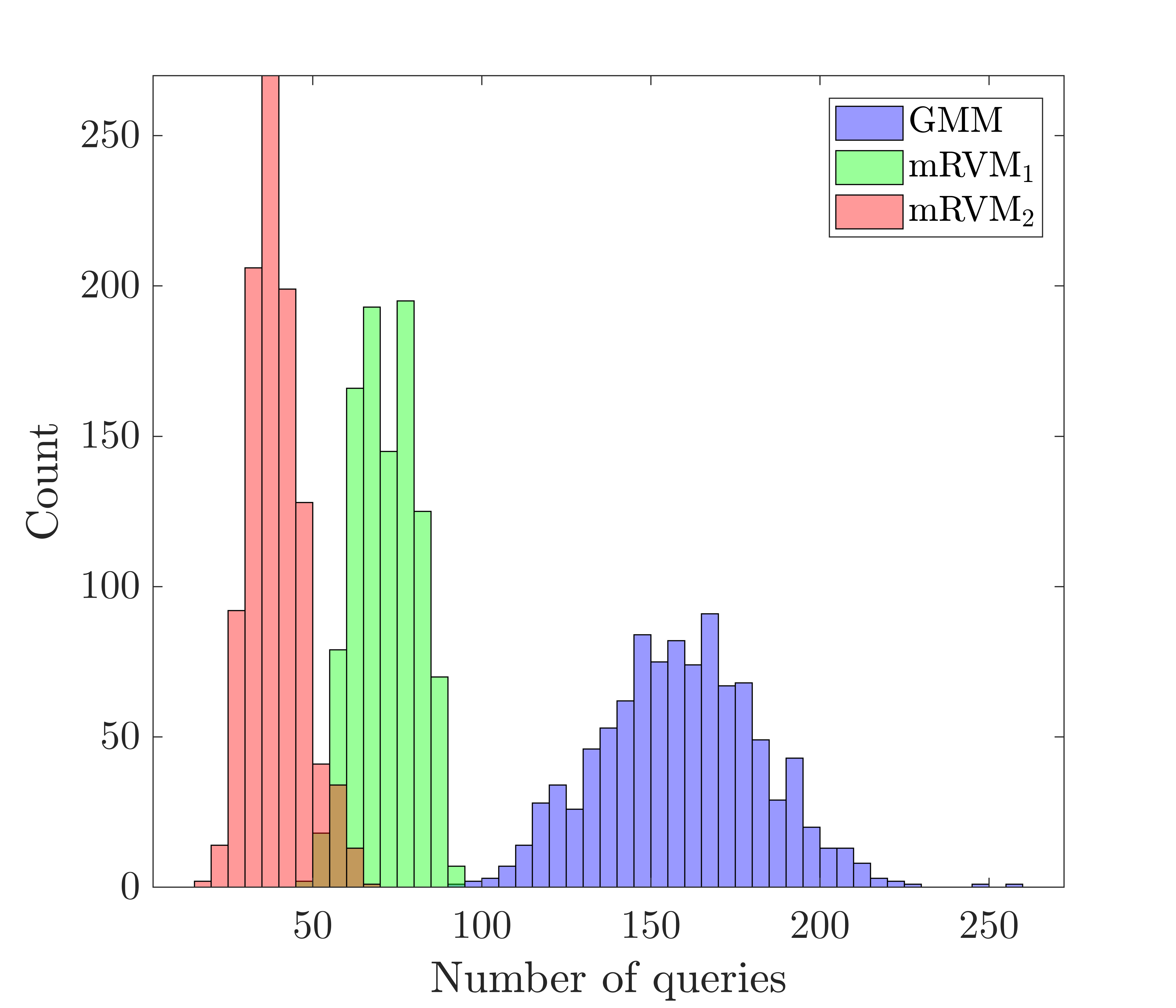}
		}
		\caption{Histograms showing the distribution of the number of queries from 1000 runs of the risk-based active learning of (i) a GMM (blue) (ii) an mRVM$_1$ (green) and (iii) an mRVM$_2$ (red) statistical classifier.}
		\label{fig:hist_rvm}
\end{figure}

\begin{figure}[ht!]
	\centering
		\scalebox{0.4}{
			\includegraphics{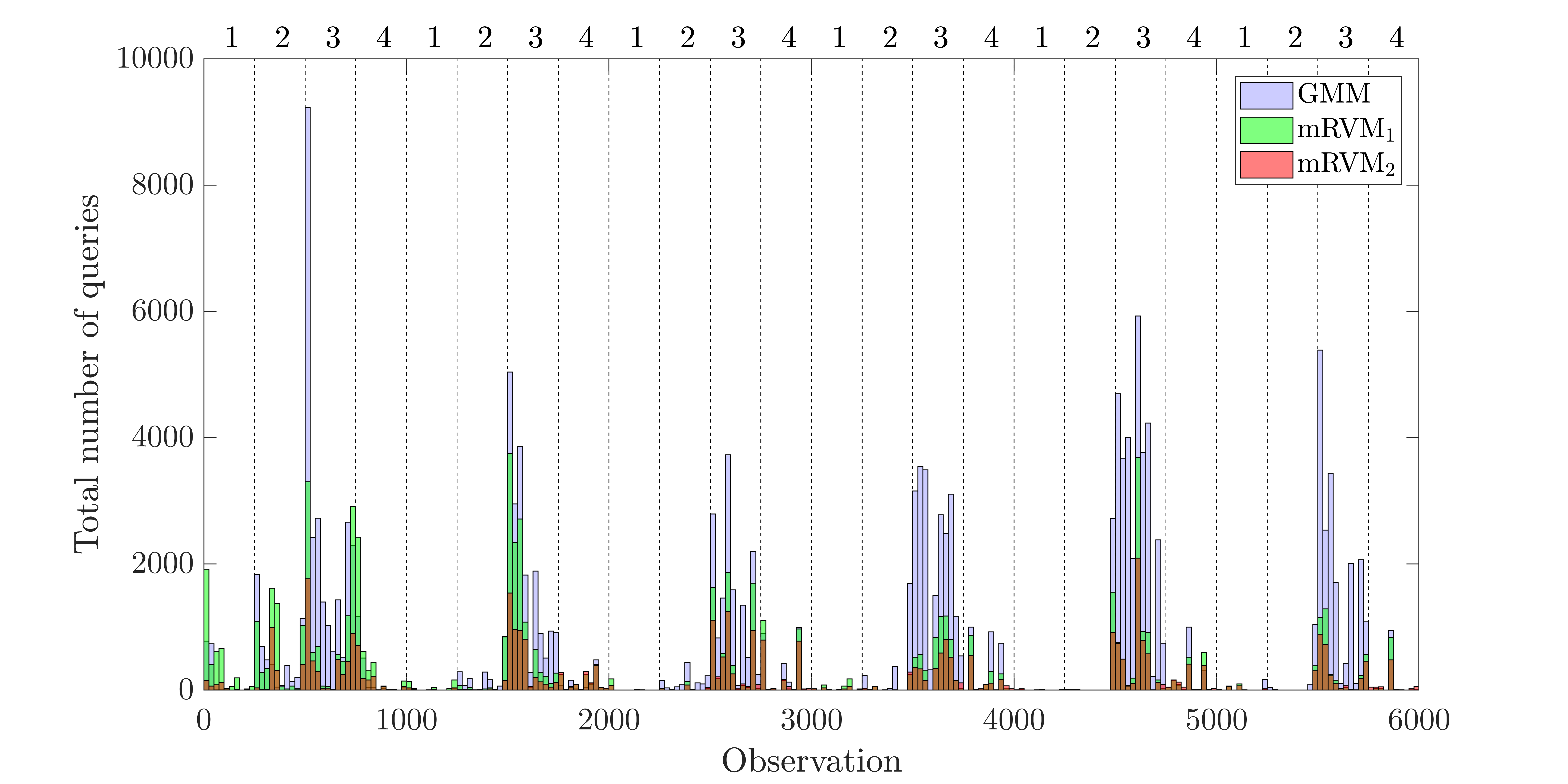}
		}
		\caption{Histograms showing the distribution of the number of queries for each observation in $\mathcal{D}_u$ from 1000 runs of risk-based active learning for (i) a GMM (blue) (ii) an mRVM$_1$ (green) and (iii) an mRVM$_2$ (red) statistical classifier. The average location of classes within $\mathcal{D}_u$ are numbered on the upper horizontal axis and transitions are denoted as a dashed line.}
		\label{fig:all_queries_rvm}
\end{figure}

\begin{figure}[ht!]
	\begin{subfigure}{.5\textwidth}
		\centering
		\scalebox{0.4}{
			\includegraphics{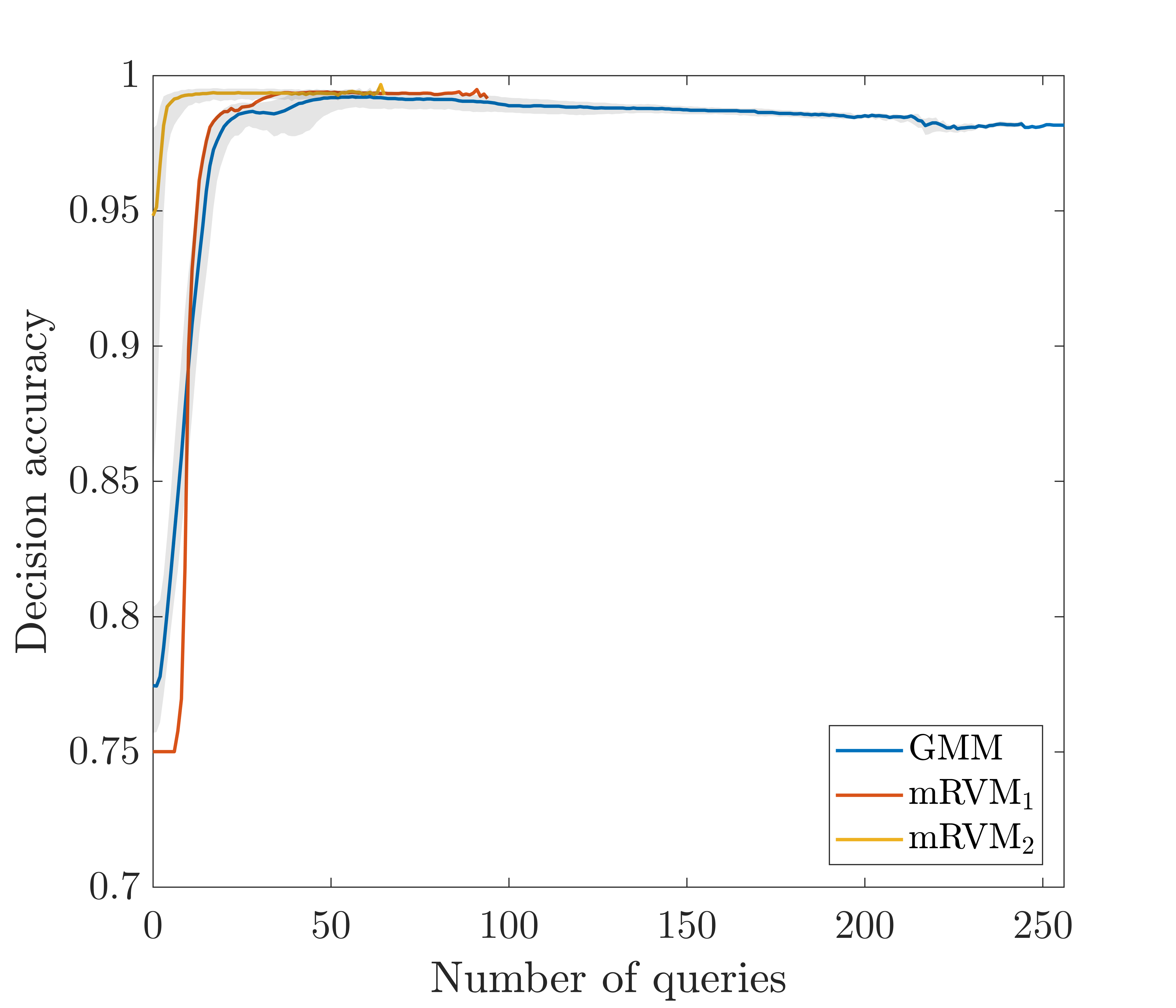}
		}
		\caption{ }
		\label{fig:dacc_rvm} 
	\end{subfigure}
	\begin{subfigure}{.5\textwidth}
		\centering
		\scalebox{0.4}{
			\includegraphics{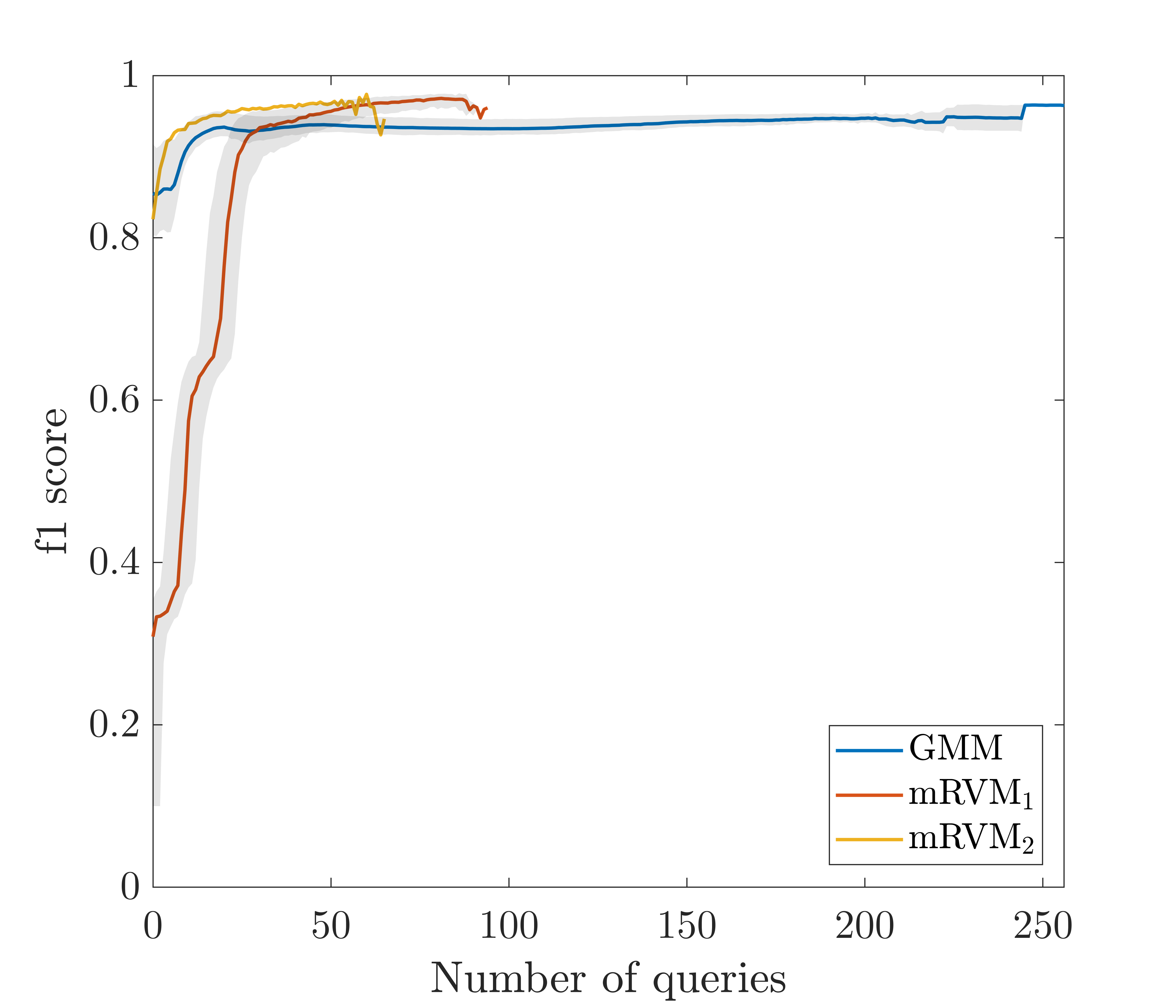}
		}
		\caption{ }
		\label{fig:f1score_rvm}
	\end{subfigure}
	\caption{Variation in median (a) decision accuracy and (b) $f_{1}$ score with number of label queries for an agent utilising (i) a GMM (ii) an mRVM$_1$ and (iii) an mRVM$_2$ statistical classifier learned via risk-based active learning. Shaded area shows the interquartile range.}
	\label{fig:performance_rvm}
\end{figure}

Figure \ref{fig:all_queries_rvm} compares the total number of queries for each index in $\mathcal{D}_u$ over the 1000 repetitions of risk-based active learning conducted using a GMM, an \mRVMa{} classifier, and an \mRVMb{} classifier. It can be seen from Figure \ref{fig:all_queries_rvm}, that data points queried by the GMM formulation also tend to be queries by the mRVM formulations, though with a greatly-reduced total. In contrast to the semi-supervised approaches presented in Section \ref{sec:results_ss_ext}, the risk-based active learning algorithms built upon mRVMs query data more heavily from the later occurrences of classes 3 and 4. This observation can, in part, be attributed to the fact that the high-EVPI regions associated with the decision boundary inferred via the mRVM models are slightly more nebulous than those obtained via semi-supervised learning. In addition, when using the mRVM classifiers, one would expect outlying data to be queried at any time during the risk-based active learning process because of the uncertainty in the class-label prediction.

Figure \ref{fig:performance_rvm} provides a comparison of the median decision accuracies and $f_1$-scores throughout the querying processes aggregated from the 1000 runs of the risk-based active learning process, applied to the \mRVMa{} and \mRVMb{} classifiers. It can be seen from Figure \ref{fig:dacc_rvm}, that both \mRVMa{} and \mRVMb{} result in superior decision accuracy performance over the GMM, plateauing at a slightly higher decision accuracy and reaching that plateau earlier. Although \mRVMa{} begins with a lower decision accuracy than the GMM, the performance of the decision-maker rapidly improves with queries made. On the other hand, \mRVMb{} begins with significantly higher decision accuracy than both the GMM and the \mRVMa{} classifiers, yet still improves rapidly with the early queries. From Figure \ref{fig:dacc_rvm}, an important observation regarding the use of the discriminative models is that, for later queries, there is no decline in decision-making performance as additional labelled data are obtained, indicating that such models, as hypothesised, possess improved robustness to sampling bias. Figure \ref{fig:f1score_rvm} shows the $f_1$-score classification performance of the models on the test dataset. Both models achieve improved classification performance over the GMM.

In summary, for the current case study, substituting the generative GMM classifier out of the risk-based decision process, in favour of RVM-based discriminative models, has mitigated the adverse effects on decision-making performance caused by the inherent bias introduced via the risk-based active learning process. Furthermore, well-defined decision boundaries were inferred from a reduced number of queries as compared to the generative model. These results were achieved as discriminative models do not rely on prior assumptions regarding the distribution of data.

\section{Case Study: Z24 Bridge Dataset}\label{sec:Z24}

In the previous section, the effects of sampling bias on decision-making performance, and approaches for counteracting these effects, were demonstrated on a somewhat idealised case study in which data were Gaussian distributed and classes were fairly separable. In the current section, the effectiveness of semi-supervised learning, and discriminative classifiers, at mitigating the effects of sampling bias, are assessed using an experimental dataset obtained from the Z24 Bridge \cite{Maeck2003}.

The Z24 bridge was a concrete highway bridge in Switzerland, situated between the municipalities of Koppigen and Utzenstorf and near the town of Solothurn. The bridge was the subject of a cross-institutional research project (SIMCES). The purpose of this project was to generate a benchmark dataset and prove the feasibility of vibration-based SHM \cite{Maeck2001,DeRoeck2003}. Because of the presence of varying structural and environmental conditions, the benchmark dataset has seen wide use in research focussed on SHM and modal analysis.

Prior to its demolition, the Z24 bridge was instrumented with sensors, and for a period of 12 months, the dynamic response of the structure and the environmental conditions were monitored. The quantities measured included the accelerations at multiple locations, air temperature, deck temperature, and wind speed \cite{Peeters2001}.

Using the dynamic response data obtained during the monitoring campaign, the first four natural frequencies of the structure were obtained \cite{Maeck2001}. A visualisation of these natural frequencies can be seen in Figure \ref{fig:data_z24}. Figure \ref{fig:overview_z24} shows the four natural frequencies in discrete time, whereas Figure ~\ref{fig:scatter_z24} shows a two-dimensional projection of these frequencies obtained via principal component analysis. The first two principal components, shown in Figure \ref{fig:scatter_z24}, account for approximately 95\% of the variability in the dataset. The principal components are provided for visualisation purposes only. In the current paper, the analyses presented for the Z24 dataset use the first four natural frequencies as features.

In total, the dataset consists of 3932 observations. Towards the end of the monitoring campaign -- from observation 3476 onwards -- incremental damage was introduced to the structure. Additionally, throughout the campaign, the bridge experienced low temperatures, and as such, the structure exhibited cold temperature effects, particularly noticeable between observations 1200 and 1500. It is believed that the increased natural frequencies observed during these periods are a result of stiffening in the bridge deck induced by very low ambient temperatures. From these data, a four-class classification problem can be defined such that $y_t \in \{ 1,2,3,4 \}$:

\begin{itemize}
	\item Class 1: normal undamaged condition (green)
	\item Class 2: cold temperature undamaged condition (blue)
	\item Class 3: incipient damage condition (orange)
	\item Class 4: advanced damage condition (pink).
\end{itemize}

In accordance with \cite{Hughes2022}, it is assumed here that the data obtained after the introduction of damage can be separated into 2 halves; the earlier half corresponding to incipient damage and the later half corresponding to advanced damage. This assumption is deemed to be reasonable, given the incremental nature of the damage progression \cite{Maeck2003}.

\begin{figure}[ht!]
	\begin{subfigure}{.5\textwidth}
		\centering
		\scalebox{0.4}{
			\includegraphics{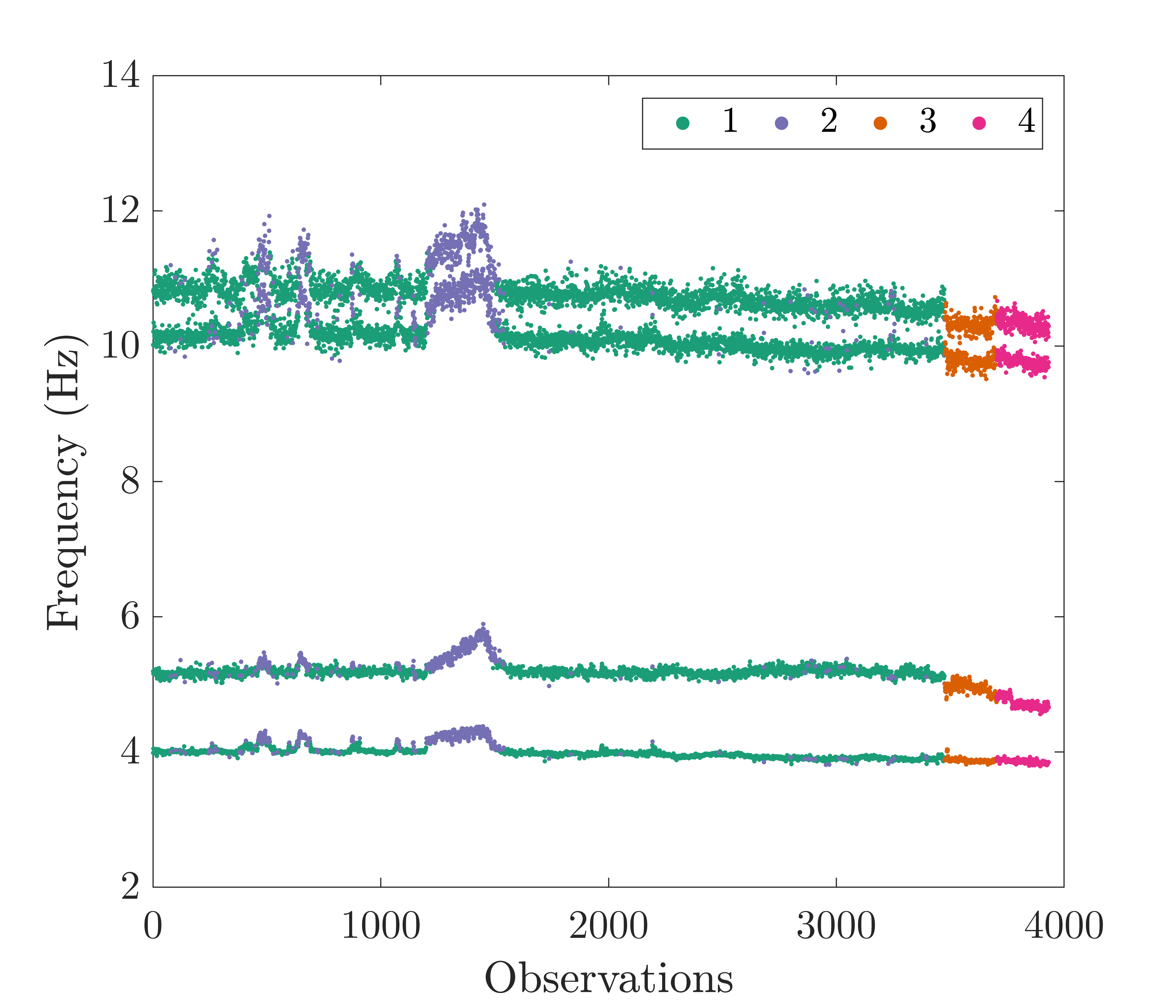}
		}
		\caption{ }
		\label{fig:overview_z24}
	\end{subfigure}
	\begin{subfigure}{.5\textwidth}
		\centering
		\scalebox{0.4}{
			\includegraphics{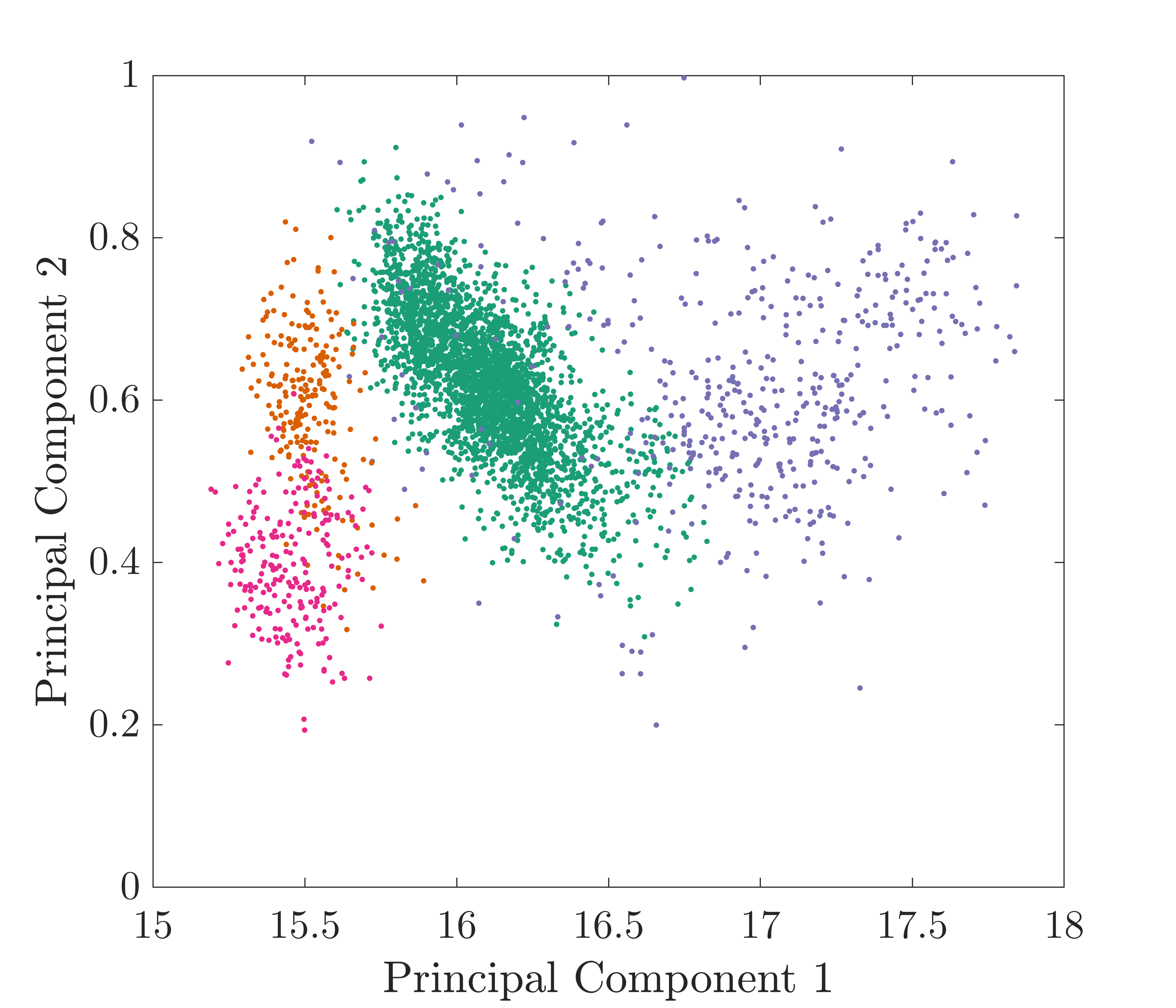}
		}
		\caption{ }
		\label{fig:scatter_z24}
	\end{subfigure}
	\caption{Visualisation of the Z24 Bridge dataset in (a) the discrete time and (b) the projected feature space.}
	\label{fig:data_z24}
\end{figure}

Upon examination of Figure \ref{fig:scatter_z24}, it becomes apparent that there is significant overlap between Class 1 and Class 2. There is also some degree of overlap between the data corresponding to classes 1 and 3, classes 2 and 3, and classes 3 and 4. Moreover, the distribution of Z24 Bridge data deviates somewhat from Gaussianity. Finally, it is obvious that class imbalance is present in the dataset, with most observations corresponding to Class 1. Combined, these characteristics of the Z24 Bridge dataset should represent a greater challenge to the risk-based active-learning approaches demonstrated on a more idealised case study in Section \ref{sec:Approaches}.

\subsection{Decision Process}

As with the previous case study, an O\&M decision process must be specified for the Z24 Bridge dataset in order to apply a risk-based active learning approach to classifier development. As the bridge was demolished over two decades ago at the time of writing, here, a simple binary decision process (`do nothing' or `perform maintenance'), similar to that detailed in Section \ref{sec:VisualExample}, is considered. Because of the similar nature of the decision processes considered, the graphical representation of the process provided by the influence diagram in Figure \ref{fig:OverallPGM2}, applies also to the current case study. Although, the influence diagram is consistent, the parameters populating the conditional probability distributions and utility functions must be specified to reflect the operational context of the Z24 Bridge. A discussion of the assumptions used to specify these distributions and utility functions is provided in \cite{Hughes2022}.

For brevity, it is assumed that utilities can be attributed directly to the four health states comprising the Z24 dataset, without the need for a fault tree. Here states 1 and 2 are assigned some small positive utility as reward for the bridge being functional, with minimal risk of failure. The incipient damage state is assigned a negative utility of moderate magnitude. This value is to reflect a possible reduction in operating capacity (e.g.\ bridge closure), and low-to-moderate risk of failure. Class 4, which corresponds to the advanced damage condition, is assigned a very large negative utility. This value reflects the severe consequences associated with the collapse of a bridge. The utility function $U(y_t)$ is provided in Table \ref{tab:Uy_z24}.

The utilities associated with the candidate actions in the domain of $d_t$ are given by the utility function $U(d_t)$ and are presented in Table \ref{tab:Ud_z24}. Once again, it is assumed that the action $d_t=0$ (`do nothing') has zero utility and that the action $d_t=1$ (`perform maintenance') has negative utility.

\begin{table}
	\begin{minipage}{.5\linewidth}
	  \centering
	  \caption{The utility function $U(y_{t+1})$.}
	  \label{tab:Uy_z24}   
	  \begin{tabular}{cc}
		  \toprule
		  \midrule
		  $y_{t+1}$ & $U(y_{t+1})$\\
		  \midrule
		  $1$ & $10$\\
		  $2$ & $10$\\
		  $3$ & $-50$\\
		  $4$ & $-1000$\\
		  \midrule
		  \bottomrule
	  \end{tabular}
	\end{minipage}%
	\begin{minipage}{.5\linewidth}
	  \centering
	  \caption{The utility function $U(d_t)$ where $d_t=0$ and $d_t=1$ denote the `do nothing' and `repair' actions, respectively.}
	\label{tab:Ud_z24}   
	\begin{tabular}{cc}
		\toprule
		\midrule
		$d_t$ & $U(d_t)$\\
		\midrule
		$0$ & $0$\\
		$1$ & $-100$\\
		\midrule
		\bottomrule
	\end{tabular}
	\end{minipage}
  \end{table}

Given the decided action $d_t = 0$, the health-state transitions are specified using the same assumptions as with the previous case study; namely, that the structure monotonically degrades and has a propensity to remain in the current state. Unlike the previous case study, however, states 1 and 2 both correspond to an undamaged condition of the bridge and instead reflect differing environmental conditions. As such, transitions both to and from these states must be possible according to the transition model. The earlier assumption that there is a propensity to remain in the current state is invoked once again; it is simply asserted that if there was a cold temperature for the most recent measurement, it is probable that the next measurement will also be made at a cold temperature. Likewise, the same reasoning is applied to the normal condition data. These assumptions are represented in the conditional probability distribution shown in Table \ref{tab:P_y1_y0_d0_z24}.

\begin{table}[ht!]
	\centering
	\caption{The conditional probability table $P(y_{t+1}|y_t, d_t)$ for $d_t = 0$.}
	\label{tab:P_y1_y0_d0_z24}       
\begin{tabular}{c c c c c c}
	\toprule
	\midrule
	& & \multicolumn{4}{c}{$y_{t+1}$}\\
	&& 1  & 2 & 3 & 4  \\ \cmidrule{3-6}
	\multicolumn{1}{c}{\multirow{4}{*}{\begin{sideways}\parbox{1.5cm}{\centering $y_t$}\end{sideways}}}   &
	\multicolumn{1}{l}{1}& 0.7 & 0.28 & 0.015 & 0.005 \\
	\multicolumn{1}{c}{}    &
	\multicolumn{1}{l}{2}& 0.43 & 0.55 & 0.15 & 0.05  \\
	\multicolumn{1}{c}{}    &
	\multicolumn{1}{l}{3} & 0 & 0 & 0.8 & 0.2  \\
	\multicolumn{1}{c}{}    &   
	\multicolumn{1}{l}{4} & 0 & 0 & 0 & 1  \\
	\midrule
	\bottomrule
\end{tabular}
\end{table}

The portion of the transition model for $d_t = 1$ is presented in Table \ref{tab:P_y1_y0_d1_z24}. Again, this distribution is specified via the assumption that the act of performing maintenance returns the bridge to an undamaged condition with high probability. Since there are two states corresponding to the undamaged condition of the bridge, the transition model must be specified such that it is ensured that the probability of returning to each of the undamaged condition states is independent of the action taken, i.e.\ it must be asserted that the weather condition is not influenced by the action taken. The procedure by which this constraint is imposed is outlined in \cite{Hughes2022}.

\begin{table}[ht!]
	\centering
	\caption{The conditional probability table $P(y_{t+1}|y_t, d_t)$ for $d_t = 1$.}
	\label{tab:P_y1_y0_d1_z24}       
	\begin{tabular}{c c c c c c}
		\toprule
		\midrule
		& & \multicolumn{4}{c}{$y_{t+1}$}\\
		&& 1  & 2 & 3 & 4  \\ \cmidrule{3-6}
		\multicolumn{1}{c}{\multirow{4}{*}{\begin{sideways}\parbox{1.5cm}{\centering $y_t$}\end{sideways}}}   &
		\multicolumn{1}{l}{1}& 0.7173 & 0.2857 & 0 & 0 \\
		\multicolumn{1}{c}{}    &
		\multicolumn{1}{l}{2}& 0.4388 & 0.5612 & 0 & 0  \\
		\multicolumn{1}{c}{}    &
		\multicolumn{1}{l}{3} & 0.5996 & 0.3904 & 0.01 & 0  \\
		\multicolumn{1}{c}{}    &   
		\multicolumn{1}{l}{4} & 0.5996 & 0.3904 & 0 & 0.01  \\
		\midrule
		\bottomrule
	\end{tabular}
\end{table}

Finally, the cost of inspection is specified to be $C_{\text{ins}} = 30$. The moderate cost is intended to reflect resources required to inspect a large-scale structure such as the Z24 Bridge.

\subsection{Results: Gaussian Mixture Model}

To facilitate comparison, risk-based active learning was applied to a Gaussian mixture model learned via the Bayesian approach outlined in Section \ref{sec:GMM}. For the Z24 Bridge case study, the features used to discriminate between the four health states of the structure were the first four natural frequencies shown in Figure \ref{fig:overview_z24}, such that $\mathbf{x}_t \in \mathbb{R}^4$.

The dataset was separated into two halves. One half formed the training dataset $\mathcal{D}$, with the other half forming an independent test set. A small (0.3\%) random subset of $\mathcal{D}$ was used to initialised the labelled dataset $\mathcal{D}_l$. The remaining data were assigned to $\mathcal{D}_u$ to be sequentially evaluated with respect to the decision process in the risk-based active-learning process. Once again, 1000 repetitions of the active learning process were conducted, each with randomly-initialised $\mathcal{D}_l$.

Figure \ref{fig:performance_z24} shows the median and interquartile range for the decision accuracy and $f_1$-score as a function of number of queries. Additionally, the performances measures are provided for an agent utilising a GMM trained using an equivalent number of samples obtained via unguided (random) querying. From Figure \ref{fig:dacc_z24}, it can be seen that risk-based active learning results in decision-making performance improving at an increased rate compared to random sampling. However, yet again, a gradual decline in decision accuracy can be observed over later queries when said queries are guided by EVPI. Figure \ref{fig:f1score_z24} shows that the classification performances of the predictive model are fairly similar for EVPI-based and random querying. Each gradually improves with number of queries, albeit non-monotonically. The large interquartile ranges for both approaches to querying indicate that the classification performance is highly sensitive to which data are available in the supervised learning process -- a phenomenon reported in \cite{Bull2019}.

\begin{figure}[ht!]
	\begin{subfigure}{.5\textwidth}
		\centering
		\scalebox{0.4}{
			\includegraphics{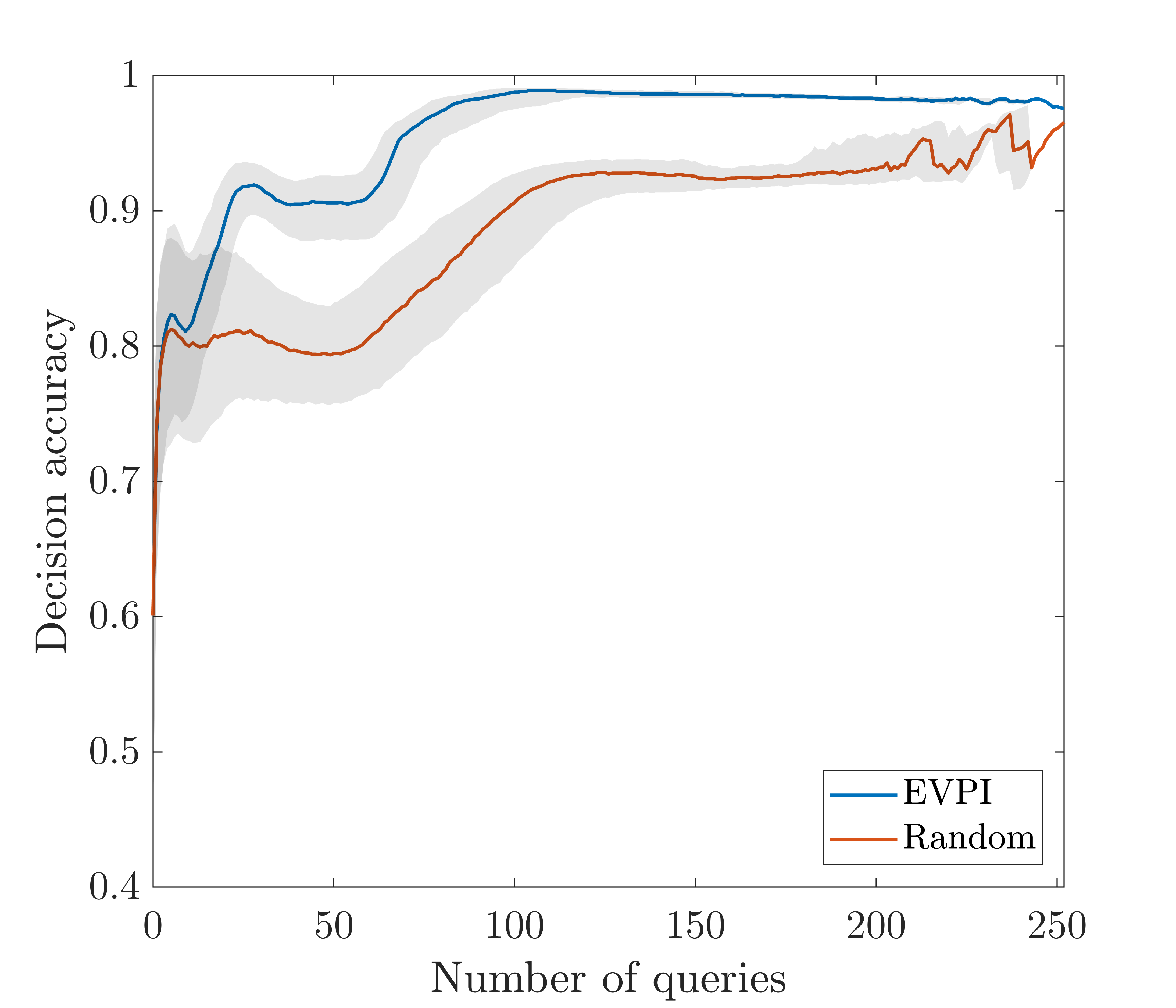}
		}
		\caption{ }
		\label{fig:dacc_z24}
	\end{subfigure}
	\begin{subfigure}{.5\textwidth}
		\centering
		\scalebox{0.4}{
			\includegraphics{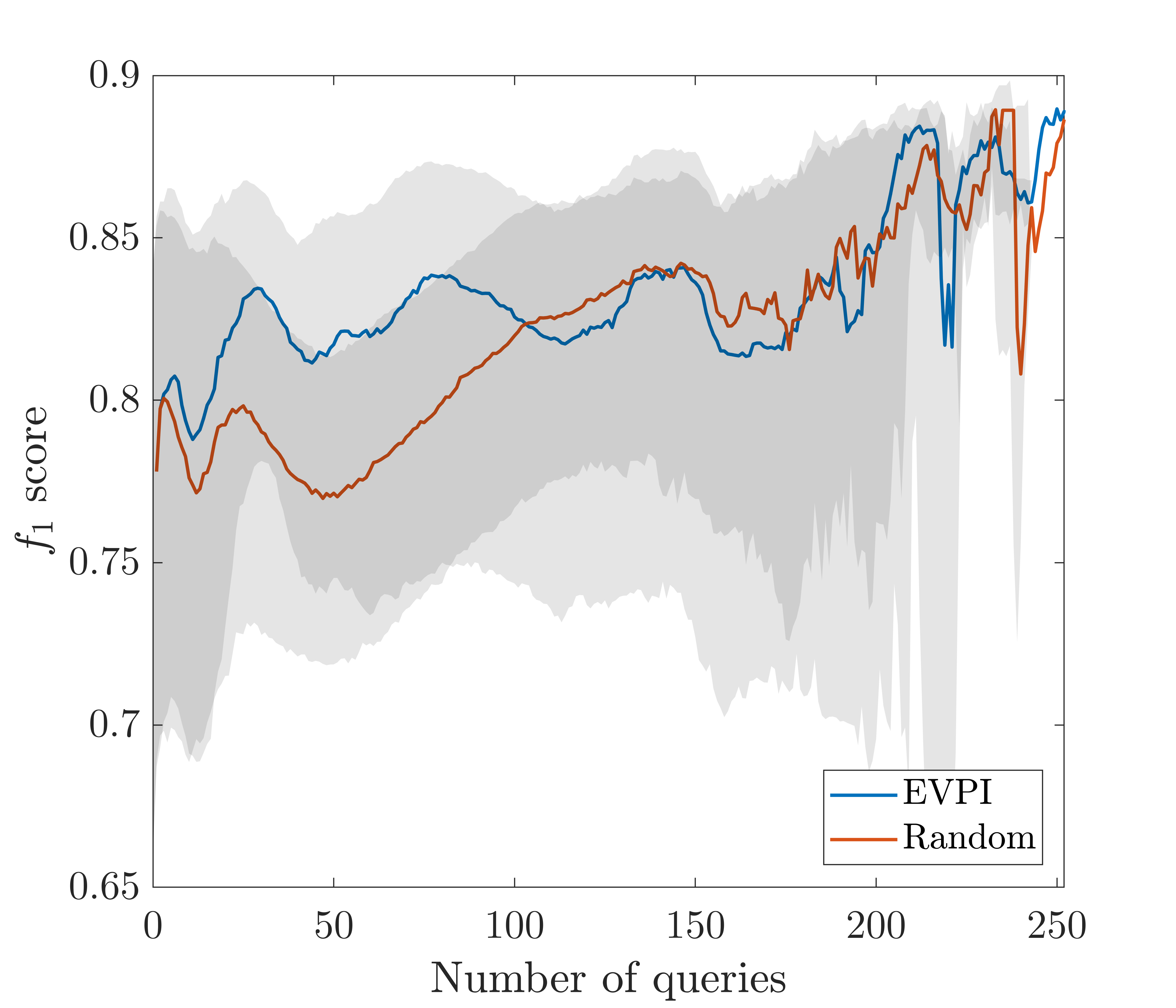}
		}
		\caption{ }
		\label{fig:f1score_z24}
	\end{subfigure}
	\caption{Variation in median (a) decision accuracy and (b) $f_{1}$ score with number of label queries for an agent utilising a GMM learned from $\mathcal{D}_l$ extended via (i) risk-based active querying (EVPI) and (ii) random sampling (Random). Shaded area shows the interquartile range.}
	\label{fig:performance_z24}
\end{figure}

\begin{figure}[ht!]
	\begin{subfigure}{.5\textwidth}
		\centering
		\scalebox{0.4}{
			\includegraphics{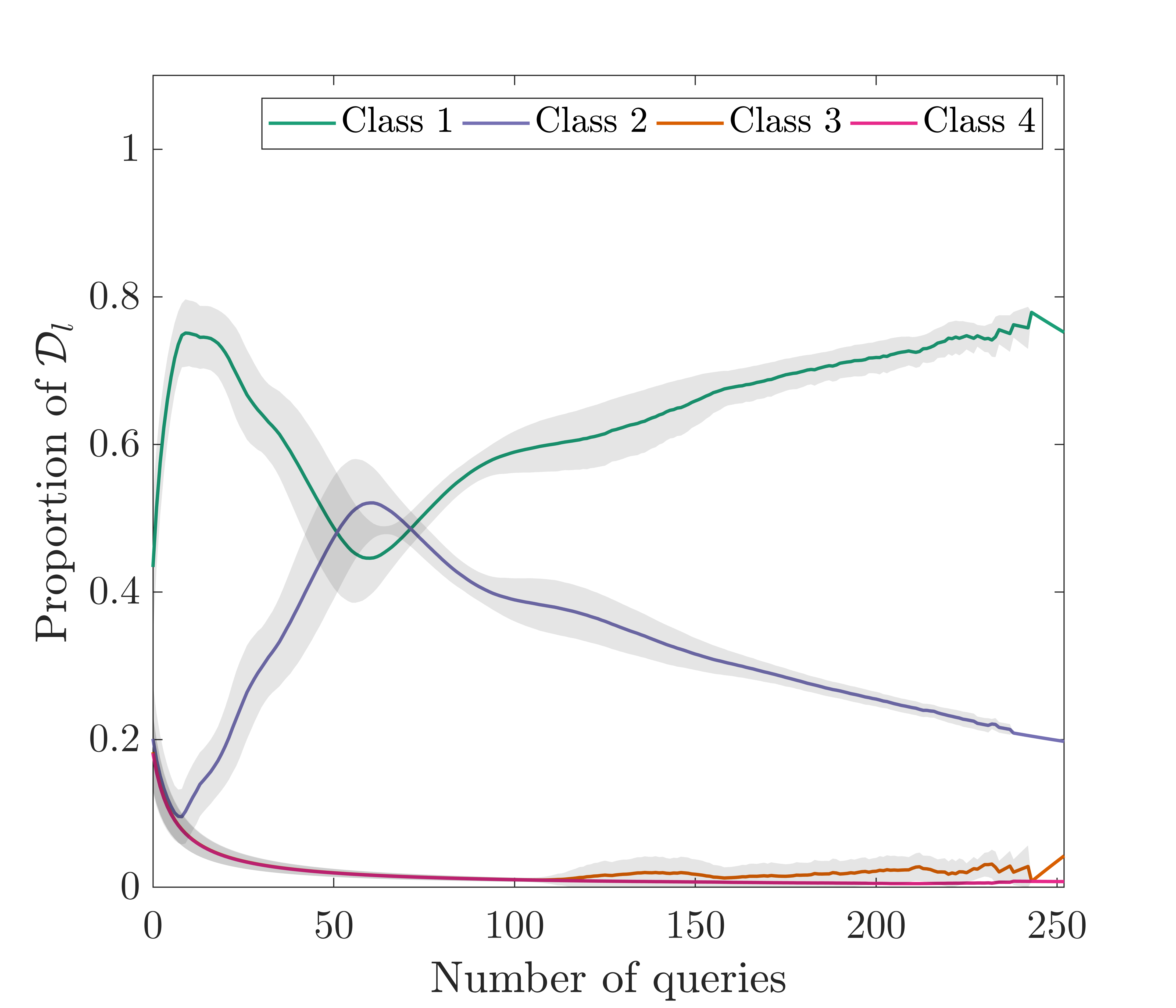}
		}
		\caption{ }
		\label{fig:cpropral_z24}
	\end{subfigure}
	\begin{subfigure}{.5\textwidth}
		\centering
		\scalebox{0.4}{
			\includegraphics{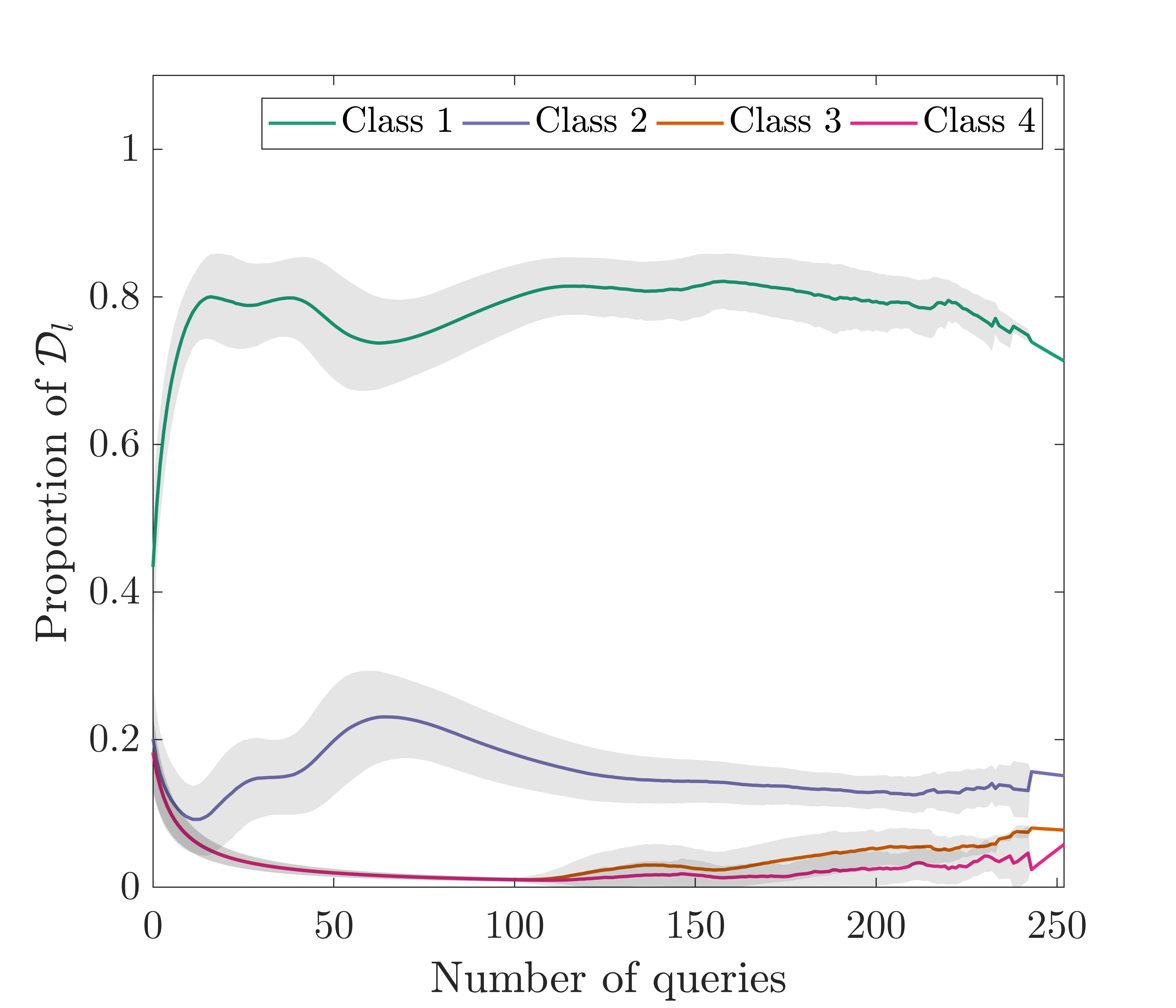}
		}
		\caption{ }
		\label{fig:cproprand_z24}
	\end{subfigure}
	\caption{Variation in class proportions within $\mathcal{D}_l$ with number of label queries for an agent utilising a GMM learned from $\mathcal{D}_l$ extended via (a) risk-based active querying and (b) random sampling. Shaded area shows $\pm1\sigma$.}
	\label{fig:cprop_z24}
\end{figure}

Figure \ref{fig:cprop_z24} shows the mean proportions that each class contributes to the labelled dataset $\mathcal{D}_l$ throughout the learning process. Figure \ref{fig:cpropral_z24} provides the class proportions for EVPI-based querying, whereas Figure \ref{fig:cproprand_z24} provides the proportions for random querying. Logically, both datasets show an increase in the proportion for Class 1 as this is the first class to present itself from the perspective of the decision-maker. One difference between datasets is that the risk-based active-learning approach results in Class 2 (cold temperature) gaining an elevated level of representation in $\mathcal{D}_l$; this is perhaps understandable if one considers the overlap present in the dataset, especially that between Class 2 and Class 3. Throughout later queries, one can see that class 1 gains increased representation in $\mathcal{D}_l$. This increase correlates with the slight decline in decision accuracy observed, indicating that data acquired during this time cause the cluster corresponding to class 1 to be altered such that it is suboptimal for decision-making. 

\begin{figure}[ht!]
	\centering
		\scalebox{0.4}{
			\includegraphics{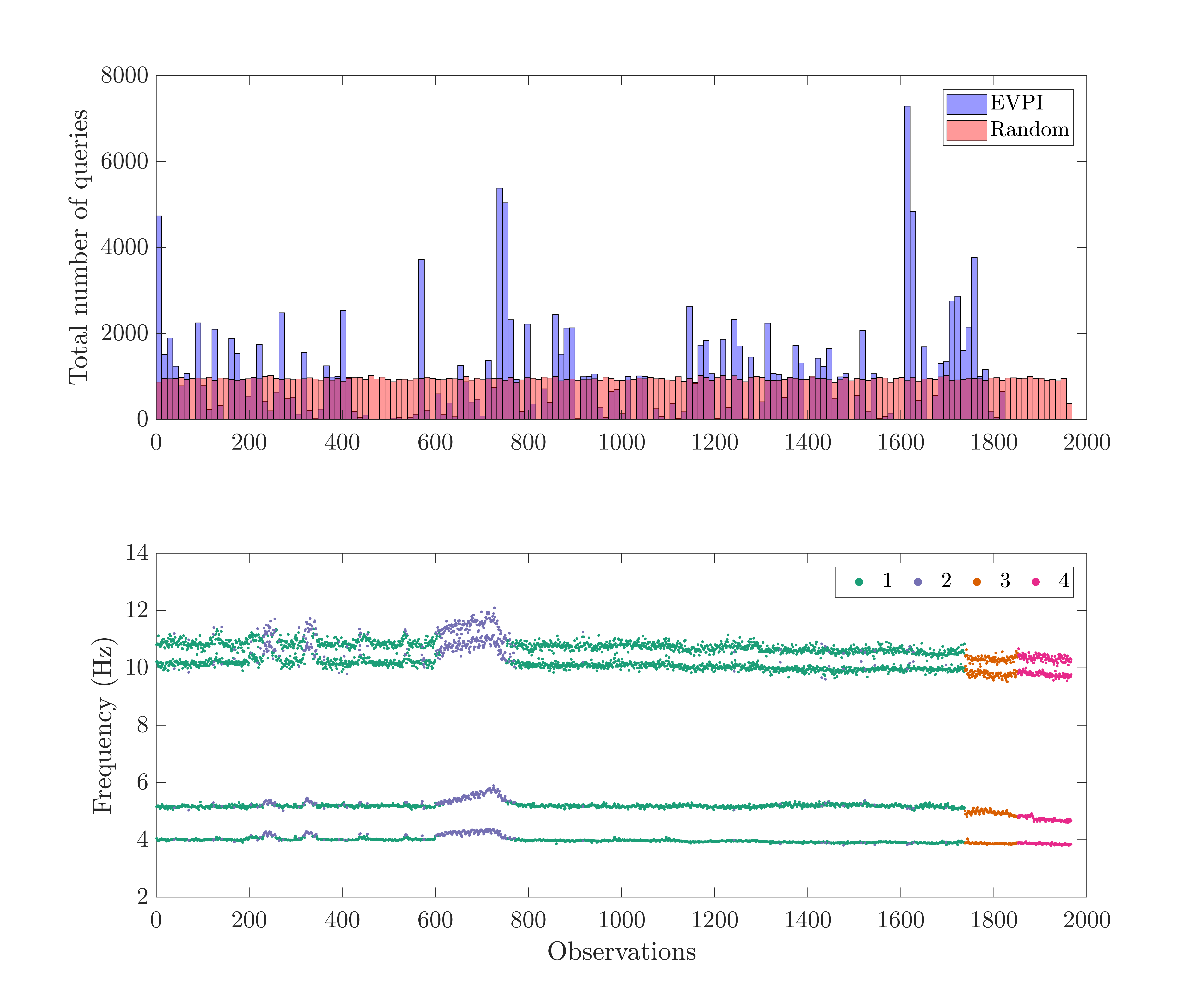}
		}
		\caption{Histograms showing the distribution of the number of queries for each observation in $\mathcal{D}_u$ over 1000 runs when adopting (i) risk-based active learning (EVPI) and (ii) random sampling (Random) in order to learn a GMM. Below the histograms, a visualisation of the unlabelled dataset $\mathcal{D}_u$ is provided for reference.}
		\label{fig:all_queries_z24}
\end{figure}

Figure \ref{fig:all_queries_z24} shows the total number of times each observation in $\mathcal{D}_u$ was queried throughout the 1000 runs, with the observations in $\mathcal{D}_u$ presented for reference. As one would expect, random sampling results in uniform querying across the dataset. On the other hand, the risk-based active learning approach exhibits definite preferences for specific observations. One tends to see an increase in observation data corresponding to previously unseen classes; however, the exception to this is Class 4 (advanced damage) -- indicating that the data in the vicinity of the cluster corresponding to Class 4 have low value of information. This result is, of course, because a decision-maker observing data in this region of the feature space can be certain that the optimal policy is `perform maintenance' because of the extremely high cost associated with this class. The most heavily-queried data points are in the vicinity of observation number 1600. These data belong to classes 1 and 2 -- both undamaged classes. The fact that they are so heavily queried, however, implies that they consistently lie close to the decision boundary between the candidate actions `do nothing' and `perform maintenance'.

To summarise, when utilising a GMM as the statistical classifier within a decision process for the Z24 Bridge dataset, one can obtain improved decision-making performance; however, as with the previous case study, a decline in performance still occurs with queries that are made later in the learning process.

\subsection{Results: Semi-supervised learning}\label{sec:Z24_Results_ss}

The modified approaches to risk-based active learning, incorporating semi-supervised learning via EM and latent-state smoothing, were each applied to the Z24 Bridge case study.

Figure \ref{fig:performance_z24_ss} shows, for each algorithm, the median decision and classification performances as a function of the number of queries. Additionally, the performances for the risk-based active-learned GMM are provided for comparison.

It is immediately apparent from both Figure \ref{fig:dacc_z24_ss} and Figure \ref{fig:f1score_z24_ss}, that the semi-supervised approaches have failed in improving robustness to sampling bias; in fact, the semi-supervised learning approaches have been detrimental to decision-making performance. Neither expectation-maximisation updates to the generative model parameters, nor pseudo-labelling of unlabelled data via a smoothing algorithm, result in a classifier that surpasses the standard GMM in terms of decision accuracy. Similarly, both approaches fail to surpass the standard GMM in terms of classification performance.

For the current case study, the poor performance of the semi-supervised methods can be attributed to the fact that the algorithms further increase dependence on the key assumption required for learning generative models; namely, that the density estimations selected are representative of the underlying distribution. This increased dependence arises as the generative models, learned from an unrepresentative dataset, are being used to inform the pseudo-labels that are subsequently used to update the model. As the Z24 Bridge data is non-Gaussian and a Gaussian density estimation was used, the semi-supervised steps in the modified risk-based learning algorithm yield biased pseudo-labels and therefore exacerbate the effects of sampling bias. These results correspond with the results of the active learning EM approach in \cite{BullThesis}, in which introducing EM to the active learning process degrades the classification performance of a GMM on the Z24 Bridge dataset.

Figure \ref{fig:performance_z24_ss} shows that the smoothing approach to semi-supervised learning outperforms EM in terms of decision accuracy and $f_1$-score. This observation is likely because the smoothing algorithm utilises temporal information from the transition model. This additional information source somewhat tempers the dependence on the (biased) classification model that one solely relies on in the EM approach.

\begin{figure}[ht!]
	\begin{subfigure}{.5\textwidth}
		\centering
		\scalebox{0.4}{
			\includegraphics{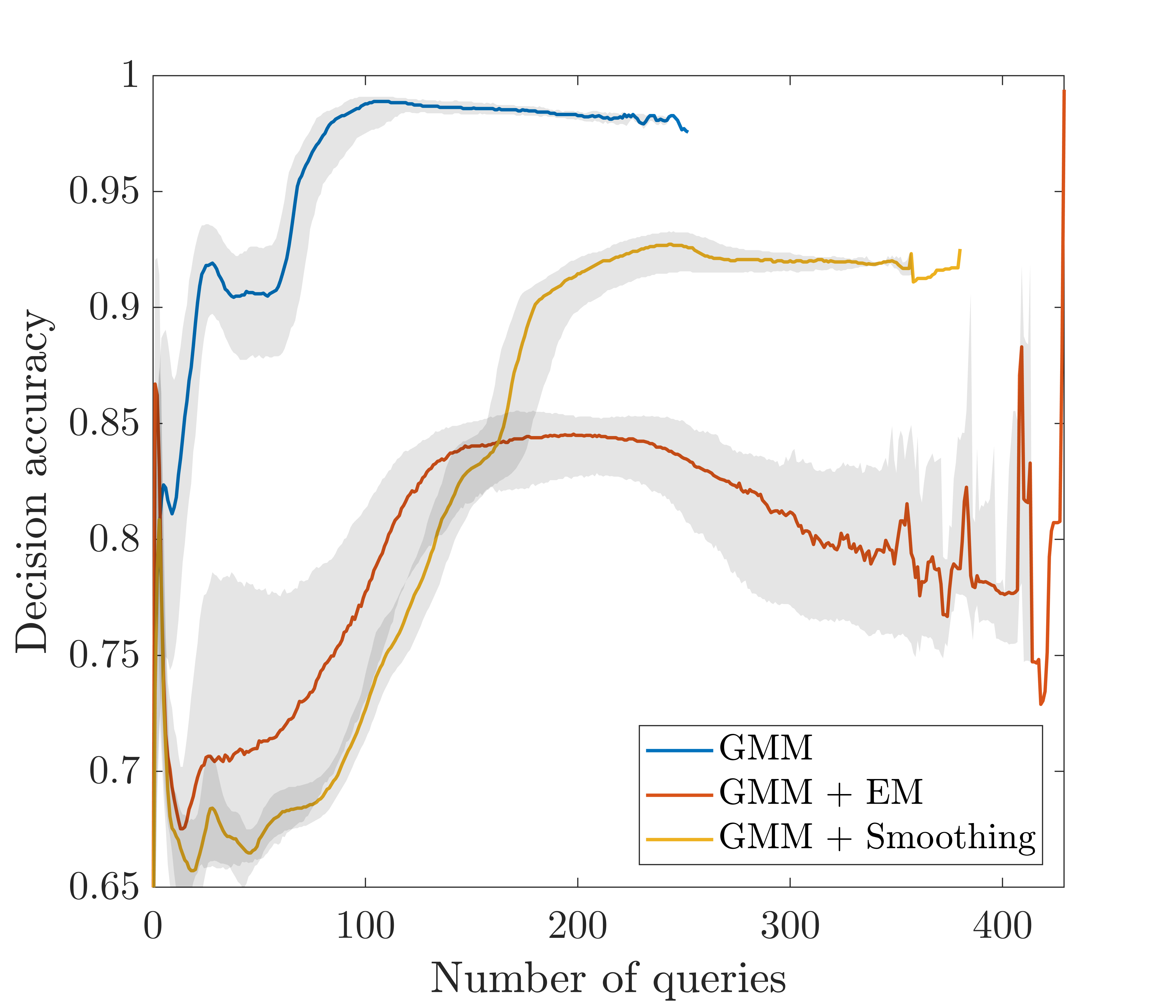}
		}
		\caption{ }
		\label{fig:dacc_z24_ss}
	\end{subfigure}
	\begin{subfigure}{.5\textwidth}
		\centering
		\scalebox{0.4}{
			\includegraphics{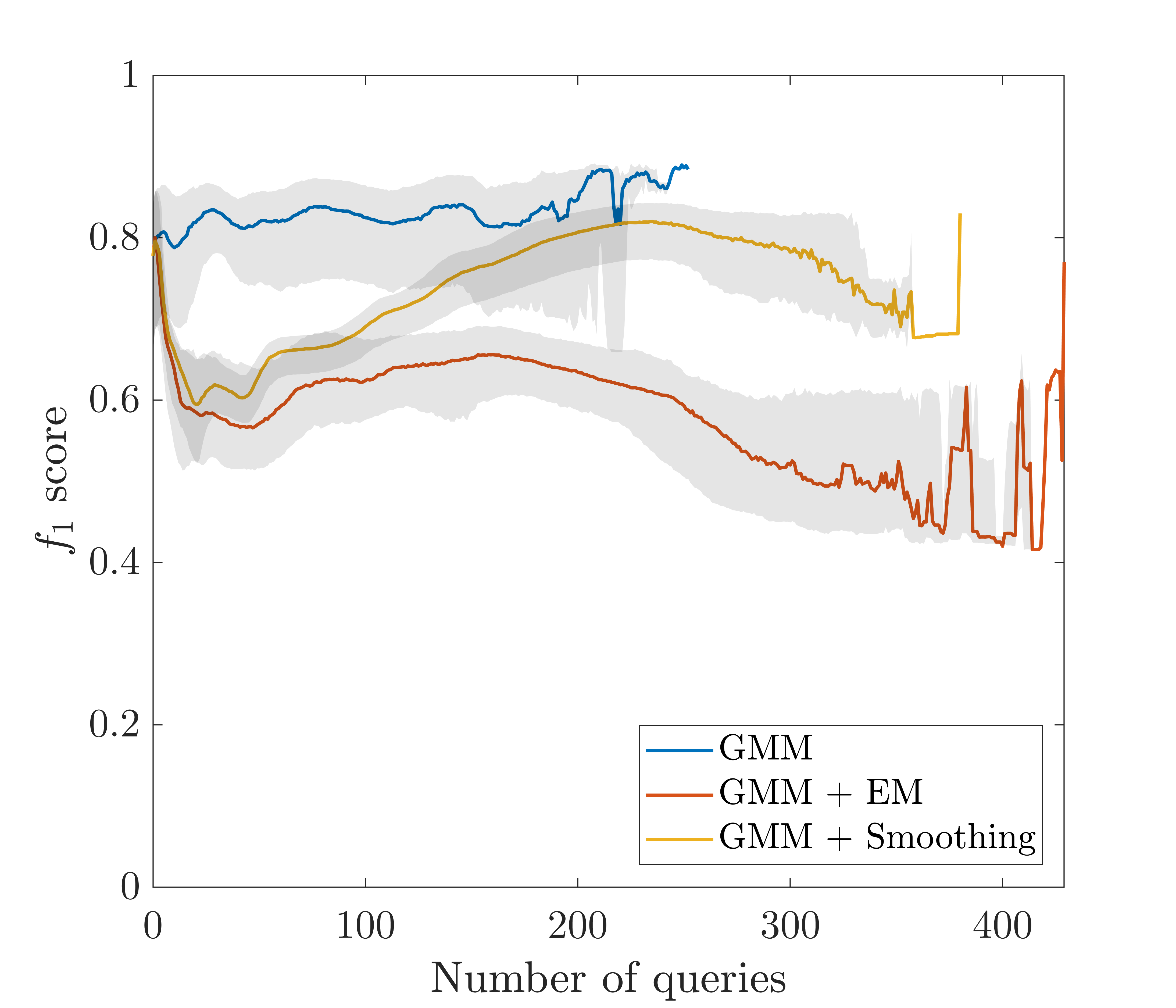}
		}
		\caption{ }
		\label{fig:f1score_z24_ss}
	\end{subfigure}
	\caption{Variation in median (a) decision accuracy and (b) $f_{1}$ score with number of label queries for an agent utilising a GMM learning via risk-based active learning (i) without semi-supervised updates, (ii) with semi-supervised updates via EM and (iii) with semi-supervised updated via smoothing. Shaded area shows the interquartile range.}
	\label{fig:performance_z24_ss}
\end{figure}

\begin{figure}[ht!]
	\begin{subfigure}{.5\textwidth}
		\centering
		\scalebox{0.4}{
			\includegraphics{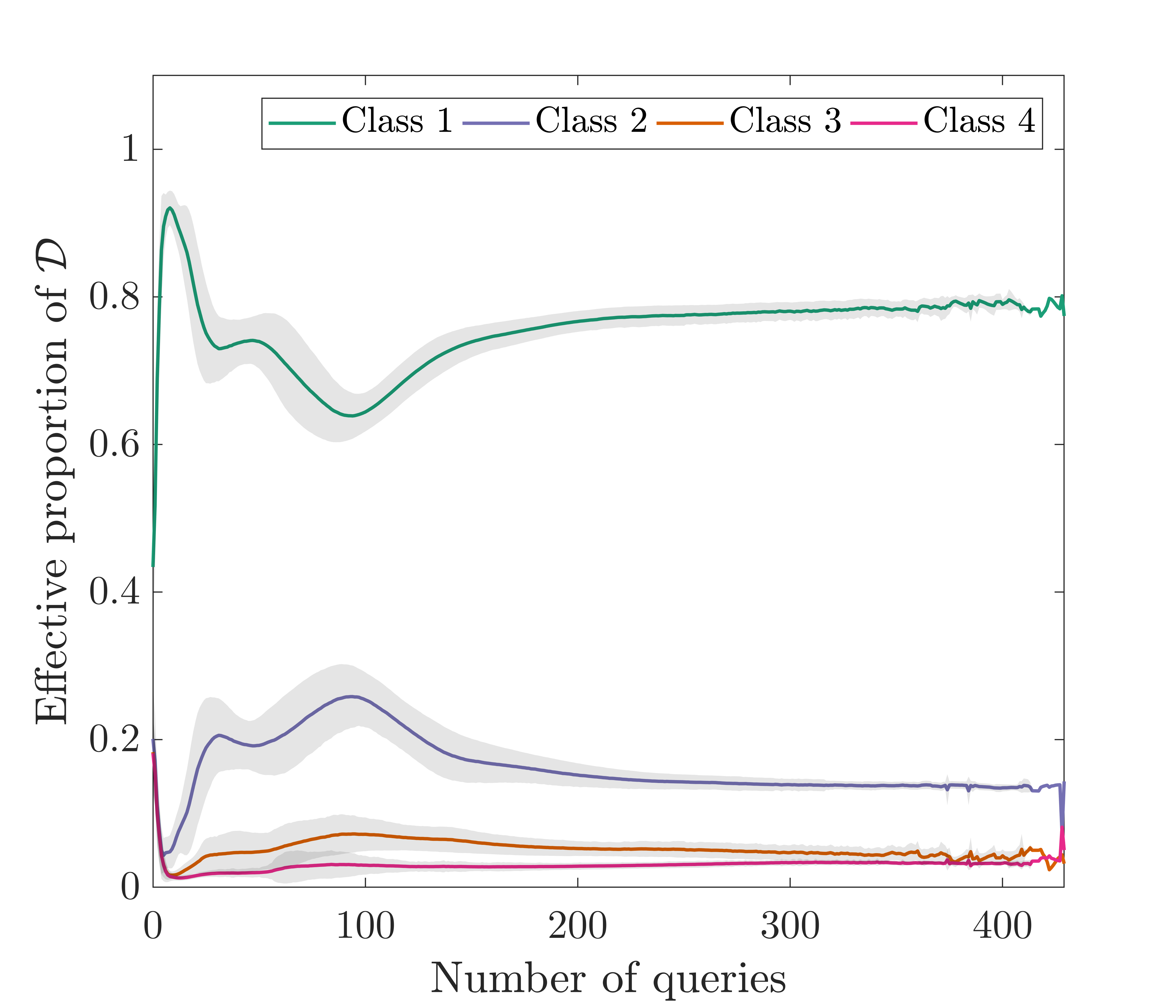}
		}
		\caption{ }
		\label{fig:cprop_em_z24}
	\end{subfigure}
	\begin{subfigure}{.5\textwidth}
		\centering
		\scalebox{0.4}{
			\includegraphics{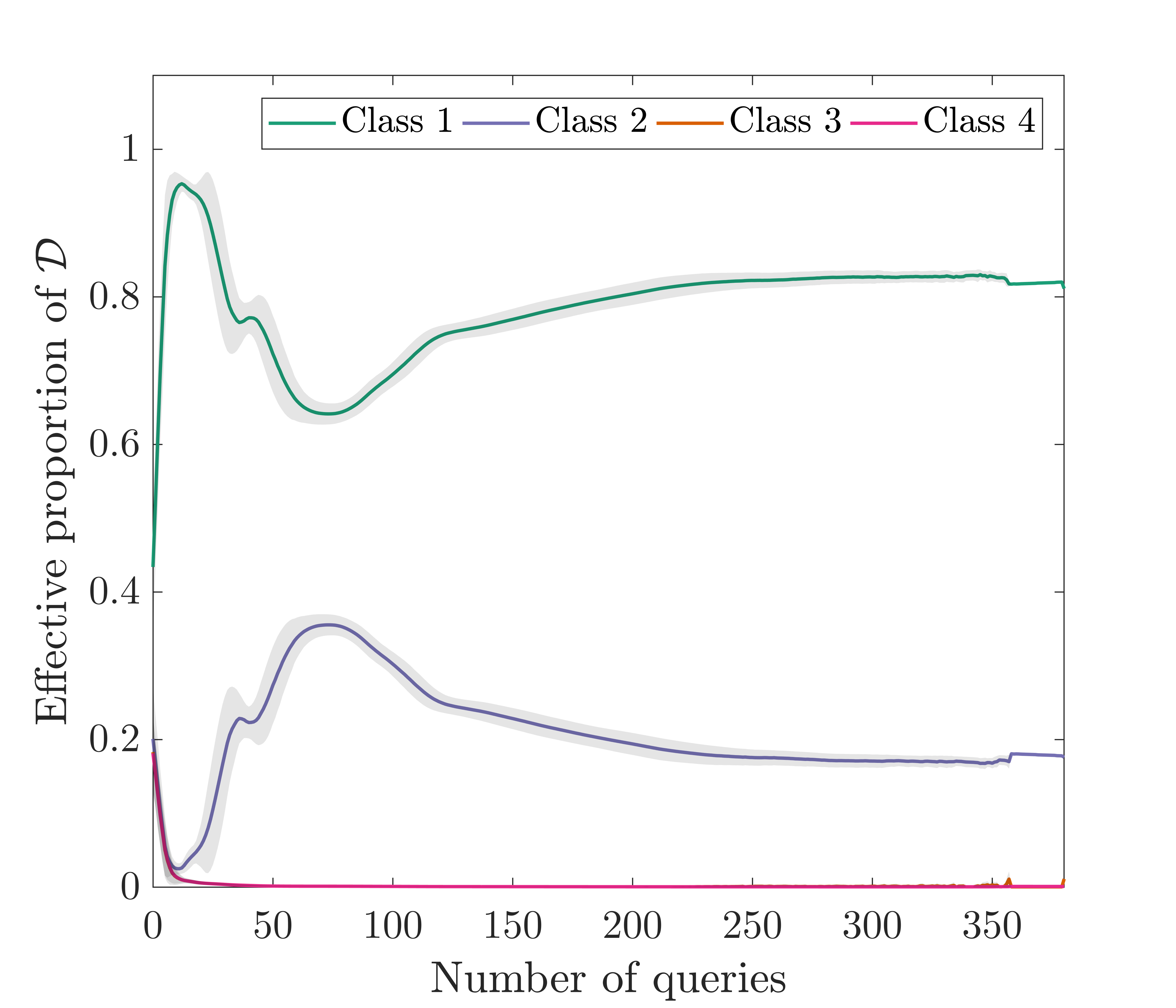}
		}
		\caption{ }
		\label{fig:cprop_smooth_z24}
	\end{subfigure}
	\caption{Variation in class proportions within $\mathcal{D}_l$ with number of label queries for an agent utilising a GMM learned from $\mathcal{D}_l$ extended risk-based active learning with semi-supervised updating via (a) expectation-maximisation and (b) latent-state smoothing. Shaded area shows $\pm1\sigma$.}
	\label{fig:cprop_ss_z24}
\end{figure}

Figure \ref{fig:cprop_ss_z24} shows the effective class proportions for data in the training set $\mathcal{D}$, throughout the risk-based active learning of semi-supervised GMMs. From these figures one can see that the semi-supervised algorithms initially induce a heavy bias in favour of Class 1 within the training dataset. These class proportions are likely to inflate both the component mixture weights and the covariance for the cluster belonging to Class 1. This initial bias explains the dramatic decrease in decision accuracy and $f_1$-score that can be observed over the first few queries in Figure \ref{fig:performance_z24_ss}.

\begin{figure}[ht!]
	\centering
		\scalebox{0.4}{
			\includegraphics{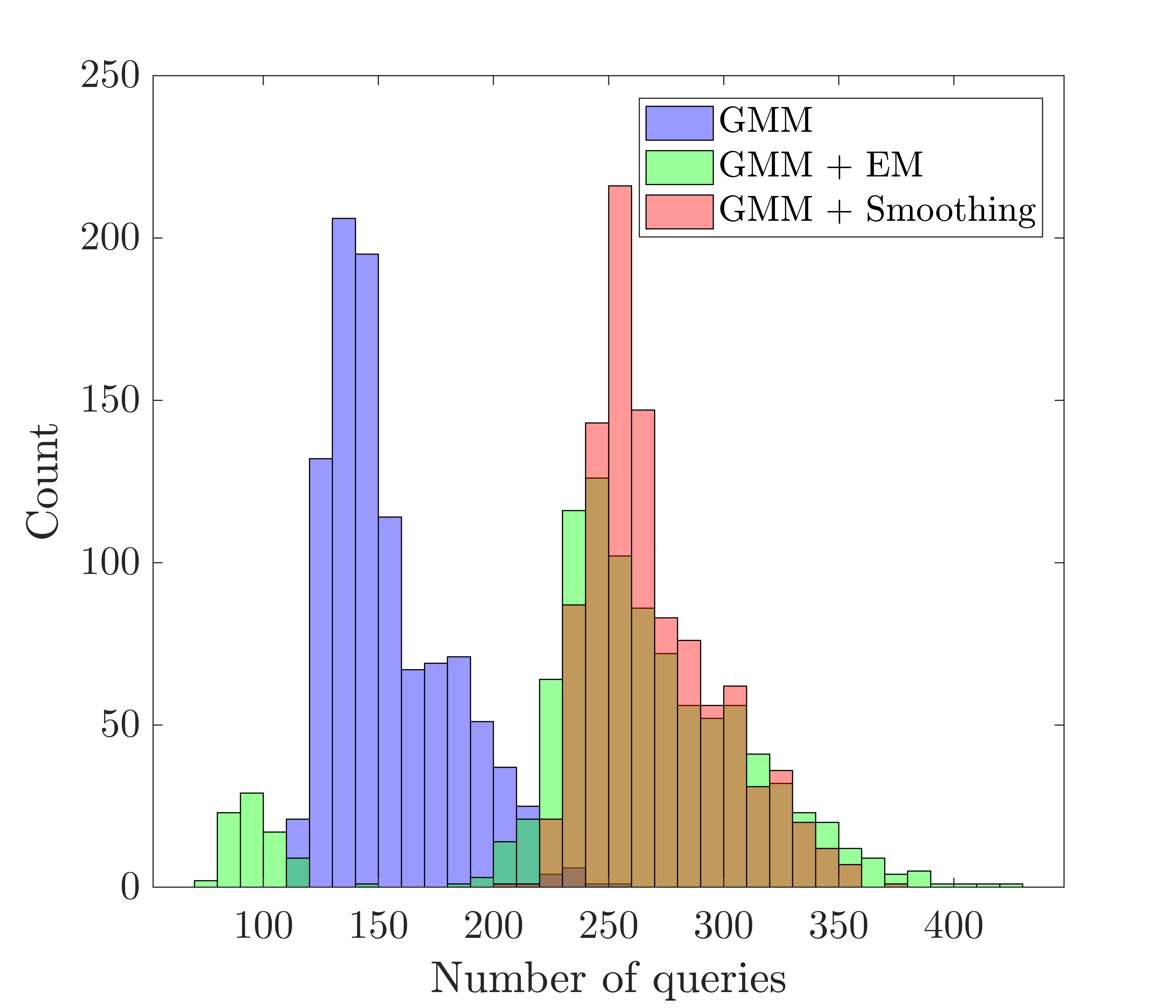}
		}
		\caption{Histograms showing the distribution of the number of queries from 1000 runs of the risk-based active learning of (i) a GMM (blue) (ii) a GMM semi-supervised via expectation-maximisation (green) and (iii) a GMM semi-supervised via latent-state smoothing (red).}
		\label{fig:hist_ss_z24}
\end{figure}

\begin{figure}[ht!]
	\centering
		\scalebox{0.4}{
			\includegraphics{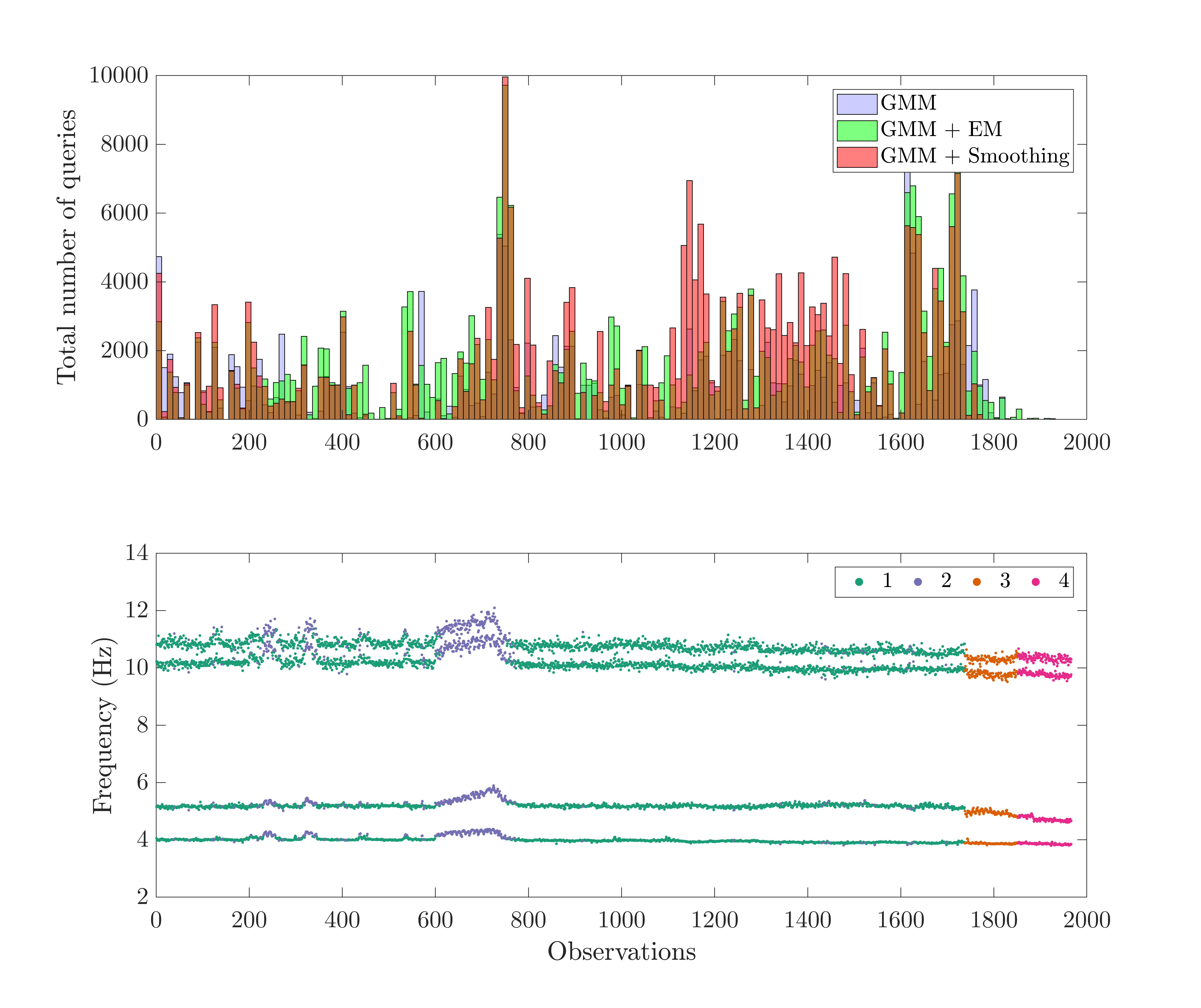}
		}
		\caption{Histograms showing the distribution of the number of queries for each observation in $\mathcal{D}_u$ from 1000 runs of risk-based active learning for (i) a GMM (blue) (ii) a GMM with EM (green) and (iii) a GMM with smoothing (red). Below the histograms, a visualisation of the unlabelled dataset $\mathcal{D}_u$ is provided for reference.}
		\label{fig:all_queries_ss_z24}
\end{figure}

In addition to decreasing the decision-making performance, the introduction of semi-supervised learning to risk-based active learning also has a detrimental effect on the number of queries made. Figure \ref{fig:hist_ss_z24} shows the distribution of the number of queries made over 1000 randomly-initialised repetitions. It is clear from Figure \ref{fig:hist_ss_z24}, that both EM and latent-state smoothing result in a higher mean for the number of queries made, when compared to the standard GMM. This result implies that using these modified approaches to risk-based active learning result in increased expenditure on structural inspections. Interestingly, the distribution of the number of queries made using the risk-based active learning algorithm that incorporates EM, is bimodal. The dominant mode corresponds to worsened performance in comparison to the standard GMM; however, the secondary mode corresponds to an improved performance. This result suggests that there exists a modality of the learning algorithm in which the EM algorithm can converge to a classifier that results in a well-defined decision boundary. As the variability in results arises because of the randomisation of the initial labelled dataset $\mathcal{D}_l$, it can be inferred that the occurrence of this secondary modality is conditional on which labelled data are available for learning the fully-supervised model, prior to the EM updates.

Figure \ref{fig:all_queries_ss_z24} shows the total number of queries for each observation index in $\mathcal{D}_u$, over the 1000 repetitions conducted for each of the two variants of the risk-based active learning process. Additionally, the queries made by the agent utilising the standard GMM are provided for comparison. As one would expect, given the histograms in Figure \ref{fig:hist_ss_z24}, the queries made by the semi-supervised algorithms all but obscure those made when using the standard GMM. These semi-supervised algorithms result in data being heavily queried across many observations in $\mathcal{D}_u$, indicating that a well-defined decision boundary is not obtained - an effect of the exacerbated sampling bias. Here, it is worth noting that, despite numerous queries elsewhere in the dataset, Class 4 remains more-or-less unqueried. Again, this result indicates that due to the extreme cost associated with the advanced damage state, a decision-maker will confidently select the action `perform maintenance' when in the vicinity of the cluster corresponding to Class 4.

In summary, the introduction of semi-supervised learning via EM, and latent-state smoothing, resulted in a degradation in the performance of an agent tasked with selecting actions within a maintenance decision process defined around the Z24 Bridge dataset. This degradation occurred because of the interaction between important characteristics of the dataset (class imbalance, non-Gaussianity, and non-separability) and the strengthened assumptions (i.i.d training data, and representative distribution forms) invoked via the semi-supervised learning algorithms. These interactions resulted in the amplification of sampling bias, leading to worsened performance.

\subsection{Results: Discriminative classifiers}\label{sec:Z24_Results_rvm}

Discriminative classifiers, in the form of \mRVMa{} and \mRVMb{}, were substituted into decision processes in place of the GMM. Risk-based active learning was employed to develop these classification models for the Z24 Bridge dataset.

The decision accuracy and $f_1$-score as functions of the number of queries made are provided for the risk-based active learning of \mRVMa{} and \mRVMb{} in Figure \ref{fig:performance_rvm_z24}.

From Figure \ref{fig:dacc_rvm_z24}, it can be seen that both \mRVMa{} and \mRVMb{} surpass the GMM in terms of decision-making performance. \mRVMa{} begins with very low performance, once again indicating that the constructive approach to selecting relevance vectors was unable to converge, because of the limiting size of $\mathcal{D}_l$. Over the first few queries, however, the performance rapidly increases as $\mathcal{D}_l$ expands such that convergence of the relevance vector selection can be achieved. As with the previous case study, \mRVMb{} achieves good performances on even the limited initial $\mathcal{D}_l$. Furthermore, improvements in decision accuracy are gained at a similar, if not a slightly greater, rate when compared to the GMM. This result means that superior decision-making performance is obtained throughout the querying process when adopting an \mRVMb{} classifier. It is again worth noting that neither formulations of RVM suffer from a degradation in decision-making performance over later queries - indicating robustness to sampling bias in $\mathcal{D}_l$. Again, this result can be attributed to the discriminative nature of the models, in addition to the fact that extraneous data are excluded from $\mathcal{A}$.

From Figure \ref{fig:f1score_rvm_z24}, it can be seen that the RVMs have inferior $f_1$-scores in comparison to the GMM. Distinct improvements in classification performance arise in correlation with appearance of new classes in $\mathcal{D}_u$ (see Figure \ref{fig:all_queries_rvm_z24}). The performances plateau at approximately 0.6 as data corresponding to Class 4 are not queried, resulting in the classifiers struggling to discriminate between classes 3 and 4. Nonetheless, as these classes share an optimal policy with respect to the decision process, decision-making performance in not impacted by these misclassifications -- as demonstrated in Figure \ref{fig:dacc_rvm_z24}.

\begin{figure}[ht!]
	\begin{subfigure}{.5\textwidth}
		\centering
		\scalebox{0.4}{
			\includegraphics{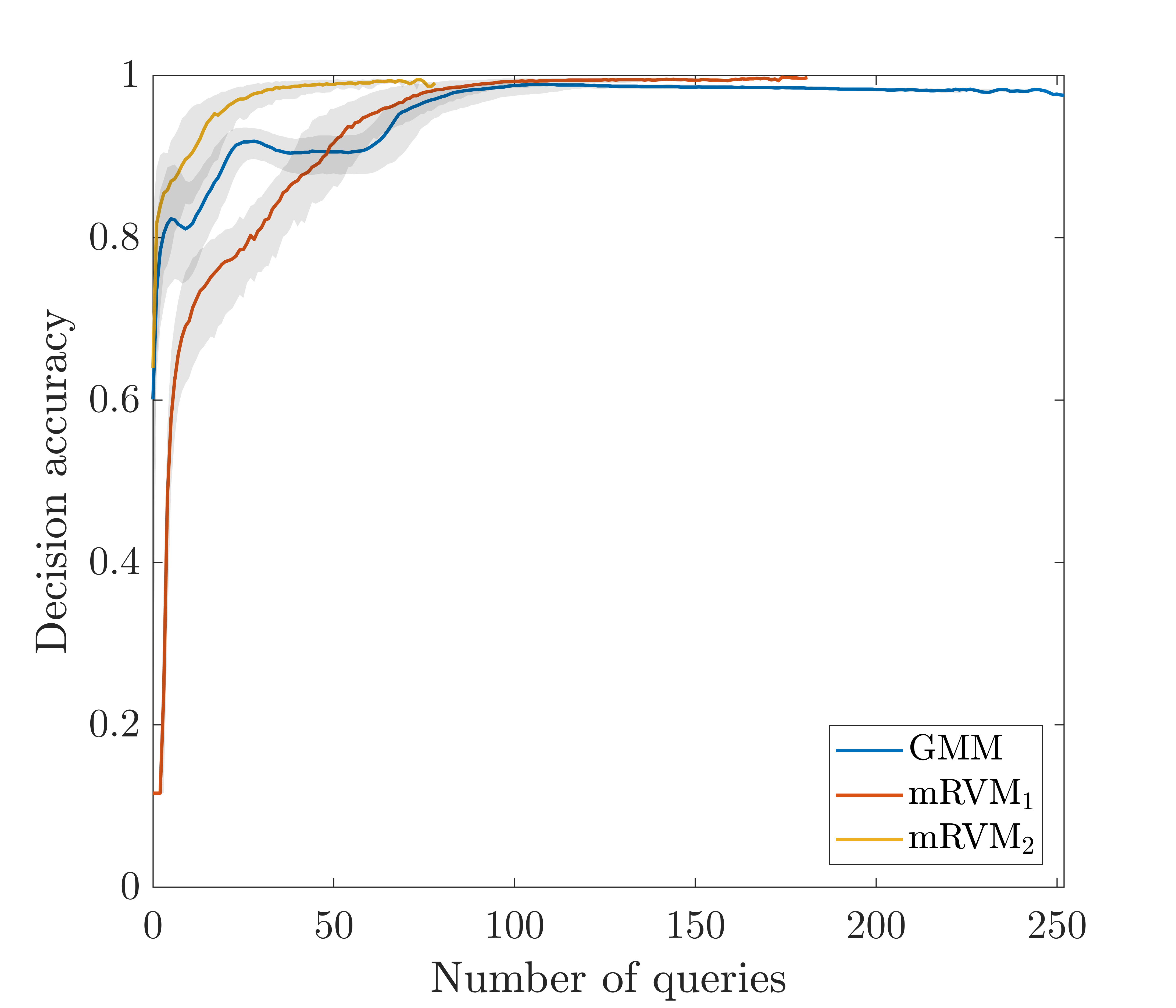}
		}
		\caption{ }
		\label{fig:dacc_rvm_z24} 
	\end{subfigure}
	\begin{subfigure}{.5\textwidth}
		\centering
		\scalebox{0.4}{
			\includegraphics{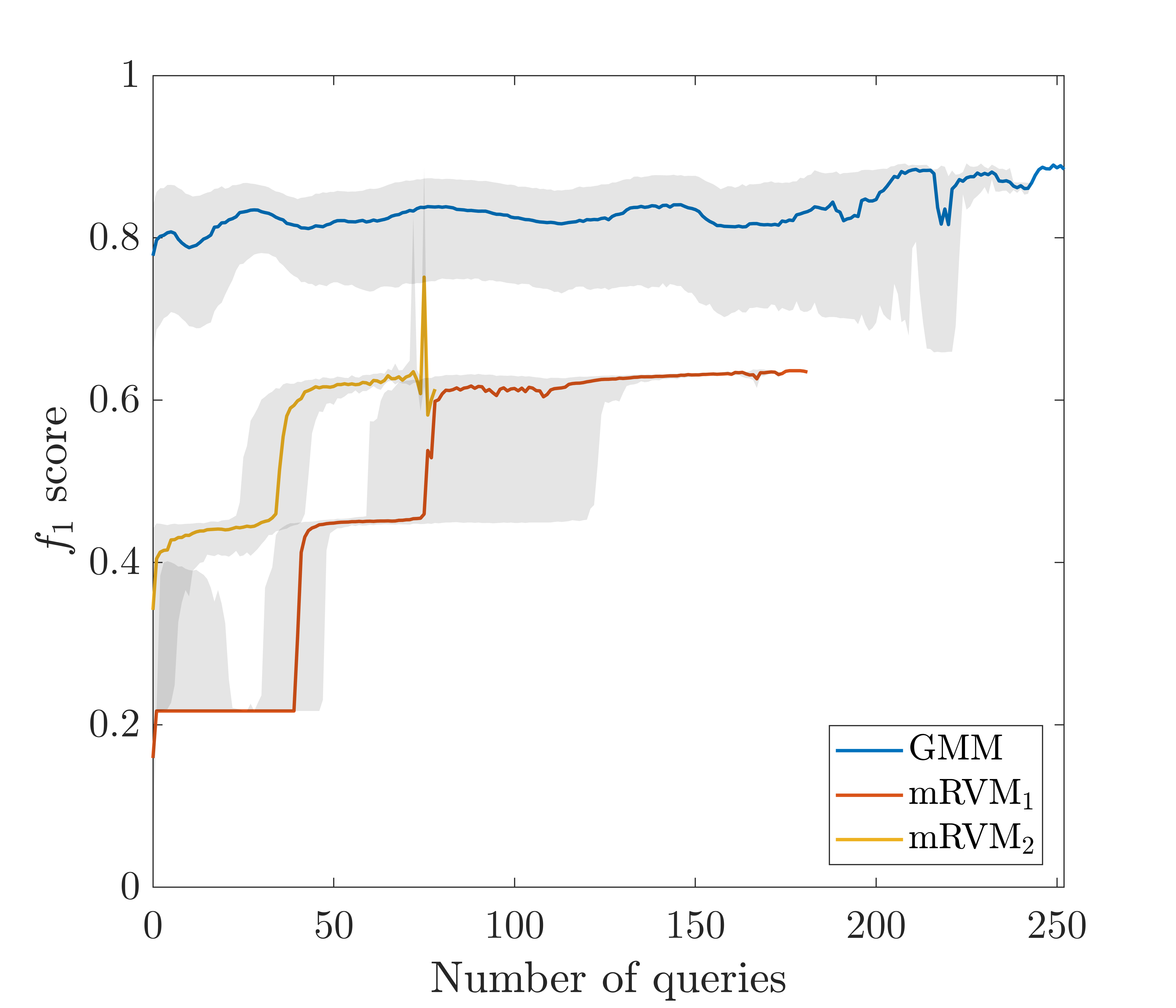}
		}
		\caption{ }
		\label{fig:f1score_rvm_z24}
	\end{subfigure}
	\caption{Variation in median (a) decision accuracy and (b) $f_{1}$ score with number of label queries for an agent utilising (i) a GMM (ii) an mRVM$_1$ and (iii) an mRVM$_2$ statistical classifier learned via risk-based active learning. Shaded area shows the interquartile range.}
	\label{fig:performance_rvm_z24}
\end{figure}

\begin{figure}[ht!]
	\begin{subfigure}{.5\textwidth}
		\centering
		\scalebox{0.4}{
			\includegraphics{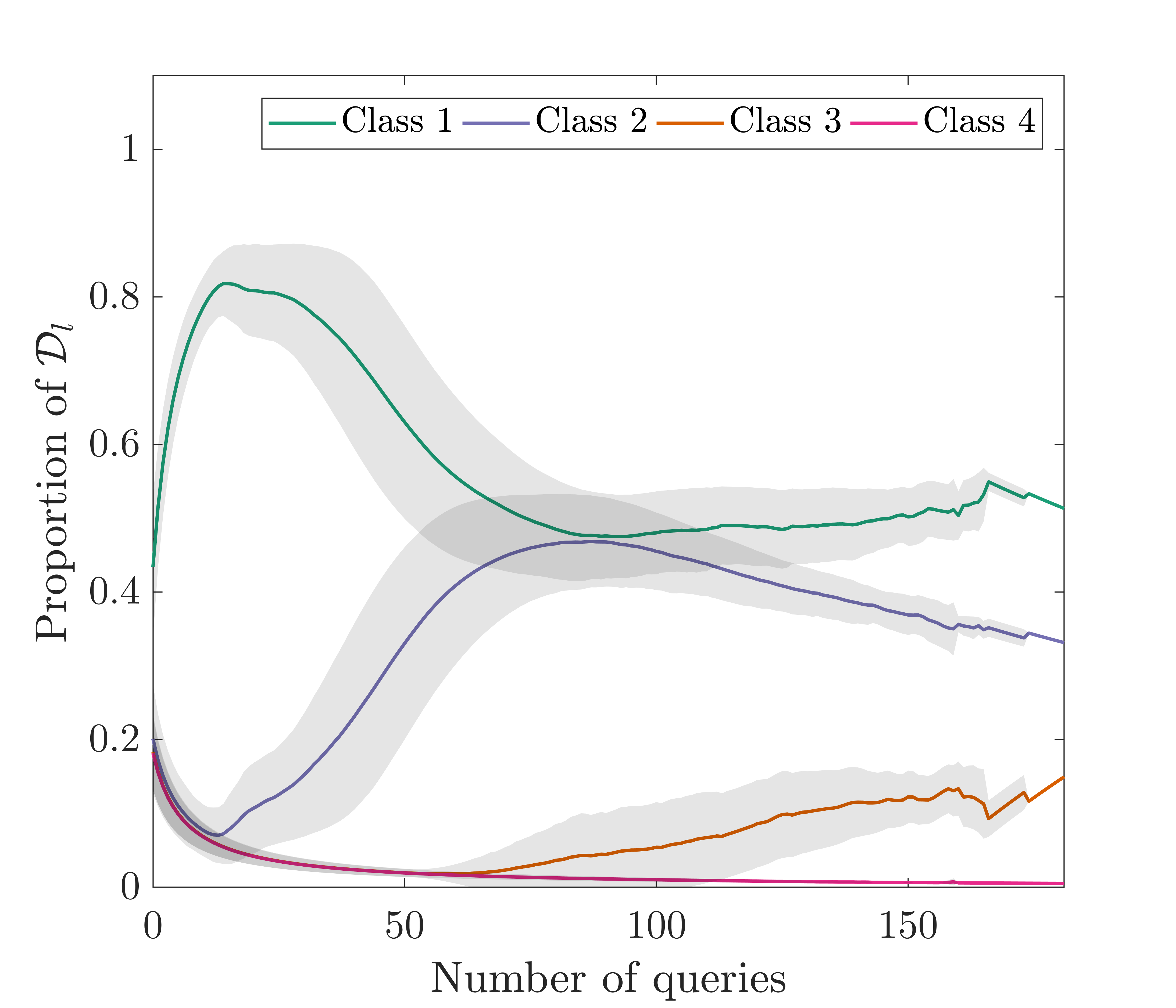}
		}
		\caption{ }
		\label{fig:cprop_mRVM1_z24}
	\end{subfigure}
	\begin{subfigure}{.5\textwidth}
		\centering
		\scalebox{0.4}{
			\includegraphics{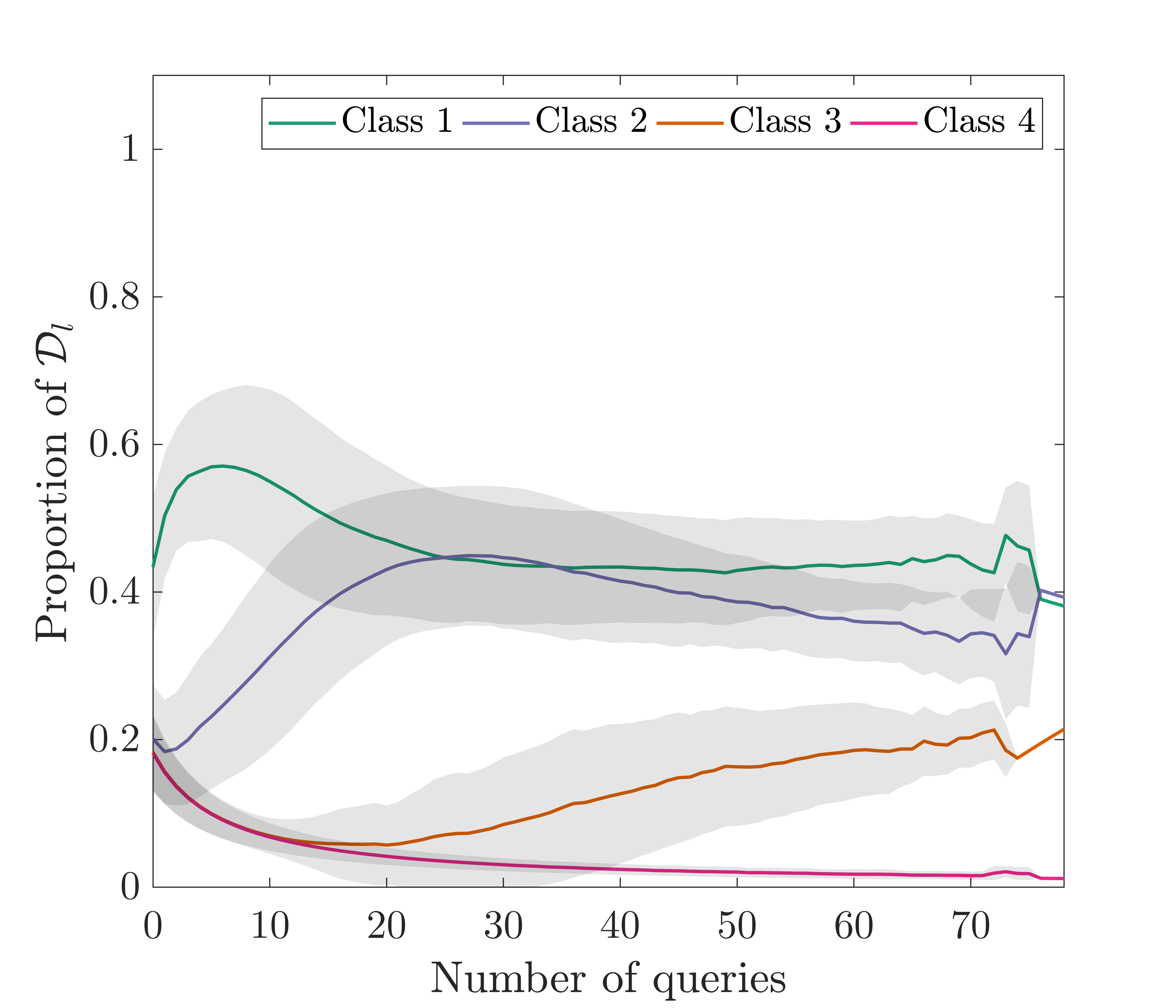}
		}
		\caption{ }
		\label{fig:cprop_mRVM2_z24}
	\end{subfigure}
	\caption{Variation in class proportions within $\mathcal{D}_l$ with number of label queries for an agent utilising an (a) mRVM$_1$ and (b) mRVM$_2$ statistical classifier learned from $\mathcal{D}_l$ extended via risk-based active learning. Shaded area shows $\pm1\sigma$.}
	\label{fig:cprop_rvm_z24}
\end{figure}

Figure \ref{fig:cprop_rvm_z24} shows the class proportions in $\mathcal{D}_l$ throughout the risk-based active learning of the mRVM classifiers. In one compares Figures \ref{fig:cprop_mRVM1_z24} and \ref{fig:cprop_mRVM2_z24} with Figure \ref{fig:cproprand_z24}, it becomes apparent that the class proportions following the active learning of \mRVMa{} and \mRVMb{} are not representative of the overall dataset. In fact, the class imbalance present in the full dataset is lessened in $\mathcal{D}_l$, with classes 1 and 2 garnering approximately 40\% representation each, and class 3 possessing the remaining 20\%. This deviation in class representation can be attributed to the RVM's sparsity -- as only a few prototypical examples are required to sufficiently represent each class; once a class is established following a few queries, data associated the class need not be queried further. This result is further supported by the step-like improvements in $f_1$-score.

\begin{figure}[ht!]
	\centering
		\scalebox{0.4}{
			\includegraphics{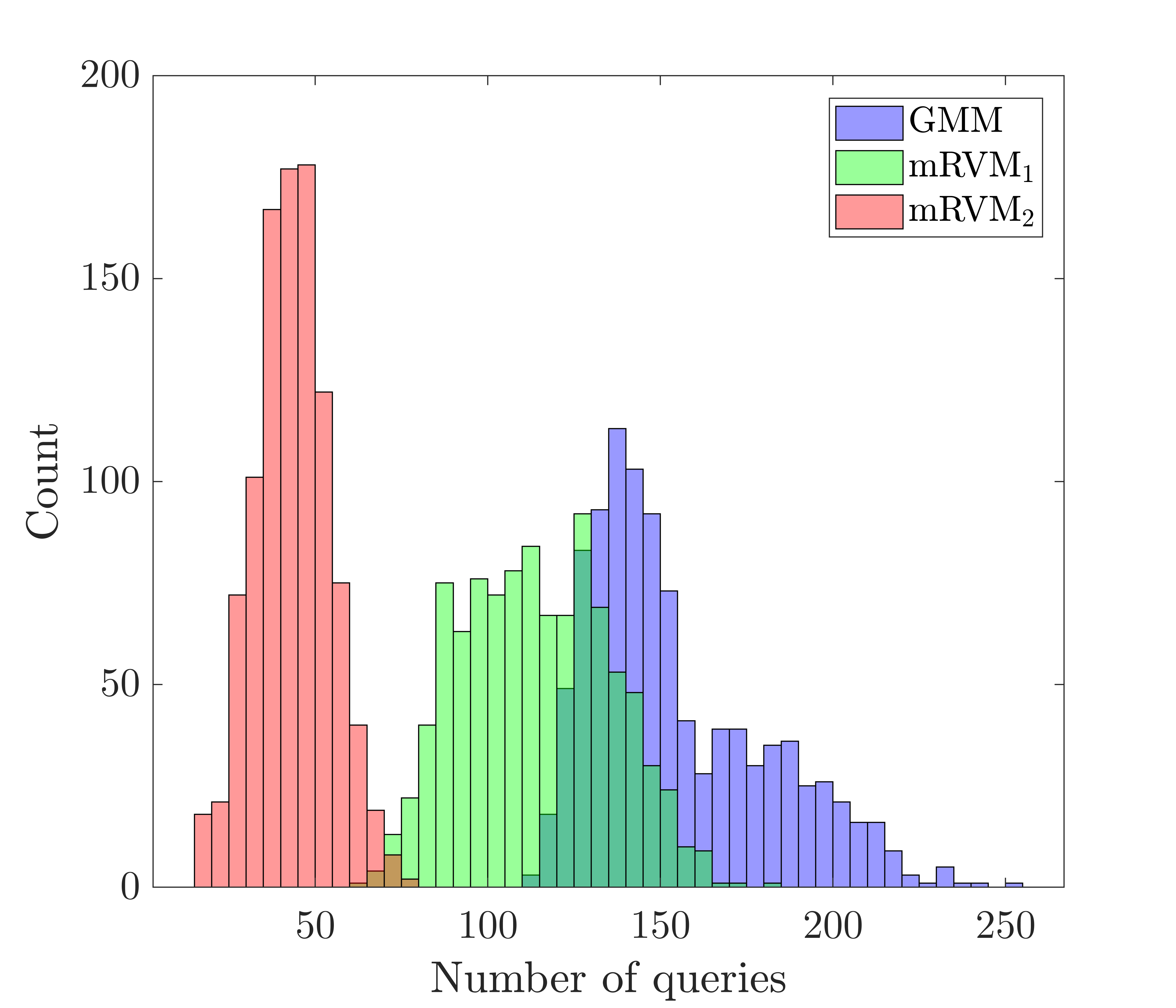}
		}
		\caption{Histograms showing the distribution of the number of queries from 1000 runs of the risk-based active learning of (i) a GMM (blue) (ii) an mRVM$_1$ (green) and (iii) an mRVM$_2$ (red) statistical classifier.}
		\label{fig:hist_rvm_z24}
\end{figure}

\begin{figure}[ht!]
	\centering
		\scalebox{0.4}{
			\includegraphics{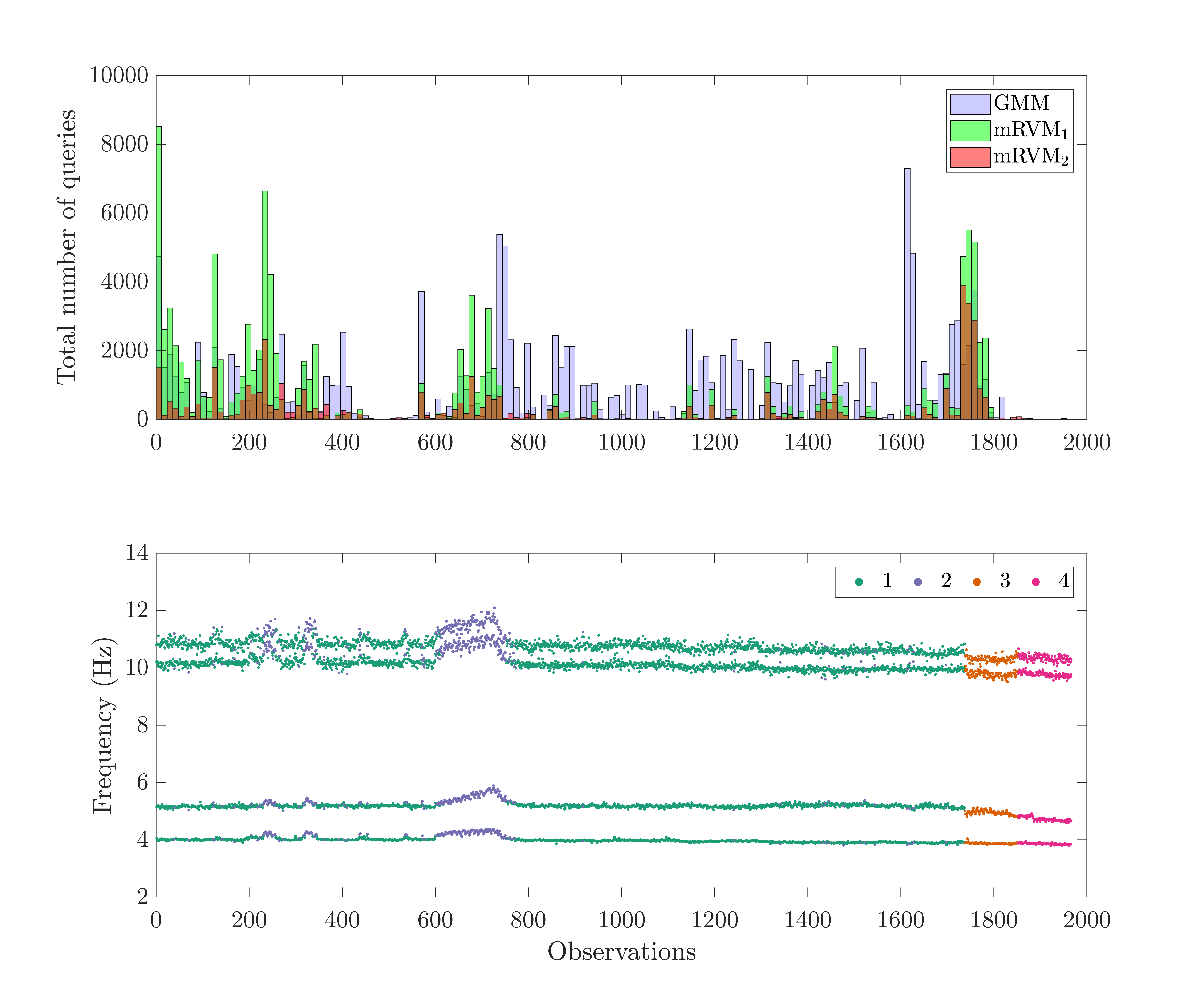}
		}
		\caption{Histograms showing the distribution of the number of queries for each observation in $\mathcal{D}_u$ from 1000 runs of risk-based active learning for (i) a GMM (blue) (ii) an mRVM$_1$ (green) and (iii) an mRVM$_2$ (red) statistical classifier. Below the histograms, a visualisation of the unlabelled dataset $\mathcal{D}_u$ is provided for reference.}
		\label{fig:all_queries_rvm_z24}
\end{figure}

The distributions for the number of queries made by each active learner are shown in Figure \ref{fig:hist_rvm_z24}. Similar to the previous case study, \mRVMa{} and \mRVMb{} both query fewer times on average when compared to the GMM. This result can also be explained by the RVM's ability to represent classes of data using only a few prototypical samples. Between \mRVMa{} and \mRVMb{}, \mRVMb{} yields superior performance in terms of the number of queries made, with both lower mean and variance. As previously mentioned, fewer queries corresponds to a reduced expenditure on inspections, with lower variance indicating more consistent performance.

Finally, Figure \ref{fig:all_queries_rvm_z24} shows the total number of queries for each observation index in $\mathcal{D}_u$. As with Figure \ref{fig:hist_rvm_z24}, it is apparent from Figure \ref{fig:all_queries_rvm_z24} that, overall, the use of RVMs as a statistical classifier reduces the total number of queries made. It can be seen that \mRVMa{} results in increased querying early in the dataset; this is to be expected as it was previously observed that this formulation struggles to construct a model capable of discriminating between classes, resulting in high EVPI for all regions of the feature space. As with the GMM, both RVM approaches result in an increase in the number of queries made as new patterns in the data present -- specifically around observation indices 200, 700, and 1800 in $\mathcal{D}_u$. Again, this result is in agreement with the trends in the $f_1$-score shown in Figure \ref{fig:f1score_rvm_z24}. The relative reduction in the number of queries made between observations 800 to 1600, for the RVMs compared to the GMM, can be attributed to the sparse representation achieved by the RVMs; as data for these classes had already been obtained, the EVPI associated with the classes was sufficiently low such that inspection was not necessary, with the exception of outlying data.

In summary, substituting mRVMs in place of the GMM, results in improved decision-making performance and robustness to sampling bias. Notably, the number of queries made throughout the learning process is also decreased by the introduction of discriminative classifiers into the decision process. These benefits are obtained by virtue of the RVM's characteristic sparsity, in addition to the reduced reliance on assumptions regarding the distribution of data. Whilst \mRVMa{} did yield noteworthy improvements in decision-making performance, \mRVMb{} proved to be the stand-out model formulation.

\section{Discussion}\label{sec:Discussions}

The results presented in Sections \ref{sec:Approaches} and \ref{sec:Z24} hold several implications with respect to the development of decision-supporting SHM systems and thus bear further discussion.

From the case studies presented, it is abundantly clear that variability in performance arises according to several factors. Specifically, variation was observed between the various types of classifier employed within a decision process, and the approach by which the models are learned.

In general, the discriminative classifiers consistently yielded better performance than the GMM (standard and semi-supervised), both in terms of decision accuracy, robustness to sampling bias, and the number of queries made. In contrast, the decision-making performance of the semi-supervised generative models were highly dependent on the assumed distributions used for density estimation and the underlying distribution of data. Based on these results, faced with the conundrum of selecting a classifier to be used in a decision-supporting SHM system, an engineer may opt to use a discriminative classifier without second thought. However, the selection may require more nuanced deliberations. For example, digital twins are a form of asset management technology highly applicable to structures and infrastructure \cite{Grieves2017,Worden2020digital,Gardner2020digitwin}. In this context, SHM systems are a necessary component of digital twins. Furthermore, recent work has identified generative models as being a core component of digital twin technology \cite{Tsialiamanis2021generative}. It follows that, if the context in which SHM system is being applied necessitates a generative model, such as within a digital twin, a discriminative model can be deemed an unsuitable choice of classifier. In this scenario, one may have to accept the effects of sampling bias, or find a method of accounting for them within a generative model. If \textit{a priori} knowledge of data distributions is available (such as via expert elicitation, or transfer learning \cite{Gardner2021b}), then density estimations can be carefully selected such that EM, latent-state smoothing, or an alternative approach to semi-supervision can be incorporated into the risk-based active learning algorithm.

Some degree of performance variability was also observed between \mRVMa{} and \mRVMb{}. While fundamentally, these two classifiers share identical model-forms, they differ in the approach used to select relevance vectors. It was observed that \mRVMb{} outperformed \mRVMa{}. This result occurred as \mRVMb{} has the advantage of beginning the training process by considering the relevance of all data in $\mathcal{D}_l$ simultaneously in the initial set $\mathcal{A}$. In contrast, \mRVMa{} is constrained to consider the contribution of data in the context of the current active set $A$, which is initialised as empty. Although \mRVMb{} proved the better of the two formulations for case studies presented in the current paper, for applications where $\mathcal{D}_l$ grows very large, \mRVMb{} becomes computationally inefficient compared to \mRVMa{}, as discussed in \cite{Psorakis2010}. It follows that, the quantity of data one expects to obtain throughout a decision-supporting SHM campaign must also factor into the selection of classification models learned via a risk-based active approach.

Another consideration that must be made in the selection of statistical classifiers used within SHM decision processes, risk-based active learning, and asset management technologies in general, is the uncertainty quantification with respect to outlying data. As previously mentioned, the generative models presented in the current paper make over-confident predictions for the class labels of outlying data -- a result of data scarcity. This observation has concerning implications for the decision-support systems, as an agent utilising an over-confident classifier will be especially prone to making suboptimal decisions. While these decisions could yield benign consequences such as an unwarranted down-time, they could also lead to missed inspections and ultimately structural failures with potentially severe consequences. Fortunately, the discriminative classifiers considered for the current paper proved to have excellent uncertainty representation for outlying data, yielding high EVPI and therefore reliably triggering inspections where necessary.

Finally, a notable observation is that the number of queries made during the risk-based active learning can be reduced via the choice of statistical classifier within a decision process. This result is highly significant in the context of SHM decision-support because \emph{queries correspond to structural inspections}. This correspondence indicates quite directly, that monetary savings can be achieved by selecting a classifier that results in minimal queries, thereby reducing the total operational costs incurred over the lifetime of structure. The aforementioned result highlights the great value to be gained via SHM systems, and provides further motivation for their development and implementation.

\section{Conclusions}\label{sec:Conclusion}

Risk-based active learning provides a methodology for the online development of statistical classifiers, specifically those being used in decision-supporting systems. In the context of SHM, this process is accomplished by querying labels for observed data via structural inspection, according to the expected value of the label information with respect to an O\&M decision process. Although risk-based active learning has been demonstrated to yield improvements to decision-making performance, the generative classification models considered prior to the current paper suffer ill-effects related to the sampling bias introduced by the guided querying. This bias typically manifests as deterioration in decision-making performance later in the querying process.

The current paper proposed two approaches to address the issue of sampling bias; semi-supervised learning and discriminative classification models. Each approach was applied to a visual example, and an experimental case study. From these case studies, it was found that semi-supervised learning yielded variable performance, dependent on the correspondence between the underlying data distributions and the density estimations selected for the generative models. On the other hand, the discriminative models were found to mitigate the deterioration in decision-making performance. In addition, the case studies provided results with high significance to SHM decision-support in general. Specifically, attention was drawn to the fact that utilising differing classifiers in a decision process model can result in the decision-making agent possessing drastically different attitudes towards outlying data. Finally, it was also shown that the choice of classifier used in the process model can greatly affect the number of queries -- and therefore inspections -- made throughout an SHM campaign. This finding was particularly significant as it implies that, with the careful selection of the classifier within an SHM system, resource expenditure can be greatly reduced.

\section*{Acknowledgements}
The authors would like to acknowledge the support of the UK EPSRC via the Programme Grants EP/R006768/1 and EP/R004900/1. KW would also like to acknowledge support via the EPSRC Established Career Fellowship EP/R003625/1. LAB was supported by Wave 1 of The UKRI Strategic Priorities Fund under the EPSRC Grant EP/W006022/1, particularly the \textit{Ecosystems of Digital Twins} theme within that grant and The Alan Turing Institute.

%
\section*{Conflict of interest}

The authors declare that they have no conflict of interest.

\bibliographystyle{elsarticle-num}
\bibliography{RBAL2_JP}

\begin{thebibliography}{10}
\expandafter\ifx\csname url\endcsname\relax
  \def\url#1{\texttt{#1}}\fi
\expandafter\ifx\csname urlprefix\endcsname\relax\def\urlprefix{URL }\fi
\expandafter\ifx\csname href\endcsname\relax
  \def\href#1#2{#2} \def\path#1{#1}\fi

\bibitem{Farrar2013}
C.~R. Farrar, K.~Worden, {Structural Health Monitoring: A Machine Learning
  Perspective}, John Wiley {\&} Sons, Ltd, 2013.

\bibitem{Grieves2017}
M.~Grieves, J.~Vickers, {Digital twin: Mitigating Unpredictable, Undesirable
  Emergent Behavior in Complex Systems}, in: Transdisciplinary Perspectives on
  Complex Systems, Berlin, Germany, 2017, pp. 85--113.

\bibitem{Niederer2021}
S.~A. Niederer, M.~S. Sacks, M.~Girolami, K.~Willcox, Scaling digital twins
  from the artisanal to the industrial, Nature Computational Science 1~(5)
  (2021) 313--320.

\bibitem{Nielsen2013}
J.~Nielsen, {Risk-Based Operation and Maintenance of Offshore Wind Turbines},
  Ph.D. thesis, Aalborg University (2013).

\bibitem{Hovgaard2016}
M.~K. Hovgaard, R.~Brincker, {Limited memory influence diagrams for structural
  damage detection decision-making}, Journal of Civil Structural Health
  Monitoring 6~(2) (2016) 205--215.
\newblock \href {https://doi.org/10.1007/s13349-016-0153-z}
  {\path{doi:10.1007/s13349-016-0153-z}}.

\bibitem{Hughes2021}
A.~J. Hughes, R.~J. Barthorpe, N.~Dervilis, C.~R. Farrar, K.~Worden, {A
  probabilistic risk-based decision framework for structural health
  monitoring}, Mechanical Systems and Signal Processing 150 (2021) 107339.

\bibitem{Bull2019}
L.~A. Bull, T.~J. Rogers, C.~Wickramarachchi, E.~J. Cross, K.~Worden,
  N.~Dervilis, {Probabilistic active learning: An online framework for
  structural health monitoring}, Mechanical Systems and Signal Processing 134
  (2019) 106294.

\bibitem{Arellano2019}
G.~Mart{\'i}nez-Arellano, S.~Ratchev, Towards an active learning approach to
  tool condition monitoring with bayesian deep learning, in: ECMS, 2019.

\bibitem{Chakraborty2015}
D.~Chakraborty, N.~Kovvali, A.~Papandreou-Suppappola, A.~Chattopadhyay, {An
  adaptive learning damage estimation method for structural health monitoring},
  Journal of Intelligent Material Systems and Structures 26~(2) (2015)
  125--143.

\bibitem{Hughes2022}
A.~J. Hughes, L.~A. Bull, P.~Gardner, R.~J. Barthorpe, N.~Dervilis, K.~Worden,
  {On risk-based active learning for structural health monitoring}, Mechanical
  Systems and Signal Processing 167 (2022) 108569.

\bibitem{Dasgupta2011}
S.~Dasgupta, {Two faces of active learning}, Theoretical Computer Science
  412~(19) (2011) 1767--1781.

\bibitem{Psorakis2010}
I.~Psorakis, T.~Damoulas, M.~Girolami, {Multiclass relevance vector machines:
  Sparsity and accuracy}, IEEE Transactions on Neural Networks 21~(10) (2010)
  1588--1598.

\bibitem{Schwenker2014}
F.~Schwenker, E.~Trentin, Pattern classification and clustering: A review of
  partially supervised learning approaches, Pattern Recognition Letters 37
  (2014) 4--14.

\bibitem{Feng2017}
C.~Feng, M.~Y. Liu, C.~C. Kao, T.~Y. Lee, {Deep Active Learning for Civil
  Infrastructure Defect Detection and Classification}, 2017, pp. 298--306.

\bibitem{Koller2009}
D.~Koller, N.~Friedman, Probabilistic Graphical Models: Principles and
  Techniques, MIT Press, 2009.

\bibitem{Kjaerulff2008}
U.~Kjaerulff, A.~Madsen, {Bayesian Networks and Influence Diagrams: A Guide to
  Construction and Analysis}, Springer, New York, 2008.

\bibitem{Sucar2015}
L.~Sucar, {Probabilistic Graphical Models: Principles and Applications},
  Springer, London, 2015.

\bibitem{Papakonstantinou2014}
K.~G. Papakonstantinou, M.~Shinozuka, Planning structural inspection and
  maintenance policies via dynamic programming and markov processes. part ii:
  Pomdp implementation, Reliability Engineering \& System Safety 130 (2014)
  214--224.

\bibitem{Hamida2020}
Z.~Hamida, J.-A. Goulet, Modeling infrastructure degradation from visual
  inspections using network-scale state-space models, Structural Control and
  Health Monitoring 27~(9) (2020) e2582.

\bibitem{Dasgupta2008}
S.~Dasgupta, D.~Hsu, Hierarchical sampling for active learning, in: Proceedings
  of the 25th International Conference on Machine Learning ACM, 2008.

\bibitem{Valkonen2021}
A.~Valkonen, B.~Glisic, Evaluation tool for assessing the influence of
  structural health monitoring on decision-maker risk preferences, Structural
  Health Monitoring (2021) 1475921721992016.

\bibitem{Vega2020a}
M.~A. Vega, M.~D. Todd, {A variational Bayesian neural network for structural
  health monitoring and cost-informed decision-making in miter gates},
  Structural Health Monitoring (2020).

\bibitem{Hughes2022b}
A.~J. Hughes, R.~J. Barthorpe, K.~Worden, On health-state transition models for
  risk-based structural health monitoring, in: Dynamics of Civil Structures,
  Volume 2, Springer International Publishing, 2022, pp. 49--60.

\bibitem{Fawcett2006}
T.~Fawcett, {An introduction to ROC analysis}, Pattern Recognition Letters
  27~(8) (2006) 861–874.

\bibitem{Damoulas2008}
T.~Damoulas, M.~Girolami, {Probabilistic multi-class multi-kernel learning: On
  protein fold recognition and remote homology detection}, Bioinformatics
  24~(10) (2008) 1264--1270.

\bibitem{Chapelle2006}
O.~Chapelle, B.~Scholkopf, A.~Zien, Semi-Supervised Learning, MIT Press, 2006.

\bibitem{Chen2014}
S.~Chen, F.~Cerda, P.~Rizzo, J.~Bielak, J.~Garrett, J.~Kovačević,
  Semi-supervised multiresolution classification using adaptive graph filtering
  with application to indirect bridge structural health monitoring, IEEE
  Transactions on Signal Processing 62~(11) (2014) 2879--2893.

\bibitem{Bull2020}
L.~A. Bull, K.~Worden, N.~Dervilis, {Towards semi-supervised and probabilistic
  classification in structural health monitoring}, Mechanical Systems and
  Signal Processing 140 (2021) 106653.

\bibitem{Dempster1977}
A.~P. Dempster, N.~M. Laird, D.~B. Rubin, {Maximum Likelihood from Incomplete
  Data via the EM Algorithm}, Journal of the Royal Statistical Society 39~(1)
  (1977) 1--38.

\bibitem{Einicke2012}
G.~Einicke, Smoothing, filtering and prediction: Estimating the past, present
  and future, BoD--Books on Demand, 2012.

\bibitem{Binder1997}
J.~Binder, K.~Murphy, S.~Russell, Space-efficient inference in dynamic
  probabilistic networks, in: Proceedings of the Fifteenth International Joint
  Conference on Artifical Intelligence - Volume 2, IJCAI'97, Morgan Kaufmann
  Publishers Inc., San Francisco, CA, USA, 1997, p. 1292–1296.

\bibitem{Cortes1995}
C.~Cortes, V.~Vapnik, Support-vector networks, Machine learning 20~(3) (1995)
  273--297.

\bibitem{Platt1999}
J.~Platt, Probabilistic outputs for support vector machines and comparisons to
  regularized likelihood methods, Advances in Large Margin Classifiers 10~(3)
  (1999) 61--74.

\bibitem{Tipping2001}
M.~E. Tipping, {Sparse Bayesian learning and the relevance vector machine},
  Journal of Machine Learning Research 1 (2001) 211--244.

\bibitem{Damoulas2009}
T.~Damoulas, M.~Girolami, {Combining feature spaces for classification},
  Pattern Recognition 42~(11) (2009) 2671--2683.

\bibitem{Manocha2007}
S.~Manocha, M.~Girolami, {An empirical analysis of the probabilistic k-nearest
  neighbour classifier}, Pattern Recognition Letters 28~(13) (2007) 1818--1824.

\bibitem{Maeck2003}
J.~Maeck, G.~De~Roeck, {Description of Z24 benchmark}, Mechanical Systems and
  Signal Processing 17~(1) (2003) 127--131.

\bibitem{Maeck2001}
J.~Maeck, B.~Peeters, G.~De~Roeck, {Damage identification on the Z24-bridge
  using vibration monitoring}, Smart Materials and Structures 10~(3) (2001)
  512--517.

\bibitem{DeRoeck2003}
G.~De~Roeck, {The state-of-the-art of damage detection by vibration monitoring:
  the SIMCES experience}, Structural Control Health Monitoring 10~(2) (2003)
  127--134.

\bibitem{Peeters2001}
B.~Peeters, G.~De~Roeck, {One-year monitoring of the Z24-Bridge: environmental
  effects versus damage events}, Earthquake Engineering Structural Dynamics
  30~(2) (2001) 149--171.

\bibitem{BullThesis}
L.~A. Bull, {Towards Probabilistic and Partially-Supervised Structural Health
  Monitoring}, Ph.D. thesis, University of Sheffield (2020).

\bibitem{Worden2020digital}
K.~Worden, E.~J. Cross, R.~J. Barthorpe, D.~J. Wagg, P.~Gardner, On digital
  twins, mirrors, and virtualizations: Frameworks for model verification and
  validation, ASCE-ASME J Risk and Uncert in Engrg Sys Part B Mech Engrg 6~(3)
  (2020).

\bibitem{Gardner2020digitwin}
P.~Gardner, M.~Dal~Borgo, V.~Ruffini, A.~J. Hughes, Y.~Zhu, D.~J. Wagg, Towards
  the development of an operational digital twin, Vibration 3~(3) (2020)
  235--265.

\bibitem{Tsialiamanis2021generative}
G.~Tsialiamanis, D.~J. Wagg, N.~Dervilis, K.~Worden, On generative models as
  the basis for digital twins, Data-Centric Engineering 2 (2021).

\bibitem{Gardner2021b}
P.~Gardner, L.~A. Bull, J.~Gosliga, N.~Dervilis, K.~Worden, {Foundations of
  population-based SHM, Part III: Heterogeneous populations – mapping and
  transfer}, Mechanical Systems and Signal Processing 148 (2021) 107142.

\bibitem{Murphy2012}
K.~P. Murphy, Machine Learning: a Probabilistic Perspective, MIT press, 2012.

\bibitem{Gelman2013}
A.~Gelman, J.~B. Carlin, H.~S. Stern, D.~B. Dunson, A.~Vehtari, D.~Rubin,
  Bayesian Data Analysis, Chapman and Hall/CRC, 2013.

\bibitem{Barber2012}
D.~Barber, Bayesian Reasoning and Machine Learning, Cambridge University Press,
  2012.

\bibitem{Tipping2003}
M.~E. Tipping, A.~C. Faul, Fast marginal likelihood maximisation for sparse
  bayesian models, in: C.~M. Bishop, B.~J. Frey (Eds.), Proceedings of the
  Ninth International Workshop on Artificial Intelligence and Statistics,
  Vol.~R4 of Proceedings of Machine Learning Research, PMLR, 2003, pp.
  276--283.

\end{thebibliography}

\begin{appendices}
	
\section{Gaussian Mixture Models}
\label{app:GMMs}
\setcounter{equation}{0}
\renewcommand{\theequation}{\thesection.\arabic{equation}}

As stated in Section \ref{sec:GMM}, the GMM represents each class $k$ via a multivariate normal distribution, parameterised by mean $\bm{\mu}_k$ and covariance $\Sigma_k$. In addition to the mean and covariance parameters of the Gaussian components, the mixture model requires specification of $p(y_t)$,

\begin{equation}
	y_t \sim \text{Cat}(\bm{\lambda})
\end{equation}

\noindent
where Cat denotes the categorical distribution parametrised by \textit{mixing proportions} $\bm{\lambda} = \{\lambda_1,\ldots,\lambda_K\}$, such that,

\begin{equation}
	P(y_t = k) = \lambda_k
\end{equation}

\noindent
and,

\begin{equation}
	\sum_{k=1}^{K} \lambda_k = 1
\end{equation}

The parameters of the GMM that specify the joint distribution $p(y_t,\mathbf{x}_t)$ can be summarised as,

\begin{equation}
	\bm{\Theta} = \{(\bm{\mu}_1,\Sigma_1,\lambda_1),\ldots,(\bm{\mu}_K,\Sigma_K,\lambda_K)\}
\end{equation}

To avoid over-fitting, a Bayesian methodology was adopted here to infer $\bm{\Theta}$ from $\mathcal{D}_l$. In the Bayesian approach, the parameters in $\bm{\Theta}$ are considered random variables with prior distributions placed over them. As a conjugate distribution to the multivariate Gaussian, a normal-inverse-Wishart prior was selected such that,

\begin{equation}
	\bm{\mu}_k , \Sigma_k | y_t=k \sim \text{NIW}(\bm{m}_0,\kappa_0,v_0,S_0)
\end{equation}

\noindent
where $\bm{m}_0$, $\kappa_0$, $v_0$ and $S_0$ are hyperparameters of the mixture model. These hyperparameters can be interpreted in the following way \cite{Murphy2012}: $\bm{m}_0$ is the prior mean for each class mean $\bm{\mu}_k$, and $\kappa_0$ specifies the strength of the prior; $S_0$ is proportional to the prior mean for each class covariance $\Sigma_k$, and $v_0$ specifies the strength of that prior. The hyperparameters were specified such that each class $y_t$ was initially represented as a zero-mean and unit-variance Gaussian distribution.

For conjugacy with the categorical distribution, a Dirichlet prior was placed over the mixing proportions $\bm{\lambda}$,

\begin{equation}
	p(\bm{\lambda}) = \text{Dir}(\bm{\alpha}) \propto \prod_{k=1}^{K}\lambda_k^{\alpha_k -1}
\end{equation}

\noindent
where $\bm{\alpha} = \{\alpha_1,\ldots,\alpha_K\}$ are hyperparameters of the mixture model. For the current paper, where $K=4$, these hyperparameters were specified such that, initially, each class is equally likely, i.e.\ the prior probability of each class in the mixture model was 0.25.

By using conjugate priors, posterior distributions over the parameters of the GMM can be calculated from $\mathcal{D}_l$ analytically. The posterior NIW distribution is given by \cite{Gelman2013},

\begin{equation}
	\bm{\mu}_k , \Sigma_k | y_t = k, \mathcal{D}_l \sim \text{NIW}(\bm{m}_n,\kappa_n,v_n,S_n)
\end{equation} 

\noindent
where $\bm{m}_n$, $\kappa_n$, $v_n$, $S_n$ are the updated parameters and are computed as follows,

\begin{equation}\label{eq:posterior_m}
	\bm{m}_n = \frac{\kappa_0}{\kappa_0 + n_k}\bm{m}_0 + \frac{n_k}{\kappa_0 + n_k}\bar{\mathbf{x}}_k
\end{equation}

\begin{equation}\label{eq:posterior_k}
	\kappa_n = \kappa_0 + n_k
\end{equation}

\begin{equation}\label{eq:posterior_v}
	v_n = v_0 + n_k
\end{equation}

\begin{equation}\label{eq:posterior_S}
	S_n = S_0 + S + \kappa_0 \bm{m}_0 \bm{m}_0^{\top} - \kappa_n \bm{m}_n \bm{m}_n^{\top}
\end{equation}

\noindent
where $n_k$ is the number of observations in $\mathcal{D}_l$ with label $k$, $\bar{\mathbf{x}}_k$ is the sample mean of observations with label $k$, and $S$ is the empirical scatter matrix given by the uncentered sum-of-squares for observations in class $k$, $S = \sum_{i \in \mathbb{I}_k}\mathbf{x}_i\mathbf{x}_i^{\top}$ where $\mathbb{I}_k$ is the set of indices for observations with label $k$.

The posterior for the mixing proportions $\bm{\lambda}$, remains Dirichlet distributed, and is given by \cite{Gelman2013},

\begin{equation}\label{eq:posterior5}
	p(\bm{\lambda}|\mathcal{D}_l) \propto \prod_{k=1}^{K}\lambda_k^{n_k + \alpha_k-1}
\end{equation}

Class predictions for unlabelled data in $\mathcal{D}_u$ can be made by obtaining the posterior predictive distributions over the labels and observations; this is achieved by marginalising out the parameters of the model. The posterior predictive distribution for unlabelled observations is obtained via the following marginalisation,

\begin{equation}
	p(\tilde{\mathbf{x}}_t|y_t = k, \mathcal{D}_l) = \int\int p(\tilde{\mathbf{x}}_t|\bm{\mu}_k,\Sigma_k) p(\bm{\mu}_k,\Sigma_k|y_t = k, \mathcal{D}_l)d\bm{\mu}_k d\Sigma_k
\end{equation}

\noindent
resulting in the Student-\textit{t} distribution \cite{Murphy2012},

\begin{equation}\label{eq:studentT}
	(\tilde{\mathbf{x}}_t | y_t = k, \mathcal{D}_l) \sim \mathcal{T}  \bigg( \bm{m}_n,\frac{\kappa_n + 1}{\kappa_n(v_n - D + 1)}S_n,v_n - D +1 \bigg)
\end{equation}

\noindent
where $\bm{m}_n$, $\kappa_n$, $v_n$, $S_n$ are the updated hyperparameters and $D$ is the dimensionality of the feature space. Here, the first two parameters of the Student-\textit{t} distribution correspond to the mean and scale, respectively. The third parameter specifies the \textit{degrees of freedom} and relates to the probability mass found in the tails of the distribution. The full functional form of the Student-\textit{t} distribution can be found in \cite{Murphy2012}.

Via a similar procedure, the posterior predictive distribution over the labels is obtained via the following marginalisation,

\begin{equation}
	p(\tilde{y}_t|\mathcal{D}_l) = \int p(\tilde{y}_t| \bm{ \lambda}) p( \bm{\lambda} | \mathcal{D}_l) d \bm{ \lambda}
\end{equation}

\noindent
resulting in,
\begin{equation}\label{eq:mixpropupdate}
	p(\tilde{y}_t = k | \mathcal{D}_l) = \frac{n_k + \alpha_k}{n + \alpha_0}
\end{equation}

\noindent
where $n = \sum_{k=1}^{K}n_k$ and $\alpha_0 = \sum_{k=1}^{K}\alpha_k$.

Finally, the predictive distribution for the class labels given a new unlabelled observation $\tilde{\mathbf{x}}_t$ can be obtained using the posterior predictive distribution and applying Bayes' rule \cite{Bull2019},

\begin{equation}\label{eq:labelpred}
	p(\tilde{y}_t = k|\tilde{\mathbf{x}}_t,\mathcal{D}_l) = \frac{p(\tilde{\mathbf{x}}_t|\tilde{y}_t = k, \mathcal{D}_l)p(\tilde{y}_t=k|\mathcal{D}_l)}{p(\tilde{\mathbf{x}}_t|\mathcal{D}_l)}
\end{equation}

\section{Expectation-Maximisation for Gaussian Mixture Models}
\label{app:EM_GMMs}
\setcounter{equation}{0}
\renewcommand{\theequation}{\thesection.\arabic{equation}}

The aim of semi-supervised learning via the EM algorithm is to infer updated distribution parameters $\bm{\Theta}$ from $\mathcal{D} = \mathcal{D}_l \cup \mathcal{D}_u$. The \textit{maximum a posteriori} MAP estimate of these updated parameters is specified as follows,

\begin{equation}\label{eq:EM1}
	\bm{\Theta} | \mathcal{D} = \text{argmax}_{\bm{\Theta}} \biggl\{ \frac{p(\mathcal{D}|\bm{\Theta})p(\bm{\Theta})}{p(\mathcal{D})} \biggr\} = \text{argmax}_{\bm{\Theta}} \biggl\{ \frac{p(\mathcal{D}_u|\bm{\Theta})p(\mathcal{D}_l|\bm{\Theta})p(\bm{\Theta})}{p(\mathcal{D}_u,\mathcal{D}_l)} \biggr\}
\end{equation}

Implicit in the factorisation of $p(\mathcal{D}|\bm{\Theta})$ in equation (\ref{eq:EM1}), is the assumption that $\mathcal{D}_l$ and $\mathcal{D}_u$ are conditionally independent. This assumption holds for random querying, as samples selected in this manner are independent and identically distributed (iid). Unfortunately, active learning violates this assumption, as data in $\mathcal{D}_l$ are not iid because of the preferential querying process and iterative model updating \cite{Dasgupta2008,Dasgupta2011}. Convenient assumptions such as this are frequently relied upon in statistical and engineering analyses (particularly in active learning contexts \cite{Bull2019}). As such, the assumption is embraced for the current case study in order to demonstrate the decision-making performance that one may achieve despite the violation.

To circumvent numerical instabilities, the MAP estimate in equation (\ref{eq:EM1}) is formulated as a maximisation of the expected joint log-likelihood across $\mathcal{D}$ \cite{Chapelle2006},

\begin{multline}\label{eq:EMloglik}
	\mathcal{L}(\bm{\Theta}|\mathcal{D}_l,\mathcal{D}_u) \propto \sum_{t=1}^{m}\log \sum_{k=1}^{K} p(\tilde{\mathbf{x}}_t|y_t = k ,\bm{\Theta})p(y_t = k | \bm{\Theta}) \\ + \sum_{t=1}^{n} \log [p(\mathbf{x}|y_t = k)p(y_t = k|\bm{\Theta})] + \log p(\bm{\Theta})
\end{multline}

\noindent where the first term corresponds to the log-likelihood of the model over $\mathcal{D}_u$, the second term corresponds to the log-likelihood of the model over $\mathcal{D}_l$ and the final term is the log-prior-likelihood of the model parameters. It should be noted that the first term contains a summation over the label space; this marginalises out $y_t$, which is considered a latent variable for data in $\mathcal{D}_u$.

During each E-step of the EM algorithm, the unlabelled observations are classified using the current estimate of the model parameters. During the M-step, updates for parameters $\bm{\Theta}$ are found using the predicted labels determined via the E-step, in addition to the acquired labelled data.

The E-step and M-step are more formally defined as follows \cite{Dempster1977}.

\textit{E-step}: A \textit{responsibility matrix} $R$ is computed for the unlabelled data corresponding to the posterior label prediction given by equation (\ref{eq:labelpred}),

\begin{equation}
	R[t,k] = r_{tk} = p(\tilde{y}_t = k | \tilde{\mathbf{x}}_t,\bm{\Theta}), \quad \forall \tilde{\mathbf{x}}_t \in \mathcal{D}_u
\end{equation}

The posterior label predictions for observations in $\mathcal{D}_l$ are given by the known labels $y_t$, and can be represented using discrete delta functions \cite{Barber2012},

\begin{equation}
	p(y_t = k | \mathbf{x}_t) = \delta_{k,y_t}, \quad \forall (\mathbf{x}_t,y_t) \in \mathcal{D}_l
\end{equation}

\noindent
where $\delta_{k,y_t}$ is the Kronecker delta function -- equal to 1 when $y_t = k$, and 0 otherwise.

Whereas, for $\mathcal{D}_l$, the number of observations corresponding to each class $n_k$ are known, these values are indeterminate for data in $\mathcal{D}_u$. As such, it is convenient to define for each class their \textit{effective counts} $r_k$ in $D_u$; these may be calculated from the responsibility matrix as follows \cite{Murphy2012},

\begin{equation}
	r_k = \sum_{t=1}^{m} r_{tk}
\end{equation}

The total (effective) counts per class over $\mathcal{D}$ can be summarised as,

\begin{equation}
	N_k = n_k + r_k
\end{equation}

\textit{M-step}: Updates for $\bm{\Theta}$ are computed via modified versions of equations (\ref{eq:posterior_m}) to (\ref{eq:posterior_S}) and equation (\ref{eq:mixpropupdate}). The mean and covariance parameters are updated as follows,

\begin{equation}\label{eq:EMupdate1}
	\bm{m}_n = \frac{\kappa_0}{\kappa_0 + N_k}\bm{m}_0 + \frac{N_k}{\kappa_0 + N_k}\bar{\mathbf{x}}_k
\end{equation}

\begin{equation}
	\bar{\mathbf{x}}_k \overset{\Delta}{=} \frac{\sum_{t=1}^{n}\delta_{k,y_t}\mathbf{x}_t + \sum_{t=1}^{m}r_{tk}\tilde{\mathbf{x}}_t}{N_k}
\end{equation}

\begin{equation}
	\kappa_n = \kappa_0 + N_k
\end{equation}

\begin{equation}
	v_n = v_0 + N_k
\end{equation}

\begin{equation}
	S_n = S_0 + S_k + \kappa_0 \bm{m}_0 \bm{m}_0^{\top} - \kappa_n \bm{m}_n \bm{m}_n^{\top}
\end{equation}

\begin{equation}
	S_k \overset{\Delta}{=} \sum_{t=1}^{n}\delta_{k,y_t}\mathbf{x}_t\mathbf{x}_t^{\top} + \sum_{t=1}^{m}r_{tk}\tilde{\mathbf{x}}_t \tilde{\mathbf{x}}_t^{\top}
\end{equation}

\noindent which lead to MAP estimates given by,

\begin{equation}
	\hat{\bm{\mu}}_k = \bm{m}_n
\end{equation}

\begin{equation}
	\hat{\Sigma}_k = \frac{S_n}{v_n+D+2}
\end{equation}

The mixing proportions $\bm{\lambda}$ are updated as follows,

\begin{equation}\label{eq:EMupdate9}
	\frac{\alpha_k + N_k -1}{\alpha_0 + N - K}
\end{equation}

\noindent where $N = |\mathcal{D}| = n + m$.

\section{Multiclass Relevance Vector Machines}
\label{app:mRVMs}
\setcounter{equation}{0}
\renewcommand{\theequation}{\thesection.\arabic{equation}}

For the multiclass classification of an unlabelled data point $\tilde{\mathbf{x}}_t$, mRVMs employ a fundamentally (generalised) linear model-form as a foundation,

\begin{equation}
	\mathbf{f}_t = \mathbf{W}^{\top}\mathbf{k}(\mathbf{x}_l,\tilde{\mathbf{x}}_t)
\end{equation}

\noindent where $\mathbf{f}_t = \{f_1,\ldots,f_K\}^{\top}$ is a vector of $K$ auxiliary variables that provide a ranking system by which the class membership of an unlabelled data point may be assessed. $\mathbf{k}(\mathbf{x}_l,\tilde{\mathbf{x}}_t)$ is an $n \times 1$ vector for which the $i^{\text{th}}$ element is specified by the \textit{kernel function} $k(\mathbf{x}_i,\tilde{\mathbf{x}}_t)$. The kernel function specifies a set of basis functions that reflect the similarity between $\tilde{\mathbf{x}}_t$ and training inputs $\mathbf{x}_l = \{\mathbf{x}_i|(\mathbf{x}_i,y_i) \in \mathcal{D}_l \}^n_{i=1}$; nonlinearity can be introduced into the RVM by selecting a nonlinear kernel function. For compactness, the vector $\mathbf{k}(\mathbf{x}_l,\tilde{\mathbf{x}}_t)$ is denoted as $\mathbf{k}_t$ herein. $\mathbf{W} = \{ \mathbf{w}_1, \ldots,\mathbf{w}_K\}$ is an $n \times K$ matrix of tunable parameters referred to as \textit{weights}, and where $\mathbf{w}_k = \{w_{1,k},\ldots,w_{n,k}\}^{\top}$. These weights act as a voting system that indicate which data in $\mathcal{D}_l$ are important, or `relevant', for discriminating between classes.

In accordance with \cite{Psorakis2010}, the auxiliary variables in $\mathbf{f}_t$ are assumed to adhere to a standardised noise model:

\begin{equation}
	f_k|\mathbf{w}_k,\mathbf{k}_t \sim \mathcal{N}(\mathbf{w}_k^{\top}\mathbf{k}_t,1)
\end{equation}

\noindent Predicted class labels $\hat{y}_t$ are assigned to otherwise unlabelled data via the auxiliary variables $\mathbf{f}_t$ by using a criterion specified by the multinomial probit link,

\begin{equation}\label{eq:RVMLabel1}
	\hat{y}_t = k \iff f_k > f_j \text{ } \forall j \neq k
\end{equation}

\noindent Moreover, a probabilistic representation of class membership can be obtained via the following marginalisation,

\begin{equation}
	p(\tilde{y}_t = k|\mathbf{W},\mathbf{k}_t) = \int p(\tilde{y}_t = k|\mathbf{f}_t)p(\mathbf{f}_t|\mathbf{W},\mathbf{k}_t) d\mathbf{f}_t
\end{equation}

\noindent where $p(\tilde{y}_t = k|\mathbf{f}_t) = \delta_{k,\hat{y}_t}$. This marginalisation yields the multinomial probit likelihood \cite{Damoulas2008},

\begin{equation}\label{eq:multi_probit_lik}
	p(\tilde{y}_t = k | \mathbf{W},\mathbf{k}_t) = \mathbb{E}_{p(u)} \Biggl[ \prod_{j \neq k} \Phi (u + (\mathbf{w}_k - \mathbf{w}_{j})^{\top}\mathbf{k}_t) \Biggr]
\end{equation}

\noindent where $u \sim \mathcal{N}(0,1)$ and $\Phi$ denotes the Gaussian cumulative distribution function. Here, it is worth noting that, in practice, equation (\ref{eq:multi_probit_lik}) cannot be computed analytically and is instead approximated via Gauss-Hermite quadrature \cite{Psorakis2010}.

Having specified the predictive model, the relevant supervised learning problem for mRVMs can be expressed as the Bayesian inference of weights $\mathbf{W}$ from $\mathcal{D}_l$. In order to conduct Bayesian inference, a prior must first be placed upon the model parameters $\mathbf{W}$; in this case,

\begin{equation}
	w_{i,k} \sim \mathcal{N}(0,\alpha_{i,k}^{-1})
\end{equation}

\noindent where the scale parameters $\alpha_{i,k}$ can be summarised in an $n \times K$ matrix $\mathbf{A}$ and are assigned the following hyperprior,

\begin{equation}
	\alpha_{i,k} \sim \Gamma(\tau,\nu)
\end{equation}

\noindent where $\tau$ and $\nu$ are hyperparameters related to the shape and scale, respectively. Given small values for the hyperparameters $\tau$ and $\nu$, the prior and hyperprior presented above result in a Student-$t$ distribution with zero-mean and small variance over the weights in $\mathbf{W}$. This restrictive distribution causes few weights to be non-zero, thereby inducing sparsity in the model \cite{Tipping2001,Psorakis2010}.

The training of the mRVM model is accomplished by updating the model parameters via an EM algorithm \cite{Dempster1977}. For a detailed exposition of the learning procedure, the reader is directed to \cite{Damoulas2009}. Here, the EM steps are provided.

The model weight parameters are updated as the MAP estimate of the posterior distribution over the weights; given by $\hat{\mathbf{W}} = \argmax_{\mathbf{W}}p(\mathbf{W}|\mathbf{F}_l,\mathbf{K}_l,\mathbf{A},\mathcal{D}_l)$, where $\mathbf{F}_l = \{ \mathbf{f}_{1},\ldots,\mathbf{f}_{K} \}^{\top}$ are auxiliary variables for data in $\mathcal{D}_l$, and $\mathbf{K}_l$ denotes $\mathbf{K}(\mathbf{x}_l,\mathbf{x}_l)$. It follows that, for a given class, the update for the weights across data in $\mathcal{D}_l$ is given by,

\begin{equation}\label{eq:update_w}
	\hat{\mathbf{w}}_k = (\mathbf{K}_l\mathbf{K}_l^{\top} + \mathbf{A}_k)^{-1}\mathbf{K}_l\mathbf{f}_k^{\top}
\end{equation}

\noindent where $\mathbf{A}_k$ is a diagonal matrix formed from the $k^{\text{th}}$ column of $\mathbf{A}$, i.e.\ $\mathbf{A}_k = \text{diag}( \alpha_{1,k},\ldots,\alpha_{n,k} )$.

Following the formulation in \cite{Psorakis2010,Damoulas2009}, the posterior distribution over the auxiliary variables is derived to be a product of conically-truncated Gaussian distributions. For a given class $j$, the auxiliary variables can be updated $\forall k\neq j$ as follows,

\begin{equation}\label{eq:update_f1}
	\tilde{f}_{k,i} \leftarrow \hat{\mathbf{w}}_{k}^{\top}\mathbf{k}_{i} - \frac{\mathbb{E}_{p(u)}\bigl[\mathcal{N}(\hat{\mathbf{w}}_{k}^{\top}\mathbf{k}_i-\hat{\mathbf{w}}_{j}^{\top}\mathbf{k}_i,1) \prod_{\kappa \neq j,k}\Phi(u + \hat{\mathbf{w}}_{j}^{\top}\mathbf{k}_i - \hat{\mathbf{w}}_{\kappa}^{\top}\mathbf{k}_i)\bigr]}{\mathbb{E}_{p(u)}\bigl[\Phi(u + \hat{\mathbf{w}}_{j}^{\top}\mathbf{k}_i - \hat{\mathbf{w}}_{k}^{\top}\mathbf{k}_i) \prod_{\kappa \neq j,k}\Phi(u + \hat{\mathbf{w}}_{j}^{\top}\mathbf{k}_i - \hat{\mathbf{w}}_{\kappa}^{\top}\mathbf{k}_i)\bigr]}
\end{equation}

\noindent and for class $j$,

\begin{equation}\label{eq:update_f2}
	\tilde{f}_{j,i} \leftarrow \hat{\mathbf{w}}_{j}^{\top}\mathbf{k}_{i} - \sum_{\kappa \neq j} (\tilde{f}_{\kappa,i} - \hat{\mathbf{w}}_{\kappa}^{\top}\mathbf{k}_{i})
\end{equation}

\noindent where $\tilde{f}_{k,i}$ denotes the mean of the distribution over the latent auxiliary variable corresponding to class $k$ for data point $i$ in $\mathcal{D}_l$.

By continuing to follow \cite{Damoulas2009}, one can derive the update for the hyperpriors $\alpha_{i,k}$ as the mean of the posterior distribution $p(\mathbf{A}|\mathbf{W},\mathcal{D}_l)$, given by,

\begin{equation}\label{eq:update_alpha}
	\tilde{\alpha}_{i,k} = \frac{2\tau + 1}{w_{i,k}^{2}+2\nu}
\end{equation}

The iterative learning process requires the repeated application of the updates given in equations (\ref{eq:update_alpha}), (\ref{eq:update_w}), (\ref{eq:update_f1}) and (\ref{eq:update_f2}), until a convergence criterion is met. Suitable convergence criteria are provided in \cite{Psorakis2010}.

As discussed in Section \ref{sec:RVMs}, RVMs utilise a subset of `relevant' samples to construct basis functions, with \mRVMa{} adopting a bottom-up approach and \mRVMb{} adopting a top-down approach.

For \mRVMa{}, $\mathcal{A}$ is initialised as an empty set with samples subsequently added or removed from $\mathcal{A}$ based upon their contribution to the objective function given by the marginal log-likelihood $\mathcal{L}(\mathbf{A}) = \log p(\mathbf{F}_l|\mathbf{K_l},\mathbf{A}) = \log \int p(\mathbf{F}_l|\mathbf{K}_l,\mathbf{W})p(\mathbf{W}|\mathbf{A})d\mathbf{W}$. In accordance with \cite{Psorakis2010}, to ensure $\mathcal{L}(\mathbf{A})$ is differentiable such that the fast type-II Maximum-Likelihood approach \cite{Tipping2003}, can be employed, it is assumed that the scale parameter for each sample is common across classes, i.e.\ $\forall k \in \{1,\ldots,K \}, \:\alpha_{i,k} = \alpha_{i}$. By following the procedure detailed in \cite{Tipping2003}, one arrives at the following marginal likelihood decomposition,

\begin{equation}
	\mathcal{L}(\mathbf{A}) = \mathcal{L}(\mathbf{A}_{-i}) + \ell(\alpha_i)
\end{equation}

\noindent where,

\begin{equation}
	\mathcal{L}(\mathbf{A}_{-i}) = \sum_{k=1}^{K}-\frac{1}{2}\biggl[ N\log2\pi +\log |\mathcal{C}_{-i}| + \mathbf{f}_k^{\top} \mathcal{C}_{-i}^{-1} \mathbf{f}_k \biggr]
\end{equation}

\noindent and,

\begin{equation}\label{eq:loglik_alpha}
	\ell(\alpha_i) = \sum_{k=1}^{K} \frac{1}{2} \biggl[ \log \alpha_i - \log(\alpha_i + s_i) + \frac{q_{k,i}^2}{\alpha_i + s_i} \biggr]
\end{equation}

\noindent with $\mathcal{C}_{-i} = \mathbf{I}_{n-1} + \mathbf{K}_{l,-i}^{\top}\mathbf{A}_{-i}^{-1}\mathbf{K}_{l,-i}$, where $\mathbf{I}_{n-1}$ denotes the identity matrix of size $n-1$, and where the subscript $-i$ is used to indicate matrices with entries corresponding to the $i^{\text{th}}$ data point in $\mathcal{D}_l$ removed. The quantities $s_i$ and $q_{k,i}$, introduced in equation (\ref{eq:loglik_alpha}) are given by \cite{Tipping2003},

\begin{equation}
	s_i = \mathbf{k}_i^{\top}\mathcal{C}_{-i}^{-1}\mathbf{k}_i \quad \text{and} \quad q_{k,i} = \mathbf{k}_i^{\top}\mathcal{C}_{-i}^{-1}\mathbf{f}_k
\end{equation}

\noindent and can be interpreted as a `sparsity factor' and `quality factor', respectively. The sparsity factor indicates how much the descriptive information provided by the $i^{\text{th}}$ data point is already provided by the existing samples. The quality factor provides a measure of the $i^{\text{th}}$ sample's ability to describe class $k$. Maximising $\mathcal{L}(\mathbf{A})$ with respect to $\alpha_i$ by following the procedure presented in \cite{Tipping2003,Psorakis2010}, one can quantify the contribution of data point $i$ to the objective function as,

\begin{equation}
	\theta_i = \sum_{k=1}^{K}q_{k,i}^{2} - Ks_i
\end{equation}

For each iteration in the learning procedure, the contribution $\theta_i$ is then used to construct $\mathcal{A}$ with samples satisfying $\theta_i > 0$ included, and other samples excluded. The model parameter update in equation (\ref{eq:update_w}) can be expressed utilising the sparse subset of data $\mathcal{A}$ as follows:

\begin{equation}
	\hat{\mathbf{W}}_{\ast} = (\mathbf{K}_{\ast}\mathbf{K}_{\ast}^{\top}+\mathbf{A}_{\ast})^{-1} \mathbf{K}_{\ast}\tilde{\mathbf{F}}^{\top}
\end{equation}

\noindent where $\mathbf{K}_{\ast}$ denotes $\mathbf{K}(\mathbf{x}_l,\mathcal{A})$ and is $n^{\ast} \times n$, and $\mathbf{A}_{\ast}$ is $n^{\ast} \times n^{\ast}$. Finally, the update given in equation (\ref{eq:update_alpha}) becomes,

\begin{equation}
	\alpha_i = \frac{Ks_i^{2}}{\theta_i}
\end{equation}

As discussed in Section \ref{sec:RVMs}, \mRVMb{} prunes samples from $\mathcal{A}$. This is achieved by excluding samples with scales $\alpha_{i,k}$ sufficiently large that $w_{i,k}$ is negligible. The $i^{\text{th}}$ data point can be considered insignificant and can be removed from $\mathcal{A}$ when $\alpha_{i,k} > 10^5 \; \forall k\in\{ 1,\ldots,K \}$.

\end{appendices}

\end{document}